\documentclass{tlp}

\newcommand\bcmdtab{\noindent\bgroup\tabcolsep=0pt%
  \begin{tabular}{@{}p{10pc}@{}p{20pc}@{}}}
\newcommand\ecmdtab{\end{tabular}\egroup}

\usepackage{fleqn}
\usepackage{epsfig}
\usepackage{amsfonts}

\def\cal{\mathcal}
\def\qed{}

\def\Fap{{\cal F}_{\alpha,\varphi}^{\Gamma}}
\def\E{{\cal E}}
\def\P{{\cal P}}
\def\Ak{{\cal A}_k}

\def\cala{{\cal A}}

\def\Gnp#1{{\Gamma^{#1}_\Phi}}

\def\ctk{\; \hbox{\bf causes to know } \;}

\def\causes{\; \hbox{\bf causes} \;}
\def\If{\; \hbox{\bf if} \;}
\def\Then{\hbox{\bf then} \;}
\def\Else{\hbox{\bf else}\;}
\def\While{\; \hbox{\bf while} \;}
\def\Do{\;\hbox{\bf do} \;}
\def\initially{\hbox{\bf initially} \;}
\def\after{\; \hbox{\bf after} \;}

\def\holdp1{ { holds(p_1,S) } }

\def\holdpn{ { holds(p_n,S) } }

\def\precon{p_1,\ldots ,p_n}

\def\sledom{=\!\!|}

\def\Not{{\bf not}\;}
\def\ol#1{\overline{#1}}

\def\holdbp1{ {\overline{holds(p_1,S)} } }

\newtheorem{proposition}{Proposition}[section]
\newtheorem{theorem}[proposition]{Theorem}
\newtheorem{corollary}[proposition]{Corollary}
\newtheorem{lemma}[proposition]{Lemma}
\newtheorem{definition}[proposition]{Definition}
\newtheorem{example}[proposition]{Example}

\def\bfAs{{\bf A}_{(\Phi,\Sigma,s)}}
\def\bfA#1{{\bf A}^{#1}_{(\Phi,\Sigma,s_0)}}

\def\Aps{A_{(\Phi_0,\sigma,s)}}
\def\Pds#1{\Pi^{#1}_{({D_\sigma},s)}}
\def\Asso#1#2#3{Asso_{#1}(#2,#3)}

\def\o#1{\bar{#1}}
\def\Find#1#2#3{find\_situation(#1,#2,#3)}
\def\Holds#1#2{holds\_after\_plan(#1,#2)}

\newcommand{\may}{\ \mbox{\bf may affect}}
\newcommand{\toknow}{\ \mbox{\bf causes to know}}
\newcommand{\After}{\ \mbox{\bf after}}

\title[Knowledge and the Action Description Language $\cala$]
{Knowledge and the Action Description Language $\cala$}

\author[Jorge Lobo, Gisela Mendez and Stuart R. Taylor]
{JORGE LOBO\thanks{Partially funded by Argonne National Laboratory 
under Contract No. 963042401. The research was partially conducted at
the EECS department of the University of Illinois at Chicago.}\\
Network Computing Research Department,
                Bell Laboratories, Murray Hill, NJ 07974, USA\\
\email{jlobo@research.bell-labs.com}
\and
GISELA MENDEZ\thanks{Work done while visiting the University of Illinois 
at Chicago and Bell Labs.}\\
Departamento de Matem\'{a}ticas,
                 Universidad Central de Venezuela, Caracas, Venezuela \\
\email{gmendez@ciens.ucv.ve}\\
\and
STUART R. TAYLOR \\
Raytheon Systems Company, 
                 Expeditionary Warfare \& Industrial Automotive,\\
                 13532 N. Central Exp., MS 37, 
                 Dallas, TX, 75243, USA\\
\email{s-taylor5@ti.com}}
\begin{document}
\maketitle
\begin{abstract}
We introduce $\Ak$, an extension of the action description language
{$\cal A\;$} \cite{gl:Alang} to handle actions which affect knowledge.
We use sensing actions to increase an agent's knowledge of the world
and non-deterministic actions to remove knowledge. We include complex
plans involving conditionals and loops in our query language for
hypothetical reasoning.  We also present a translation of $\Ak$ domain
descriptions into epistemic logic programs.\footnote{This paper
extends the results of the work first presented in \cite{LMT97}.}
\end{abstract}

%%%%%%%%%%%%%%%%%%%%%%%%%%%%%%%%%%%%%%%%%%%%%%%%%%%%%%%%%%%%%%%%%%%%
%%%%% Introduction %%%%%%%%%%%%%%%%%%%%%%%%%%%%%%%%%%%%%%%%%%%%%%%%%
%%%%%%%%%%%%%%%%%%%%%%%%%%%%%%%%%%%%%%%%%%%%%%%%%%%%%%%%%%%%%%%%%%%%

\section{Introduction} \label{sec:intro}

Since its introduction, the action description language $\cala$ has
served as a platform to study several aspects that arise when we try
to formalize theories of actions in logic \cite{gl:Alang}.  $\cala$ was
designed as a minimal core of a high level language to represent and
reason about actions and their effects.  Domain descriptions written
in this language have direct translations into extended logic
programs.  Extensions to $\cala$ have been developed to study and
reason about the concurrent execution of actions \cite{bg:ReCa}, the
non-deterministic effects of some actions \cite{th:Raelp} and to study
many instances of the qualification and ramification problems
\cite{kl:Aie}, \cite{gkl:Ram}, \cite{mct:Qual}.

In this paper we propose a new action description language called
${\cal A}$$_{k}$.  ${\cal A}$$_k$ is a minimal extension of $\cala$ to
handle {\em sensing\/} actions.  A sensing action is an action that does
not have any effect in the world.  The effect is only in the
perception of the reasoning agent about the world.  The execution of a
sensing action will increase the agent's knowledge about the current
state of the world.  Take for example a deactivated agent placed
inside a room. The agent has duties to carry out and will be activated
by a timer. Let us assume the agent is always placed facing the door.
The agent, once activated, may become damaged if it attempts to leave
the room and the door is closed.  Before the agent tries to leave
the room it needs to perform some act of sensing in order to determine
whether the door is opened or not.  The agent has incomplete knowledge
with respect to the door. A sensing action such as {\em looking at the
  door\/} would provide information to the agent concerning the status
of the door.

In our simple model there will be two sources of knowledge available
to an agent: initial knowledge, i.e., knowledge provided to
the agent at initialization time, and knowledge gained from sensing
actions.  We will assume that the agent is acting in isolation.  Thus,
once an agent has gained knowledge about its world, only its actions or
limitations of its reasoning mechanism (such as limited memory) could
make the agent lose knowledge.  We will assume an ideal agent and
expect that only actions can remove knowledge.  An action can cause
the loss of knowledge if its effect is non-deterministic.  Take for
example the action of tossing a coin.  We know it will land with
either heads showing or with tails showing, but exactly which cannot
be predicted.  Non-deterministic actions and sensing actions have
opposite effects on an agent's knowledge.

The main contributions of this paper are: 
\begin{itemize}

\item The language $\Ak$, which
incorporates sensing and non-deterministic actions.

\item A query sub-language with complex plans that allow hypothetical 
reasoning in the presence of incomplete information.  These complex plans
include conditionals (if-then-else) and routines (while-do).

\item
A sound and complete translation of domain descriptions written in
$\Ak$ into epistemic logic programs.  

\end{itemize}

The rest of this paper is organized as follows.  In Section
\ref{sec:domain}, we start with the syntax and semantics of domains
with deterministic and sensing actions only.
Section~\ref{sec:queriesI} presents the query sub-language of $\Ak$
with conditional plans.  In Section~\ref{sec:non-det}, the language is
extended to include non-deterministic actions and
Section~\ref{sec:queries} adds loops to the query language.
Section~\ref{ELP} gives an outline of epistemic logic programs as they
pertain to $\Ak$. In Section~\ref{sec:translation}, we present the
translation of domains in $\Ak$ into epistemic logic programs. 
%In section 7, we present an example of the translation. 
In Section~\ref{sec:other-work}, we discuss how our work relates to
other work in the field.  Section~\ref{sec:future} presents a few
directions for future work and concluding remarks.

%%%%%%%%%%%%%%%%%%%%%%%%%%%%%%%%%%%%%%%%%%%%%%%%%%%%%%%%%%%%%%%%%%%%
%%%%% Domain Language %%%%%%%%%%%%%%%%%%%%%%%%%%%%%%%%%%%%%%%%%%%%%%
%%%%%%%%%%%%%%%%%%%%%%%%%%%%%%%%%%%%%%%%%%%%%%%%%%%%%%%%%%%%%%%%%%%%

\section{$\Ak$: Domain Language} \label{sec:domain}

%%%% Syntax of Ak %%%%%%%
\subsection{Syntax of $\Ak$} 
The language of $\cal A$$_{k}$ consists of two non-empty disjoint sets
of symbols $F$, $A$.  They are called {\em fluents\/}, and {\em
actions\/}.  As in $\cala$, fluents are statements or observations about
the world.  The set $A$ consists of two disjoint sets of actions, {\em
sensing\/} actions and {\em non-sensing\/} actions.  Actions will be
generically denoted by $a$, possibly indexed.  A {\em fluent
literal\/} is a fluent or a fluent preceded by a $\neg$ sign.  A fluent
literal is negative when preceded by $\neg$ and is positive otherwise.
Fluent literals will be denoted by $f$, $p$ and $q$ possibly indexed.

There are three kinds of propositions in $\Ak$, {\em object effect\/}
propositions, {\em value\/} propositions and {\em non-deterministic
effect\/} propositions.  We discuss non-deterministic effect
propositions in Section~\ref{sec:non-det}.

\noindent
Object effect propositions are
expressions of the form
\begin{equation}\label{eq1}
  a {\bf \;causes}\; f {\bf \;if}\; \precon
\end {equation}
where $a$ is a non-sensing action, and $f$ and
$\precon$,  with $n\leq 0$, are fluent literals.
%$\varphi$ a propositional formula built using fluent literals.  
This expression intuitively means that in a situation where $\precon$ are
true, the execution of $a$ causes $f$ to become true. 
%If the fluent literal in (\ref{eq1}) is negative then the object 
%effect proposition becomes

When  $n = 0$ in the preconditions of (\ref{eq1})
we will write the proposition as
\begin{equation}\label{eq3}
  a {\bf \;causes}\; f
\end {equation}
%
%\begin{equation}\label{eq4}
%  a {\bf \;causes}\; \neg g
%\end {equation}

A {\em value proposition\/} is an expression of the form
\begin{equation} \label{eq7}
{\bf \;initially}\;  f
\end {equation}
where $f$ denotes a fluent literal. Value propositions describe 
the initial knowledge the agent has about the world.

%\noindent
There are also {\em knowledge\/} laws.  Knowledge laws are expressions
of the form
\begin{equation}\label{eq5}
a_s {\bf \;causes\;to\; know}\; f {\bf \;if}\; \precon
\end {equation}
where $a_s$ is a sensing action, $f$ is a fluent and $\precon$ are
preconditions as in (\ref{eq1}). Intuitively this expression says that in a
situation where $\precon$ are true the execution of $a_s$ causes the
agent to realize the current value of $f$ in the world.
%We will call the action $a$ in (\ref{eq3}) a {\em sensing} action.  
We do not allow sensing actions to occur in any effect
proposition.

If $n = 0$ in (\ref{eq5}), we will write the
knowledge law as
\begin{equation}\label{eq6}
a_s {\bf \;causes\;to\; know}\; f
\end {equation}

At this point we should remark that we are assuming the agent may have
incomplete but always correct knowledge about the world.  Propositions
and laws in $\Ak$ describe how the knowledge of the agent changes, but
if these changes are the result of propositions like (\ref{eq1}) we
assume that the effects in the world would be the same as if the world
were in a state where $\precon$ are true, that is, there are not
external entities that modify the world and the specification of the
laws are correct and deterministic.

%\medskip \noindent
\begin{definition} \label{domains-def}
A collection of the above propositions and laws is called a {\em
domain description\/}. A domain description $D$ is {\em simple\/} if
for any sensing action $a_s$ and any fluent $f$ there exists at most
one knowledge law in $D$ of type (\ref{eq5}).
\end{definition}
The following example illustrates how knowledge laws can be used
to reason about actions.

\begin{example} \label{ex:bulb1}
  A robot is instructed to replace the bulb of a halogen lamp. If the
  lamp is on when the bulb is screwed in, the robot's circuits will get
  burned out from the heat of the halogen bulb, and it will not be
  able to complete the task.  The robot will have 
  to find a sequence of actions that will allow it to complete the task 
  without burning out. We assume that
  the robot is already at the lamp. This is represented by the
  following domain description,
  
%A robot is instructed to replace the halogen bulb of lamp and it needs
%to find a sequence of actions to complete the task without burning out
%its circuits.  Assume that its world is described by the following
%domain description.
\[D_1\left\{
\begin{array}{ll}
r_1 : {\bf \;initially}\; \neg burnOut \\
r_2 : {\bf \;initially}\; \neg bulbFixed \\
r_3 : changeBulb {\bf \;causes}\; burnOut {\bf \;if}\; switchOn \\
r_4 : changeBulb {\bf \;causes}\; bulbFixed {\bf \;if}\; \neg switchOn \\
r_5 : turnSwitch {\bf \;causes}\; switchOn {\bf\;if}\; \neg switchOn \\
r_6 : turnSwitch {\bf \;causes}\; \neg switchOn {\bf\;if}\; switchOn \\
\end{array} \right. \] 
It follows from $D_1$ that in the initial state the robot does not
know the state of the switch in the lamp.
Hence, there does not exist a way to determine before hand what will be
the result of the action $changeBulb$.  
When the robot goes to carry out the action $changeBulb$, it could
end up in a resulting state in which either $bulbFixed$ is true or in
a state where it will be burned out and unable to complete the task.
Without knowing whether the switch is on or off, the robot will not be
able to find a plan to accomplish its task.  The robot must first
check the state of the switch.  After realizing whether the switch is on or
off, it will take the appropriate actions to complete the task.  The
robot will need a knowledge law such as:

$r_7$ : $checkSwitch$ ${\bf\;causes\;to\;know}$ $switchOn$ ${\bf
  \;if}$ $\neg burnOut$

\noindent
After checking the switch the robot will {\em know\/} whether the switch
is on or off.  Sensing gives the robot that extra knowledge it would
need to accomplish the task without burning out and provides a
branching point in its hypothetical reasoning. If the switch is on it
will turn the switch and replace the bulb.  If the switch is off it
will directly replace the bulb.  This conditional reasoning will
enable the robot to show that there is a sequence of actions to
accomplish the task.
\end{example}
%

%%%% Semantics of Ak %%%%%%
\subsection{Semantics of $\Ak$}

The semantics of $\Ak$ must describe how an agent's knowledge changes
according to the effects of actions defined by a domain description.
We begin by presenting the structure of an agent's knowledge.  We will
represent the knowledge of an agent by a set of possibly incomplete
worlds in which the agent believes it can be.  We call these worlds
{\em situations\/} and a collection of worlds an {\em epistemic state\/}.
A situation, since it could be an incomplete description of the world,
will be represented by a collection of sets of fluents.  A set of
fluents will be called a {\em state\/}.  If a formula is true in an
epistemic state of an agent (to be defined later), by our assumption
it means that the agent knows that the formula is true in the real
world.  Epistemic states will also allow us to distinguish when the
agent knows that the disjunction $f_1 \vee f_2$ is true from when it
either knows $f_1$ or knows $f_2$.\footnote{Note the similarity with a
collection of belief sets in \cite{gp:Epi}.}

We will say that a fluent $f$ is true or holds in a state $\sigma$
(denoted by $\sigma \models f$) iff $f\in\sigma$.  A fluent $f$ does
not hold in a state $\sigma$ (denoted by $\sigma \not\models f$) iff
$f \not \in \sigma$. $\sigma\models \neg f$ iff $\sigma \not\models
f$.  For more complex formulas, their truth value can be recursively
defined as usual.  A formula $\varphi$ made of fluents is true in (or
modeled by) a situation $\Sigma$ (denoted by $\Sigma\models
\varphi$) if the formula is true in every state in $\Sigma$; it is
false if $\neg\varphi$ is true in every state $\Sigma$.  A formula
is true in an epistemic state
if is true in every situation in the epistemic state; it is
false if its negation is true.

A situation is {\em consistent\/} if it is non-empty; otherwise it is
{\em inconsistent\/}.  A situation is {\em complete\/} if it contains a
single state; otherwise it is {\em incomplete\/}. An epistemic state is
inconsistent if it is empty or contains an inconsistent situation;
otherwise it is consistent.  An epistemic state is complete if it
contains only one complete situation.  Figure 1 shows two consistent
epistemic states in which the fact ``{\em Ollie is wet\/}'' (represented
by $wet$) is known by Agent A and Agent B.  In the epistemic state
(a), containing an incomplete situation, Agent A does not have
knowledge about the weather.  In the other epistemic state (b),
containing two complete situations, Agent B either knows it is raining
or knows that it is not raining outside.  Recall that epistemic states
will be used in the context of plans for hypothetical reasoning. That
is, predicting properties if the plan were executed.  Thus, if an
agent plans to execute a series of actions that takes it to the
epistemic state (a), it will not know how to dress if it needs to go
outside and does not want to get wet.  In the epistemic state (b), the
agent will know how to proceed.

\begin{figure} \label{fig:toknow}
\epsfxsize=5.0in
\centerline{\epsffile{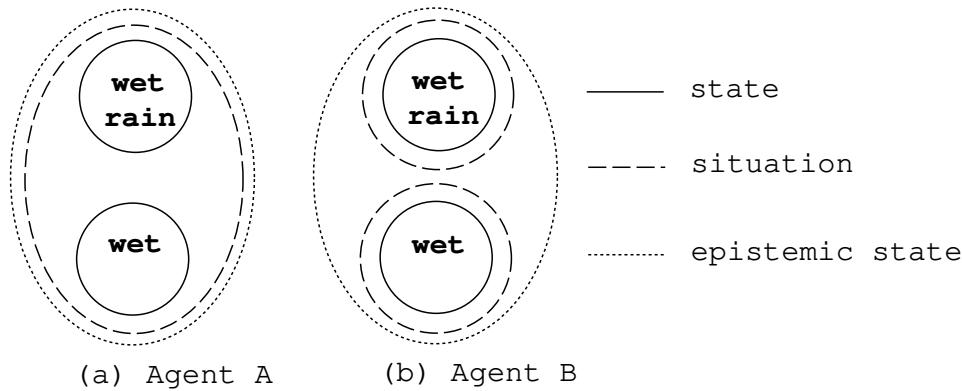}}
\caption{Epistemic states for Agent A and Agent B.}  
\end{figure}

{\em Interpretations\/} for $\Ak$ are transition functions that map pairs of 
actions and situations into situations.  To define when an
interpretation models a domain description, we will define an auxiliary
function that interprets the effect of actions at the state level.  We
call this function a {\em 0-interpretation\/}. 
{\em 0-interpretations\/} are functions that map actions and
states into  states\footnote{0-interpretations and 0-models are similar to
interpretations and models for domains in $\cala$.}. 
A 0-interpretation $\Phi$$_0$ is a {\em 0-model\/} of a domain
description $D$ iff for every state $\sigma$

\begin{enumerate}
\item For a fluent $f$ of any effect proposition of the form ``$a$
  {\bf causes} $f$ {\bf if} $\precon$" in $D$, the fluent $f$ holds in
  $\Phi_{0} (a,\sigma)$ if its preconditions $\precon$ holds in $\sigma$,

\item For a fluent literal $\neg f$ of any effect proposition of the
  form ``$a$ {\bf causes} $\neg f$ {\bf if} $\precon$" in $D$, the
  fluent $f$ does not hold in $\Phi_{0} (a,\sigma)$ if its preconditions 
  $\precon$ holds in $\sigma$,

\item For a fluent $f$, if there are no effect propositions of
  the above types, then $f \in \Phi_{0} (a,\sigma)$ if and only if $f
  \in \sigma$.
\end{enumerate}

\medskip \noindent Before we define when an interpretation $\Phi$ is a
{\em model\/} of a domain description $D$, we need the following
definition that will let us interpret knowledge laws.  The interest of
the defintion will become clear after we explore the scenarion in
Example \ref{ex:door}.

\begin{definition}\label{def:comp}
  Let $\Sigma$ be a consistent situation, $f$ a fluent and $\varphi$ a
  disjunction of conjunctions of fluent literals (preconditions).  A
  consistent situation $\Sigma'$ is ``$f,\varphi$-{\em compatible\/}"
  with $\Sigma$ iff $\Sigma' = \Sigma$ whenever $f$ is either true or
  false in $\Sigma$.  Otherwise $\Sigma'$ must satisfy one of the
  following conditions:
\begin{enumerate}
\item $\Sigma'  = \{\sigma \in \Sigma \;|\; \varphi$ is not
  true in $\sigma \}$

\item $\Sigma' = \{\sigma \in \Sigma \;|\; \varphi$ is true in
  $\sigma,\; f \not \in \sigma\}$

\item $\Sigma' = \{\sigma \in \Sigma \;|\; \varphi$ is true in
  $\sigma,\; f \in \sigma\}$

\end{enumerate}
\end{definition}

\begin{example} \label{ex:door}
  Let us return to the agent scenario from the introduction.  Imagine
  that currently the agent is deactivated in the room.  The agent will
  be automatically activated by an internal clock.  Then it needs to
  find the door, leave the room, and perform some duties.  When the
  agent is initially activated it will know nothing about its
  surroundings and will remain ignorant of its surroundings until it
  performs a sensing action. We will show how the conditions presented
  in Definition \ref{def:comp} are enough to represent the result of
  sensing.  Its only action is to $look$. We assume the action consist
  of opening its ``eyes" and looking.
%The knowledge law will be
 This domain is represented below with only one knowledge law,
%
%\[
%\begin{array}{ll}
% look {\ctk}  doorOpened {\bf\;if}\; facingDoor
%\end{array}  \] 
%
\[D_2\left\{
\begin{array}{ll}
r_1 : look {\ctk}  doorOpened {\bf\;if}\; facingDoor
\end{array} \right. \] 
\noindent This initial situation of complete ignorance is
represented by the situation 

\{\{\}, \{$doorOpened$\}, \{$facingDoor$\}, \{$doorOpened,facingDoor$\}\}.  

\noindent
If the action $look$ is executed in the real world the agent may find that
it is not facing the door and will not know whether the door is opened 
or not, this is represented by the situation

\{ \{ \}, \{$doorOpened$\} \}.

\noindent
Another possibility could be that the agent was facing the door and after
it is activated, it will know that it is facing the door and will also know 
that the door is not opened

\{ \{$facingDoor$\} \}.

\noindent Still another possibility could be that the agent
was facing the door and after being activated, it will learn that it
is facing the door and that the door is opened

\{ \{$doorOpened,\;facingDoor$\} \}.

\noindent Since the agent will be doing hypothetical
reasoning (i.e. planning) it will have no way of knowing  which
situation it will be in until the action is actually executed. Thus,
the agent can only assume that it will be in one of the three
situations, so when the agent analyzes what would be the consequences of
executing $look$ it concludes that the result will take it to the
epistemic state that consists of the following three situations.

\begin{enumerate}

\item \{ \{ \}, \{$doorOpened$\} \}

\item \{ \{$facingDoor$\} \}

\item \{ \{$doorOpened,\;facingDoor$\} \}

\end{enumerate}

\noindent Each situation is $doorOpened,facingDoor-compatible$. The
first situation corresponds to the first case of Definition
\ref{def:comp}. The agent knows it is not facing the door since
$facingDoor$ is false in all states contained in the situation.  The
same cannot be said for $doorOpened$ since in one state it is false
and the other state it is true. This is to be expected since in this
situation the agent is not facing the door,
and it cannot know if the door is opened or closed.

\medskip \noindent
The second situation corresponds to the second case of Definition
\ref{def:comp}. This situation contains all the states in which the 
precondition $facingDoor$ is true and the fluent $doorOpen$ is false. 
The agent not only knows it is facing the door but also knows the door
is not opened. 

\medskip \noindent
The last situation is from the last case of Definition \ref{def:comp}.
In this situation the agent knows it is facing the door and also knows
that the door is opened. 

\medskip \noindent
Observe that a result of sensing is that the preconditions of the sensing 
action will become known to the agent if the value of the fluent being
sensed is initially unknown.  This occurs even if the effect of the action 
remains unknown after executing the action, which hapens in the situation
coming from the states where  
the preconditions of the execution of the sensing action in a knowledge 
laws are not true. 
\end{example}

%%%%%%%%%%%%%%%%%%%%%%%%%%%%%%%%%%%%%%%%%%%%%%%%%%%%%%%%%%%%%%%%%%%%

\begin{definition}

  A state $\sigma$ is called an {\em initial state\/} of a domain
  description $D$ iff for every value proposition of the form ``{\bf
    initially}$\; \varphi$'' in $D$, $\varphi$ is true in $\sigma$.
  The {\em initial situation\/} $\Sigma_0$ of $D$ is the set of all
  the initial states of $D$.
\end{definition}

%\msg{ Should we include initial situations in this def? }

\begin{definition}
A fluent $f$ is a {\it potential sensing effect} of a sensing action
$a_s$ in a domain $D$ if there is a knowledge law of the form

$\;\;\;\;\;\;\;\;\;\;\;$
$ a_s {\bf \;causes\; to\; know\;} f \;{\bf if} \; \varphi$ 

\noindent
in $D$.  We will also say that $f$ is the {\it potential sensing
effect} of the knowledge law.

The {\it knowledge precondition} of a fluent $f$ with respect to a sensing
action $a_s$ in a domain $D$ is the disjunction $\varphi_1
\vee\ldots\vee \varphi_n$ if and only if 

$\;\;\;\;\;\;\;\;\;\;\;$
$ a_s {\bf \;causes\; to\; know\;} f \;{\bf if} \; \varphi_1$ 
 
$\;\;\;\;\;\;\;\;\;\;\;\;\;\;\;\;\;\;\;\;\;\;\;\;\;\;\;\;\;\;\;$
$\vdots$

$\;\;\;\;\;\;\;\;\;\;\;$
$ a_s {\bf \;causes\; to\; know\;} f \;{\bf if} \; \varphi_n$ 

\noindent
are all the knowledge laws in which $a_s$ occurs and $f$ is a potential
sensing effect.
\end{definition}       

Note that if the domain is simple (Definition \ref{domains-def}) then
the knowledge precondition of any fluent in the domain with respect to
any sensing actions is either empty or it has only one disjoint.

\begin{definition}
\label{d-model}
  Given an interpretation $\Phi$ of $\Ak$,
  $\Phi$ is a {\em model\/} of a domain description $D$, if and only if
  for any consistent situation $\Sigma$:
\begin{enumerate}

\item\label{0-model} There exists a 0-model $\Phi_0$ of $D$, such that
  for any non-sensing action $a$, 

  $\Phi(a,\Sigma)$ = \( \displaystyle \bigcup_{\sigma\in\Sigma} 
  \{\Phi_0(a,\sigma)\}\).

\item For each sensing action $a_s$, let $f_1,\ldots,f_n$ be the
potential sensing effects of $a_s$ and $\varphi_i$ the knowledge
precondition of $f_i$ with respect to $a_s$.
%
%$\;\;\;\;\;\;\;\;\;\;\;$
%$ a_s {\bf \;causes\; to\; know\;} f_1 \;{\bf if} \; \varphi_1$ 
% 
%$\;\;\;\;\;\;\;\;\;\;\;\;\;\;\;\;\;\;\;\;\;\;\;\;\;\;\;\;\;\;\;$
%$\vdots$
%
%$\;\;\;\;\;\;\;\;\;\;\;$
%$ a_s {\bf \;causes\; to\; know\;} f_n \;{\bf if} \; \varphi_n$ 
%
%be all the knowledge laws in which $a_s$ occurs. 
Then, $\Phi(a_s,\Sigma)$ must be consistent and if $n = 0$,
$\Phi(a_s,\Sigma) = \Sigma$ otherwise $\Phi(a_s,\Sigma) =
\bigcap_{i\in[1..n]} \Sigma_i$, such that each $\Sigma_i$ is a
situation $f_i,\varphi_i - compatible$ with $\Sigma$.

\end{enumerate}

\noindent
$\Phi(a,\Sigma) = \emptyset$ for any action ${a}$ if 
$\Sigma = \emptyset$.
\end{definition}

\begin{example} \label{ex:room1}
The third floor agent of a building has the job of making sure the
white-board in a room on that floor is clean.  The agent will approach
the room, look into the room, clean the white-board if it is not
clean, and then leave the room.  We focus here on ``looking into the
room".  When the agent looks into the room it will know whether the
white-board in that room is clean.  Also if the curtains are open the
agent will learn whether it is raining outside.  Sensing actions can not 
appear in object effect propositions, but there is no
restriction on the number of knowledge laws associated with a sensing
action.  Thus, the action could affect the truth value of several
fluents simultaneously.  In this example the sensing action
$lookInRoom$ will appear in two knowledge laws.  We will see how the
resulting situations are $f,\varphi$-compatible with the initial
situation and briefly discuss the models of this domain description.
%When the whiteboard agent looks into a room it will know whether the
%Assume that when an agent assigned to clean white-boards looks into a
%room (denoted by lookInRoom) it will know whether the white-board in that
%room is clean(denoted by boardClean).  It will also learn whether it is
%raining outside(denoted by rainOutside) if the curtains in the room are
%open(denoted by curtainOpen).  We assume that because of building policy, the
%agent will initially know that the lights of the room will be
%on(denoted by lightOn) and that the curtains of the room will be open.  A
%simple domain description for this situation will be:
%
The following simple domain description illustrates the scenario,
\[D_3\left\{
\begin{array}{ll}
r_1 : {\bf \;initially}\; curtainOpen \\
r_2 : {\bf \;initially}\; lightOn \\
r_3 : lookInRoom {\ctk} rainOutside {\bf\;if}\; curtainOpen \\
r_4 : lookInRoom {\ctk} boardClean {\bf\;if}\; lightOn \\
\end{array} \right. \] 

\noindent The initial situation $\Sigma_0$ of $D_3$
has four states.\footnote{Observe that the initial epistemic state of
the robot has always a single situation.  To be able to specify more
complex initial epistemic states the language must be changed.}
\[
\begin{array}{ll}
\Sigma_0 = \{ & \{curtainOpen,lightOn\},\\
&
\{rainOutside,curtainOpen,lightOn\},\\
&
\{boardClean,curtainOpen,lightOn\}, \\
&
\{rainOutside,boardClean,curtainOpen,lightOn\}\}
\end{array}
\]
%
%\[
%\begin{array}{l}

%\noindent
%$\Sigma_0=\{\{curtainOpen,lightOn\},\{rainOutside,curtainOpen,lightOn\},$
%%\Sigma_0=\{\{curtainOpen,lightOn\},\{rainOutside,curtainOpen,lightOn\},\\

%$\{boardClean,curtainOpen,lightOn\},\{rainOutside,boardClean,curtainOpen,lightOn
%\}\}$
%%\end{array}
%%\]
%%

\noindent
There is only one action in $D_3$, and any model of $D_3$ applied to
the initial situation $\Sigma_0$ may behave in one of the following forms:
% $\Phi (lookInRoom,\Sigma_0)$.

\medskip
$\Phi_1(lookInRoom,\Sigma_0)=\{\{curtainOpen,lightOn\}\}$

$\Phi_2(lookInRoom,\Sigma_0)=\{\{rainOutside,curtainOpen,lightOn\}\}$

$\Phi_3(lookInRoom,\Sigma_0)=\{\{boardClean,curtainOpen,lightOn\}\}$

$\Phi_4(lookInRoom,\Sigma_0)=\{\{rainOutside,boardClean,curtainOpen,lightOn\}\}$

\medskip \noindent Models may differ in how they behave when they are
applied to other situations different to $\Sigma_0$, but for
$\Sigma_0$ they must be equal to one of the $\Phi_i$ above.  Unlike
domains in ${\cal A}$ in which given an initial situation there is
only one model for the domain, our language allows for several models.

Observe too that since {\em lookInRoom\/} is a sensing action, its
occurrence does not change any fluent's value. If we start from
%$\Sigma_0$, after we reach one the four situations, any new execution
$\Sigma_0$, and then reach one of the four situations, any new execution
of {\em lookInRoom\/} will result in the same situation.

To verify that each of the $\Phi_i$ can be a partial description of a model
of $r_3$ and $r_4$, let
\[
\begin{array}{ll}
\Sigma_1=\{ &
\{curtainOpen,lightOn\},\{boardClean,curtainOpen,lightOn\}\} \\
\Sigma_2 = \{ &
\{rainOutside,curtainOpen,lightOn\}, \\
&\{rainOutside,boardClean,curtainOpen,lightOn\}\} \\
\Sigma_3=\{ &
\{curtainOpen,lightOn\},\{rainOutside,curtainOpen,lightOn\}\} \\
\Sigma_4 = \{ &
\{boardClean,curtainOpen,lightOn\}, \\
& \{rainOutside,boardClean,curtainOpen,lightOn\}\}
\end{array}
\]
Note that $\Sigma_1$ and $\Sigma_2$ are
$rainOutside,curtainOpen$-compatible with $\Sigma_0$, and that
$\Sigma_3$ and $\Sigma_4$ 
are $boardClean,lightOn$-compatible with $\Sigma_0$, and
\[\begin{array}{l}
\Phi_1(lookInRoom,\Sigma_0) = \Sigma_1\cap\Sigma_3\\
\Phi_2(lookInRoom,\Sigma_0) = \Sigma_2\cap\Sigma_3 \\
\Phi_3(lookInRoom,\Sigma_0) = \Sigma_1\cap\Sigma_4\\
\Phi_4(lookInRoom,\Sigma_0) = \Sigma_2\cap\Sigma_4
\end{array}
\]

\noindent Note also that none of the situations are
$f,\varphi$-compatible with $\Sigma_0$ by part (1) of Definition
\ref{def:comp} because there is no knowledge precondition $\varphi$ of
either $rainOutside$ or $boardClean$ with respect to $lookInRoom$ in
the domain description $D_3$
%\footnote{aqui tu tenias $D_3$}
that is false in any of the states in the initial situation
$\Sigma_0$.
\end{example}

%%%%%%%%%%%%%%%%%%%%%%%%%%%%%%%%%%%%%%%%%%%%%%%%%%%%%%%%%%%%%%%%%%%%
%%%%% Query Language %%%%%%%%%%%%%%%%%%%%%%%%%%%%%%%%%%%%%%%%%%%%%%%
%%%%%%%%%%%%%%%%%%%%%%%%%%%%%%%%%%%%%%%%%%%%%%%%%%%%%%%%%%%%%%%%%%%%

\section{$\Ak$: Query language--part I} \label{sec:queriesI}
%As one may have noticed, value propositions of the type (\ref{hypact}) 
%is missing from our domain language. 
%are not part of our domain language. 
%There is no separation of domain and query language in $\cala$. This 
%presents a problem with the value proposition (\ref{hypact}). This 
%value proposition may have several meanings. One is ``The sequence of
%actions $A_1,\ldots,A_m$ is executed in the initial situation making $F$
%true (false)." Another way is, ``If the sequence $A_1,\ldots,A_m$ could be
%executed in the initial situation then $F$ would be true afterwards."
%The first corresponds to an actual observation and the second, to a 
%hypothetical observation. A consequence of this semantic ambiguity is 
%that there is no clear division between actual and hypothetical 
%observations. To avoid this, we adopt a solution given in 
%\cite{lif:TwoComp}. We separate the language of $\cala$ into
%a query language and a domain language. We then go even further
%and restrict (\ref{hypact}) to the query language and 
%give propositions like (\ref{hypact}) the name queries.
%
%\subsection{Plans}

Given a domain description, an agent would like to ask how the
world would be after the execution of a sequence of actions 
starting from the initial situation. 
Using actions as in $\cala$, queries in $\Ak$ can be of the form
\begin{equation}\label{eq8}
\varphi \;{\bf after \;} [a_1,\ldots ,a_n]
\end{equation}
where $\varphi$ is a conjunction of fluent literals.  The answer to
this query will be {\em yes\/} (or true) in a domain $D$ if for every
model $\Phi$ of $D$ the test condition $\varphi$ is true in the
situation \[\Phi(a_n,\Phi(a_{n-1},\ldots\Phi(a_1,\Sigma_0)\cdots))\]
i.e. the situation that results after the execution of
$a_1,\ldots,a_n$ from the initial situation $\Sigma_0$ of $D$.  The
answer will be {\em no\/} (or false) if for every model $\Phi$ of $D$
$\varphi$ is false in
$\Phi(a_n,\Phi(a_{n-1},\ldots\Phi(a_1,\Sigma_0)\cdots))$.  Otherwise
the answer will be $unknown$.  With this notion we can define an
entailment relation between domain descriptions and queries.  We say
that a domain $D$ entails a query $Q$, denoted by $D\models Q$, if the
answer for $Q$ in $D$ is yes.  For example, if we add $\initially
switchOn$ to $D_1$, it can be easily shown that
\[D_1 \models bulbFixed \after [turnSwitch,changeBulb]\] 
However, from the original $D_1$ (even including $r_7$) there does not
exist a sequence of actions $\alpha$ such that $D_1 \models bulbFixed
\after \alpha$.  The inferences from $D_1$ are conditioned to the
output of the sensing action: if the switch is on then the sequence
$[turnSwitch,changeBulb]$ will cause the light to be fixed, else the
single action $[changeBulb]$ will fix it.  Reasoning in the presence
of sensing actions requires the projections to be over plans more
complex than a simple sequence of actions.

We recursively define a {\em plan\/} as follows,\footnote{We will use
the list notation of Prolog to denote sequences.}
\begin{enumerate}

\item an empty sequence denoted by $[]$ is a plan.
\item If $a$ is an action and $\alpha$  is a plan then the
concatenation of $a$ with $\alpha$ denoted by $[a|\alpha]$ is also a plan.

\item If $\varphi$ is a conjunction of fluent literals and $\alpha$,
  $\alpha_1$ and $\alpha_2$ are plans then $[\If \varphi \Then
  \alpha_1|\alpha]$ and $[\If \varphi \Then \alpha_1 \Else
  \alpha_2|\alpha]$ are (conditional) plans.

\item Nothing else is a plan.

\end{enumerate}
Now we redefine a query to be a sentence of the form
\begin{equation}\label{eq8.1}
\varphi \;{\bf after \;} \alpha
\end{equation}
Where $\varphi$ is a test condition (a conjunction of fluent literals) 
and $\alpha$ is a plan.

\begin{example} (Conditionals) \label{ex:bulb2}
Here we add the knowledge law to $D_1$ and rename it $D_{1'}$. 

\[D_{1'}\left\{
\begin{array}{ll}
r_1 : {\bf \;initially}\; \neg burnOut \\
r_2 : {\bf \;initially}\; \neg bulbFixed \\
r_3 : changeBulb {\bf \;causes}\; burnOut {\bf \;if}\; switchOn \\
r_4 : changeBulb {\bf \;causes}\; bulbFixed {\bf \;if}\; \neg switchOn \\
r_5 : turnSwitch {\bf \;causes}\; switchOn {\bf\;if}\; \neg switchOn \\
r_6 : turnSwitch {\bf \;causes}\; \neg switchOn {\bf\;if}\; switchOn \\
r_7 : checkSwitch {\bf\;causes\;to\;know}\;switchOn{\bf\;if}\;\neg burnOut \\
\end{array} \right. \] 
We can define a conditional plan to fix the bulb:
\[\begin{array}{lll}
bulbFixed \after & [checkSwitch, \\
                  & \If \neg switchOn &\Then [changeBulb]\\
                  &                      &\Else
                                        [turnSwitch,changeBulb]]. 
\end{array}
\]
\end{example}

\medskip \noindent
The above query provides two alternatives for reasoning. The ${\bf else}$ 
clause is followed if the test condition is false. 
A conditional can be expanded to a ${\bf case}$ statement in general 
when reasoning needs to be done along several different sequences of plans. 
Note that if the conditional plan was attempted before or without the 
sensing action $checkSwitch$, the query may not succeed because the 
test condition could evaluate to neither true nor false but rather unknown. 
Sensing actions need to be executed before the conditionals to 
ensure the test conditions will evaluate to either true or false.

\subsection{Plan Evaluation Function and Query Entailment}

To formally define entailment we need to define first the evaluation
of a plan in terms of interpretations.  In other words, we define how the
plan will change an initial situation based on an interpretation.

\begin{definition}\label{exe-func} The plan evaluation function
  $\Gamma_\Phi$ of an interpretation $\Phi$ is a function such that
  for any situation $\Sigma$
\begin{enumerate}

\item $\Gamma_\Phi([],\Sigma) = \Sigma$.

\item $\Gamma_\Phi([a|\alpha],\Sigma) =
\Gamma_\Phi(\alpha,\Phi(a,\Sigma))$ for any action $a$.

\item $\Gamma_\Phi([\If \varphi \Then \alpha_1|\alpha],\Sigma) =
\Gamma_\Phi(\alpha,\Sigma')$, where $$\Sigma' =
\left\{ 
\begin{array}{ll}
  \Gamma_\Phi(\alpha_1,\Sigma)&\mbox{if $\varphi$ is true in $\Sigma$}\\
    \Sigma & \mbox{if $\varphi$ is false in $\Sigma$}\\
 \emptyset & \mbox{otherwise} 
\end{array}
\right.$$

\item $\Gamma_\Phi([\If \varphi \Then \alpha_1 \Else
\alpha_2|\alpha],\Sigma)
 = \Gamma_\Phi(\alpha,\Sigma')$, where $$\Sigma' = 
 \left\{
\begin{array}{ll}
  \Gamma_\Phi(\alpha_1,\Sigma)& \mbox{if $\varphi$ is true in $\Sigma$}\\ 
  \Gamma_\Phi(\alpha_2,\Sigma)&\mbox{if $\varphi$ is false in $\Sigma$}\\ 
  \emptyset & \mbox{ otherwise}
\end{array}
\right.$$
\end{enumerate}
\end{definition}

\begin{definition}
A query $\varphi \after \alpha$ is entailed by a domain description
$D$ ($D \models \varphi \after \alpha$) iff for every model
$\Phi$ of $D$, $\varphi$ is true in
$\Gamma_\Phi(\alpha,\Sigma)$.
\end{definition}
It is easy to check that 
\[\begin{array}{lll}
D_{1'} \models bulbFixed \after & [checkSwitch, \\
                  & \If \neg switchOn &\Then [changeBulb]\\
                  &                      &\Else
                                        [turnSwitch,changeBulb]]. 
\end{array}
\]
It is easy to see the task will be completed regardless
of what model we are in. This is due in part to the combination of the
sensing action and the conditional plan.

\section{Actions with non-deterministic effects}\label{sec:non-det}
There are several different reasons why knowledge may be removed from
the set of facts known by the agent. There may be decay of the
knowledge, difficulty accessing the knowledge, or it may execute an 
action that makes
a particular knowledge no longer valid. In our description we assume
an ideal agent; an agent whose knowledge persists and is not subject to
any type of failure or obstacles preventing the quick access of its
knowledge. Given this assumption, the first two possibilities for the
removal of knowledge are impossible.  However,
 {\em non-deterministic actions\/} may remove
knowledge. A non-deterministic action is an action in which the
outcome cannot be predicted beforehand. An example of such an action
with an unpredictable outcome is the toss of a coin. A coin on a table
will show either heads or tails. Looking at the coin, one can gain
knowledge of which side of the coin shows. Once the action of tossing
the coin takes place we are no longer certain of which side will
show. The coin will land and will show either heads or tails. We will
not know which side shows until we do the sensing action of
looking. 
We describe the removal of knowledge as no longer knowing the truth 
value of a fluent.

%\medskip \noindent
A non-deterministic effect proposition is an expression of the form
\begin{equation}\label{eq10}
a \;{\bf may \;affect\;}\; f {\bf \;if}\; \precon
\end {equation}
where $a$ is a non-sensing action and $f$ is a fluent. 
The preconditions $\precon$ are defined as in equation (1).
Intuitively the proposition states that the truth value of $f$ may
change if $a$ is executed in a situation where $\precon$ is true.
%Out of the two possible outcomes one must be the true outcome of the
%non-deterministic action.

%\medskip \noindent
When $n = 0$, equation (10) becomes
\begin{equation}\label{eq11}
a \;{\bf may \;affect\;}\; f 
\end {equation}
We now re-define 0-interpretations to take into account
non-deterministic actions.  A 0-interpretation $\Phi$$_0$ is a {\em
0-model\/} of a domain description $D$ iff for every state $\sigma$,
$\Phi_0(a,\sigma)$ is such that

\begin{enumerate}
\item For a fluent $f$ of any effect proposition of the form ``$a$
  {\bf causes} $f$ {\bf if} $\precon$" in $D$, $f\in\Phi_0(a,\sigma)$
  if $\precon$ holds in $\sigma$,

\item For a fluent literal $\neg f$ of any effect proposition of the
  form ``$a$ {\bf causes} $\neg f$ {\bf if} $\precon$" in $D$, the
  $f\not\in\Phi_0(a,\sigma)$ if $\precon$ holds in $\sigma$,

\item For a fluent $f$ such that there are no effect propositions of
the above types, $f \in\Phi_0(a,\sigma)$ if and only if $f \in \sigma$ 
unless there is a non-deterministic effect proposition of the
form ``$a \;{\bf may \;affect\;}\; f\If \precon$" for which $\precon$
holds in $\sigma$.
\end{enumerate}
%%%%%%%%%%%%%%%%%%%%%%%%%%%%%%%%%%%%%%%%%%%%%%%%%%%%%%%%%%%%%%%%%%%

\begin{example}
  Our agent is ordered at this time to put ice from a bag into cups.
  The ice in the bag is solid. The agent needs to break the ice into
  pieces that are able to fit in the cups. The agent decides to drop
  %the bag of ice in hope that the action will produce smaller pieces
  the bag of ice as a means to complete the task.
  %Our agent is ordered this time to put ice into cups.
  %The ice cannot fit in the cups. The agent decides to drop
  %the ice to produce smaller pieces of ice that will fit.
\[D_5\left\{
\begin{array}{ll}
t1 : {\bf initially}\; inHandIceBag \\
t2 : {\bf initially}\; solidIce \\
t3 : {\bf initially}\; noDrops \\
t4 : pickUp {\bf \;causes}\; inHandIceBag {\bf \;if}\; \neg inHandIceBag \\
t5 : drop {\bf \;causes}\; \neg inHandIceBag {\bf \;if}\; inHandIceBag \\
t6 : drop {\bf \;may\;affect}\; solidIce {\bf \;if}\; noDrops \\
t7 : drop {\bf \;may\;affect}\; solidIce {\bf \;if}\; fewDrops \\
t8 : drop {\bf \;causes}\; fewDrops {\bf \;if}\; noDrops \\
t9 : drop {\bf \;causes}\; enoughDrops {\bf \;if}\; fewDrops \\
t10 : drop {\bf \;causes}\; \neg solidIce {\bf \;if}\; enoughDrops \\
t11 : checkIce {\bf \;causes\; to\; know}\; solidIce \\
t12 : putIceInCups {\bf \;causes}\; iceInCups {\bf \;if}\; \neg solidIce \\
\end{array} \right. \] 

\noindent
This example combines many of the ideas previously presented. Let us examine 
this domain description to see how this all fits together.

\begin{itemize}
\item
Rules t1 - t3 establish what is initially known in the world. 
The values of all other fluents are unknown at this time.

\item
Rules t4 and t5 describe the effect that theactions $Drop$ and
$pickUp$ have on $inHandIceBag$.

\item
Rules t6 and t7 describe the non-deterministic effect of the
action $drop$ on the ice.  

\item
Rules t8 - t10 are object effect propositions which ensure that the
ice will break after no more than three $drops$ (i.e. the execution of
the action $drop$ three times).  In the example $noDrops$ is equated
with 0 drops, $fewDrops$ with 1 drop, and $enoughDrops$ with 2 drops.
%\footnote{este 3 lo coloque yo}.

\item
Rule t11 is the sensing action which allows the agent to know whether the ice 
is broken or not after the execution of the non-deterministic action $drop$. 
Rule t12 is the goal of the task the agent is to perform. 
\end{itemize}

\noindent 
The non-determinism appears in the action of dropping the bag of
ice. Before the action is carried out, the agent knows that the ice is
solid. After the non-deterministic action, the agent is no longer
certain if the ice is still solid or in pieces.  The knowledge of
knowing the ice is solid has been removed.  The agent can only regain
that knowledge by performing a sensing action.

If the robot wants to fill the cup with ice it will iterate the
process of dropping the ice until it breaks.  A plan to accomplish this 
goal will look like:
\[
\While \neg solidIce \Do [drop,pickup,checkIce], putIceInCups]
\]
Adding loops to plans is the topic of the next section.
%\hfill $\Box$
\end{example}

\section{$\Ak$: Query language--part II} \label{sec:queries}

If we allow while-loops in our plan we could verify that $D_5$
%\footnote{aqui tenias $D_3$} 
entails the following 
query,

$iceInCups$ ${\bf after }$ 
[${\bf while}$ $\neg solidIce$ ${\bf do}$
$[drop,pickUp,checkIce]$,
$putIceInCups$]

\medskip
Similar to conditional plans, a sensing action is placed before
checking the exit condition of the loop.
We extend the definition of plans to include
loops as follows.

\begin{enumerate}

\item An empty sequence denoted by $[]$ is a plan.
\item If $a$ is an action and $\alpha$  is a plan then the
concatenation of $a$ with $\alpha$ denoted by $[a|\alpha]$ is also a plan.

\item If $\varphi$ is a conjunction of fluent literals and $\alpha$,
  $\alpha_1$ and $\alpha_2$ are plans then $[\If \varphi\; \Then
  \alpha_1|\alpha]$ and $[\If \varphi\; \Then \alpha_1 \Else
  \alpha_2|\alpha]$ are (conditional) plans.

\item If $\varphi$ is a conjunction of fluent literals and $\alpha$
and $\alpha_1$ are plans then \\
$[\While \varphi \Do \alpha_1|\alpha]$ is also a (routine) plan.

\item Nothing else is a plan.

\end{enumerate}

\subsection{Plan Evaluation Function and Query Entailment}

To extend the definition of entailment to plans with while loops we
need to extend the definition of the plan evaluation function
$\Gamma_\Phi$.  We will define this function using very elementary
tools from denotational semantics for programming languages (as in
Chapter 4 of \cite{DP90}).  The intuitive idea of the denotational
semantics is to associate the execution of a plan (or a program) of
the form ``$\While
\varphi \Do
\alpha$'' with one of the while-free plans:\footnote{Recall that the
situation $\emptyset$ represents inconsistency.}

$\;\;\;\;\;\;\;\;\;\;\;$
$\If \varphi\; \Then \emptyset$ 
 
$\;\;\;\;\;\;\;\;\;\;\;$
$\If \varphi\; \Then [\alpha, \If \varphi\; \Then \emptyset]$ 
 
$\;\;\;\;\;\;\;\;\;\;\;$
$\If \varphi\; \Then [\alpha,\If \varphi\; \Then [\alpha, \If \varphi\; \Then \emptyset]]$
 
$\;\;\;\;\;\;\;\;\;\;\;\;\;\;\;\;\;\;\;\;\;\;\;\;\;\;\;\;\;\;\;$
$\vdots$

\medskip\noindent
If the while-plan terminates then there exists a $n$ such that the
$n$th plan in this infinite sequence computes exactly the same
function that the while-plan computes.  Moreover, for each $m<n$, the
$m$th plan is an approximation of the computation of the while-plan.
If the while-plan does not terminate, any plan in the sequence is an
approximation of the while-plan but none is equivalent since the
while-plan computation is infinite.  Thus, to define this sequence we
start by defining a partial order over the set of functions that map
situations into situations.  The order will arrange the functions as in
the sequence of plans above.
\begin{definition}
  Let $\E$ be the set of all situations and $\P$ the set of all total
  functions $f$  mapping situations into situations, ${\P} = \{f \;|\;
  f:\E \rightarrow \E$ \}.  We say that for any pair of functions
  $f_1,f_2 \in \P, f_1 \leq f_2$ if and only if for any $\Sigma \in
  \E$ if $f_1(\Sigma) \neq \emptyset$, then $f_1(\Sigma) =
  f_2(\Sigma)$.
\end{definition}   
Then, we associate a (continuous) transformation inside this order to
each plan $\alpha$.  Informally speaking, the transformation starts
with the first plan in the sequence and in each application returns
the next element in the sequence.  Finally, we will define the meaning
of the plan based on the least fix-points of these transformations.

Let $f_\emptyset$ denote the function that maps any situation into the 
empty situation $\emptyset$. 

\begin{definition}
  Let $\alpha$ be a plan and $\Gamma$ a function that maps plans and
  situations into situations. Let $\varphi$ be a conjunction of fluent 
literals.
  Then, we define the function $\Fap:\P \rightarrow \P$ such that for
  any function $f \in \P,$ 

  $\Fap(f)(\Sigma) = \left\{
  \begin{array}{ll}
    \ \Sigma & \mbox{if $\varphi$ is false in $\Sigma$} \\
  f(\Gamma(\alpha,\Sigma)) & \mbox{if
  $\varphi$ is true in $\Sigma$ }\\
  \emptyset & \mbox{otherwise}
  \end{array}\right.$
\end{definition}
\noindent
We can define the powers of $\Fap$ as follows:
\begin{enumerate}
\item $\Fap\uparrow 0 = f_{\emptyset}$.
\item $\Fap\uparrow n+1 = \Fap(\Fap\uparrow n)$.
\item $\Fap\uparrow\omega = 
\ldots\Fap\ldots(\Fap(\Fap(\Fap\uparrow 0))\ldots)\ldots$, i.e. the
infinite composition of $\Fap$ applied to $f_\emptyset$.
\end{enumerate}
It can be shown that this power is correctly defined.  Proof and a
formal definition of powers can be found in Appendix \ref{denotational sem}.
%%%%%%%%%%%%%%%%%%%%%%%%%%%%%%%%%%%%%%%%%%%%%%%%%%%%%%%%%%%%%%%%%%%
%%%%%%%%%%%%%%%%%%%%%%%%%%%%%%%%%%%%%%%%%%%%%%%%%%%%%%%%%%%%%%%%%%%
%%%% COMMENT %%%%%%%%%%%%%%%%%%%%%%%%%%%%%%%%%%%%%%%%%%%%%%%%%%%%%%

We now extend the definition of the evaluation function $\Gnp{}$
to apply to plans with routines by adding item

\begin{itemize}
\item[5.]
$\Gnp{}([\While \varphi \Do \alpha_1|\alpha],\Sigma) =
\Gamma_\Phi(\alpha,\Sigma')$, where $\Sigma' = 
  {\cal F}_{{\bf if}\varphi{\bf
      then}\alpha_1,\varphi}^{\Gnp{}}\!\uparrow\!\omega$

\end{itemize}
to Definition~\ref{exe-func}.   The definition of entailment remains
unchanged.  That is, $D \models \varphi \after \alpha$ iff for every
model $\Phi$ of $D$, $\varphi$ is true in
$\Gamma_\Phi(\alpha,\Sigma_0)$. 

\begin{example}
$$
\begin{array}{ll}
D_5 \models iceInCups \after 
[ & \While \neg solidIce\; \Do [drop,pickup,checkIce], \\
  & putIceInCups]
\end{array}
$$
\end{example}

\subsection{Plan Termination}

Notice that the query above with the while loop could have been
written using three nested conditionals.  A more natural example will
replace rules $t6-10$ with the single rule
\[
 drop \;\mbox{\bf may affect}\; solidIce
\]
However, in this domain we are not be able to prove termination.  The
verification of termination is a difficult task, especially for planning.
How do we really know that the ice will eventually break? Or how do we
know that the cup is filling up?  With time the ice will either melt
or break, and if we do not place infinitesimally small amounts of ice
in the cup the cup will eventually fill up or we will run out of ice.
We have simplified the problem in our example by adding $t6-t10$.
These propositions state that the ice will break with no more than three
``drops".

We are faced with a similar situation in the following example.

\begin{example}  \label{ex:cans}
Consider the following situation.  On the floor of a room there are
  cans. An agent is given an empty bag and instructed to fill the bag
  with cans. We assume that there are more than enough cans on the
  floor to fill the bag. The domain description for this task is
\[D_4\left\{
\begin{array}{ll}
r_1 : {\bf \;initially}\; \neg bagFull \\
r_2 : drop {\bf \;causes}\; \neg canInHand \\
r_3 : drop {\bf \;causes}\; canInBag {\bf\;if}\;  canInHand \\
r_4 : lookInBag \;\ctk \;  bagFull \\
r_5 : pickUp {\bf \;causes}\; canInHand {\bf\;if}\;  \neg canInHand \\
\end{array} \right. \] 
This task of picking up cans and dropping them into the bag involves
the repetition of a small sequence of actions. There is a degree of
uncertainty inherent in this task because it is unknown how many cans
are needed to fill the bag. Therefore a loop that executes
the sequence of actions repeatedly until the task is completed is needed. 
If the number of cans needed to fill the bag is known beforehand, then the
set of actions would be repeated sequentially for those number of
times.
%\hfill $\Box$
\end{example}
%

%It cannot be shown that $D_4$ entails the following query,
\noindent A query that we would like to prove is 
\[\begin{array}{ll}
bagFull \after [ lookInBag, 
                  \While \neg bagFull \Do
                  [ & pickUpCan,\\ & dropCanInBag, \\ & lookInBag]]
\end{array}
\]
Ideally, we would like to use a routine which could solve any type of
task that involves uncertainty of its end.  However, each task has its
own conditions for termination.  For example, filling the volume of a
bag differs from finding an unfamiliar store in an unfamiliar area
based on the vague directions of a stranger. Do we really know the bag
will become full? Or how useful are vague directions such as, ``Just
walk down Lincoln Avenue, you can't miss it" when generating a plan. A
hole may tear in the bag, or suppose that the stranger who had all the
best intentions was mistaken about the location of the store.
To ensure termination (either with success or failure) we need to add
to our domain descriptions general axioms or constraints.  We do not
have constraints in $\Ak$ but we may be able to add them by using
other extensions of $\cala$ such as the one in \cite{bgp:RepAct}. To
address this problem we should first look at the standard techniques
of problem verifications such as the ones founded in \cite{AU95} or
\cite{Cousot90}.  These classical ideas have been used
by Manna and Waldinger to prove termination of plans with loops but
without sensing actions \cite{MW87}.  For sensing, it might also  be
useful to consider the techniques described in \cite{gb98} to detect
loop-termination using probability approaches.  However, analysis of
the termination of plans is outside the scope of this paper.  We will
discuss in Section~\ref{sec:future} how some of the problems of
termination may be addressed in simple situations.

%%%%%%%%%%%%%%%%%%%%%%%%%%%%%%%%%%%%%%%%%%%%%%%%%%%%%%%%%%%%%%%%%%%%
%%%%% Epistemic Logic Programs %%%%%%%%%%%%%%%%%%%%%%%%%%%%%%%%%%%%%
%%%%%%%%%%%%%%%%%%%%%%%%%%%%%%%%%%%%%%%%%%%%%%%%%%%%%%%%%%%%%%%%%%%%

\section{Epistemic Logic Programs}\label{ELP}
In the past, domain descriptions of dialects of $\cala$ 
have been translated into extended logic programs 
\cite{bg:ReCa,gl:Alang}.  
Extended logic programs use two types of negation to represent
incomplete information. There is strong or classical negation $\neg$
and negation as failure $not$.  The semantics of extended logic
programs is defined by a collection of sets of literals called answer
sets \cite{gl:ExLogic}.  However, we are required to represent
incomplete information that crosses over multiple sets of answer
sets. This will be the case in our translation of domain descriptions
into logic programs where situations will be closely related to sets
of answer sets, and domain descriptions act over epistemic states
which are sets of situations.  In this case, extended logic programs
will no longer be sufficient to codify domain descriptions.

Gelfond has extended disjunctive logic programs to work with sets of sets of 
answer sets \cite{gl:ExLogic}. He calls his new programs 
{\em epistemic logic programs\/}.  In epistemic logic programs, the 
language of extended logic programs is expanded with two
modal operators $K$ and $ M$. $KF$ is read as ``$F$ is known to be true"
and $MF$ is read as ``$F$ may be believed to be true."  

Universal and existential quantifiers are also allowed as well 
as the epistemic disjunctive ``$or$''  which the semantics is based on the
minimal model semantics associated with disjunctive logic programs
\cite{LMR92}. As an example, when $F \;or\;G$ is  defined as 
a logic program, its models are exactly ${F}$ and ${G}$.  Note that
the classical $F \vee G$ cannot be defined as a logic program,
because it has models which are not minimal.
  
In the rest
of this section we will review the syntax and the semantics of the
subclass of epistemic logic programs that will be required to represent
our domain descriptions.  Readers interested in more details about
epistemic logic programs are referred to \cite{gel:LpRii}.

\medskip \noindent The semantics of an epistemic logic program is
defined by pairs $ \langle A,W \rangle $.  $A$ is a collection 
of sets of
ground literals called the set of {\em possible beliefs\/}. Each set in
$A$ can be indexed as $A$ = $\{ A_1 \dots A_n \} $.  $W$ is a set in
$A$ called the {\em working set of beliefs\/}.  
To define the semantics, we restrict our formulas
to be: ground literals, a ground literal preceded by a modal operator,
a ground literal preceded by a modal operator and $\neg$, or a
conjunction of such formulas.  The truth of a formula $F$ in $\langle
A,W\rangle$ is denoted by $\langle A,W\rangle\models F$ and the
falsity by $\langle A,W\rangle\sledom F$, and
are defined as follows.

\medskip
$\langle A,W \rangle \; \models \; F$ iff $F \in W$, when $F$ is a
ground atom.

\medskip
$\langle A,W \rangle \; \models \; KF$ iff $ \langle A,A_i \rangle
\models F ,\; \forall A_i \in A $.

%\medskip
%$\langle A,W \rangle \; \models \; MF$ iff $ \langle A,A_i \rangle
%\models F ,\; \exists A_i \in A $.

\medskip
$\langle A,W \rangle \; \models \; F \wedge G$ iff $ \langle A,W
\rangle \models F \; $ and $ \langle A,W \rangle \models G $.

%\medskip
%$\langle A,W \rangle \; \models \; \exists xF $ iff there is a ground
%term such that $ \langle A,W \rangle \models F(t) $.

\medskip
$\langle A,W \rangle \; \models \; \neg F$ iff $ \langle A,W \rangle
\; \sledom \; F$.

\medskip
$\langle A,W \rangle $ $\sledom$ $F$ iff $\neg
F\in W$, when $F$ is a ground atom.

\medskip
$\langle A,W \rangle \; \sledom \; KF$ iff $ \langle A,W \rangle
\not\models KF$

%\medskip
%$\langle A,W \rangle \; \sledom\; MF$ iff $ \langle A,W \rangle \not
%\models MF $.

\medskip
$\langle A,W \rangle \; \sledom \; F \wedge G$ iff $ \langle A,W
\rangle \sledom F \; $ or $ \langle A,W \rangle \sledom G $.

%\medskip
%$\langle A,W \rangle \; \sledom \; \exists xF $ iff there is a ground
%term such that $ \langle A,W \rangle \sledom F(t) $.

\medskip
$\langle A,W \rangle \; \sledom \; \neg F$ iff $ \langle A,W \rangle
\; \models \; F$.

\medskip
$\langle A,W \rangle \models
 \;F\;or\;G$ iff $\langle A,W \rangle \models
 \;\neg(\neg F\;\wedge\;\neg G)$

\medskip \noindent Note that when a formula $G$ is of the form $KF$,
or $\neg KF$, its evaluation in $\langle A,W \rangle$ does not depend
on $W$.  Thus, we will write $A\models\; G$ or $A\sledom\; G$.
Moreover, the evaluation of object formulas does not depends on
$A$. If $G$ is objective we sometimes write $W\models\; G$ or
$W\sledom\; G$.

An epistemic logic program is a collection of rules of the form

\begin{equation}
F_1\;or\ldots or\; F_k\; \leftarrow \;G_1,\ldots,G_m, {\bf not}\;
F_{m+1}, \ldots ,{\bf not}\;F_n
\end{equation}

\noindent
where $F_1\ldots F_k$ and $F_{m+1}\ldots F_n$ 
are (not necessarily ground) objective literals 
(without $K$ or $M$) and $G_{1} \ldots G_m$ are (not necessarily ground) 
subjective (with $K$ or $M$) or objective literals.  

\medskip \noindent Let $\Pi$ be an epistemic logic program without
variables, ${\bf not}$, or modal operators. A set $W$ of ground
literals is a belief set of $\Pi$ if it is a minimal set of ground
literals, satisfying the following properties:

\begin{enumerate}
%\item $W \models F_1\; or\ldots or \;F_k$ for every rule 
%$F_1 \;or\ldots or \;F_k\;\leftarrow\;G_1 \ldots G_m$ in $\Pi$ which 
\item $W \models F$ for every rule 
$F\;\leftarrow\;G_1 \ldots G_m$ in $\Pi$ which 
$W\models G_1 \wedge \ldots \wedge G_m$.
\item If there is a pair of complementary literals, i.e $F\; and \; \neg F$, 
in $W$ then $W$ is the set of all literals.
\end{enumerate}

\medskip \noindent Let $\Pi$ be an epistemic logic program with ${\bf
not}$ and variables but does not contain any modal operator. Let
$Ground(\Pi)$ be the epistemic logic program that is obtained from $\Pi$
by replacing each rule in $\Pi$ with all its ground instances.  Let
$W$ be a set of ground literals ( literals in $W$ and $\Pi$  are from
the same language). $\Pi^W$ is obtained from $\Pi$ by
removing from $Ground(\Pi)$

\begin{enumerate}
\item All the rules which contain formulas of the form ${\bf not}\; G$
such that $W \models G$.
\item All occurrences of formulas of the form ${\bf not}\; G$ from the
remaining rules.
\end{enumerate}
$W$ is a belief set of $\Pi$ if and only if $W$ is a belief set 
of $\Pi^W$.

Let $\Pi$ be any epistemic logic program, and $\bf A$ a collection of sets
of literals. $[\Pi]_{\bf A}$ is the epistemic logic program obtained by
removing from $Ground(\Pi)$

\begin{enumerate}
\item All rules with formulas of the form $G$ such that $G$ contains
  $M$ or $K$, and ${\bf A} \not \models G,$
\item All occurrences of formulas containing $M$ or $K$ from the
  remaining rules.
\end{enumerate}

\noindent A set $\bf A$ is a $world \;view$ of $\Pi$ if $\bf A$ is the
collection of all belief sets of $[\Pi]_{\bf A}$.  A world view of
$\Pi$ is consistent if it does not contain the belief set of all
literals.  An epistemic logic program is consistent if it has
at least one consistent non-empty world view.  In epistemic logic
programs the only {\em working sets of  beliefs\/} that are considered
are world views and the {\em possible belief\/} is always a member of
the working set under consideration (i.e. a belief set).

\medskip \noindent Let $\Pi$ be an epistemic logic program and $\bf A$
be a world view of $\Pi$. A literal $L$ is true in $\bf A$ iff for
every ground instance $F$ of $L$, $\langle {\bf A},A_i \rangle \models
F$ for all $A_i$ in $\bf A$.  $F$ is true in $\Pi$, denoted by $\Pi\models
F$, iff ${\bf A} \models F$ for every world view $\bf A$ of $\Pi$.

\begin{example} \label{pq}
The epistemic program

\begin{enumerate}
\item $q(a)\;\leftarrow \neg Kp(a).$
\item $p(a) \; \leftarrow \neg Kq(a).$
\end{enumerate}
has two world views
%\footnote{aqui es world views} 

$\;\;\;\;\;\;\;\;\;\;$
$\{ \{ p(a) \} \}$         %\footnote{debe tener doble llave}
$\;\;\;\;\;\;\;\;\;\;$
$\{ \{ q(a) \} \}$         %\footnote{debe tener doble llave}

\noindent
In the first world view $Kp(a)$  is true and $Kq(a)$ is true in the
second.

\medskip
\noindent
The epistemic program
\begin{enumerate}
\item $q(a)\;or\;q(b).$
\item $p(a) \; \leftarrow \neg Kq(a).$
\end{enumerate}
has one world view

$\;\;\;\;\;\;\;\;\;\;$
$\{ \{p(a), q(a)\}, \{p(a), q(b)\} \}$

\medskip
\noindent
Note that $Kq(a)$ is not true in this world view
 because $q(a)$ is not member of the
second belief set.
%Thus, this epistemic logic program gives us a belief set which is able to
%answer $yes$ to the query $Interview(mike)$ just as we would expect.
The main intuition to have when reading a formula of the form $KF$ is
that it will be true iff $F$ is true in every belief set of the program.
%\hfill $\Box$
\end{example}

%%%%%%%%%%%%%%%%%%%%%%%%%%%%%%%%%%%%%%%%%%%%%%%%%%%%%%%%%%%%%%%%%%%%
%%%%% Translation Epistemic Logic Programs %%%%%%%%%%%%%%%%%%%%%%%%%
%%%%%%%%%%%%%%%%%%%%%%%%%%%%%%%%%%%%%%%%%%%%%%%%%%%%%%%%%%%%%%%%%%%%

\section{Translation to Epistemic Logic Programs} \label{sec:translation}

In this section we start with a sound and complete translation of {\em
simple\/} domain descriptions into epistemic logic programs.  This will
let us explain the logic program rules under the simple scenario and
will make clear the rules for the general case.

Our epistemic logic programs will use variables of three sorts: {\em
  situation\/} variables denoted by $S$ or $S'$ possibly indexed, {\em
  fluent\/} variables denoted by $F$ or $F'$ possibly indexed, {\em
  action\/} variables denoted by $A$ or $A'$ possibly indexed, and the
special situation constant $s_0$ that represents the initial
situation.  We will also have a constant symbol for each fluent
symbol $f$ in the language and we add the constant symbol $\o{f}$ to
represent $\neg f$. For simplicity we will denote the fluent
literal constants by the fluent literal they represent.  We will also
add the special constant symbol $true$ to the set of fluent literal
constants.

\subsection{The Domain Independent Translation}
We start by first giving the rules for inertia. These rules encode
that a fluent remains unchanged if no actions that affect the fluent is
executed.   Whenever a fluent literal appears as an
argument in a predicate, it is representing a corresponding constant in
the program.  For any fluent literal $l$, if $l = \neg f$, $\o{l}$ will
denote  $f$ in the program.

\medskip\noindent 
For every fluent literal $f$ there is an inertia rule of the
form:

\medskip
$holds(f,res(A,S))\leftarrow$ $holds(f,S),{\bf not}$ $ab(\o{f}, A,S).$

%\noindent
%$holds(\o{F},res(A,S))\leftarrow$ $holds(\o{F},S),{\bf not}\; ab(F,A,S).$

\medskip \noindent For every fluent symbol $f$ there is an
or-classicalization rule of the form

\medskip
$holds(f,s_0) \;or\; holds(\o{f},s_0)$

\medskip \noindent 
The above rule states that our belief sets are complete in the sense that
either $holds(f,s_0)$ or  $holds(\o{f},s_0)$  must be true since
$f \vee \neg f$ is a tautology in every state.  Note that because of
the minimal model semantics interpretation of the ``$or$'' we will not
have both $holds(f,s_0)$ or $holds(\o{f},s_0)$ holding simultaneously.

\medskip\noindent 
We will also have two more domain independent rules that we will
call rules of {\em suppression\/}.  

\medskip
$holds(true,res(A,S)) \leftarrow holds(true,S)$

$holds(F,S) \leftarrow holds(true,S)$

\medskip
\noindent
These rules will be used to implement compatibility.  For example, if
a situation $\Sigma = \{\sigma_1, \sigma_2\}$ with two states is split
into two situations $\Sigma_1 = \{\sigma_1\}$ and $\Sigma_2 =
\{\sigma_2\}$, for compatibility after the execution of a sensing action,
$\Sigma_1$ will be generated by suppressing $\sigma_2$ from $\Sigma$
using these rules.  How this is accomplished will become apparent when
we introduce the domain dependent rules produced by the knowledge
laws.

\subsection{The Domain Dependent Translation}
\label{domain dependent example}
Value propositions of the form ``${\bf initially}$ $f$" 
are translated into
\[
holds(f,s_{0})
\]
The translation of effect propositions of the form ``$a$ ${\bf
causes}$ $f$ ${\bf if}$ $p_{1},\ldots ,p_{n}$" is the standard
translation for effect propositions introduced by Gelfond and
Lifschitz in \cite{gl:Alang} for $\cala$.  The translation produces
two rules.  The first one is:
\[
holds(f,res(a,S))\leftarrow\holdp1,\ldots,\holdpn
\]
It allows us to prove that $f$ will hold after the result of the
execution of $a$ if preconditions are satisfied. The second rule is:
%
%$holds(f,result(a,S))\leftarrow\holdp1$,$\ldots$,$\holdpn$.
%
\[
ab(f,a,S)\leftarrow\holdp1,\ldots,\holdpn, \Not holds(true,res(a,S))
\]
%
%$abnormal(f,a,S)\leftarrow\holdp1$,$\ldots$,$\holdpn$.
%
\noindent 
where the predicate $ab(f,a,S)$ disables the inertia rule in the 
cases where $f$ can be affected by $a$.

We will introduce the domain dependent translation of knowledge laws
using the following domain description.
\begin{example}~\\
\label{ex-d01}
\[
D_{1}^0\left\{
\begin{array}{ll}
%r_1 : {\bf \;initially}\; \neg burnOut \\
r_1 : {\bf \;initially}\; \neg bulbFixed \\
r_2 : checkSwitch {\bf\;causes\;to\;know}\;switchOn{\bf\;if}\;\neg burnOut \\
\end{array} \right. \] 
%
%$\;\;\;\;\;\;\;\;\;\;$
%\noindent
%\noindent
\end{example}

\noindent
In this example the initial situation is:
\[
\begin{array}{ll}
\Sigma =& \{
\{burnOut, \neg bulbFixed, switchOn\}, \\ & 
\{burnOut, \neg bulbFixed, \neg switchOn\}, \\
&
\{\neg burnOut, \neg bulbFixed, switchOn\}, \\ &
\{\neg burnOut, \neg bulbFixed, \neg switchOn\}\} 
\end{array}
\]
after the robot executes the action $checkSwitch$ we will have
the following  resulting situations:
\[
\begin{array}{ll}
\Phi_1 (checkSwitch,\Sigma) = \{ & \{\neg burnOut,
\neg bulbFixed, switchOn\}\} \\
\Phi_2 (checkSwitch,\Sigma) = \{ & \{\neg burnOut,
\neg bulbFixed, \neg switchOn\}\}\\
\Phi_3 (checkSwitch,\Sigma) = 
\{ &
\{burnOut, \neg bulbFixed, switchOn\}, \\
&
\{burnOut, \neg bulbFixed, \neg switchOn\}\}
\end{array}
\]
These correspond to the three $(switchOn,\neg burnOut)$-compatible
sub-sets of $\Sigma$ (see Definition~\ref{def:comp}).  Note also that
\[\begin{array}{l}
\Phi_1(checkSwitch,\Phi_1 (checkSwitch,\Sigma)) = 
\Phi_1(checkSwitch,\Sigma)\\
\Phi_2(checkSwitch,\Phi_2
(checkSwitch,\Sigma)) = \Phi_2 (checkSwitch,\Sigma) \\
\Phi_3(checkSwitch,\Phi_2
(checkSwitch,\Sigma)) = \Phi_3 (checkSwitch,\Sigma)
\end{array}
\]
Our logic program translation of this domain will have three world
views, one corresponding to each of the transition functions $\Phi_1$
$\Phi_2$, and $\Phi_3$.  $\Phi_1$ is depicted on the left hand side of
the figure below, $\Phi_2$ on the right hand side and $\Phi_3$ in the
middle.

%\begin{center}
\noindent $\:\:\:\:\:\:\:\:\:\:\:\:$
\begin{picture}(0,0)%
\epsfig{file=newnew.pstex}%
\end{picture}%
\setlength{\unitlength}{3947sp}%
\begingroup\makeatletter\ifx\SetFigFont\undefined%
\gdef\SetFigFont#1#2#3#4#5{%
  \reset@font\fontsize{#1}{#2pt}%
  \fontfamily{#3}\fontseries{#4}\fontshape{#5}%
  \selectfont}%
\fi\endgroup%
\begin{picture}(5934,5382)(2479,-6286)
\put(6811,-1651){\makebox(0,0)[b]{\smash{\SetFigFont{12}{14.4}{\rmdefault}{\mddefault}{\updefault}$\neg switchOn$}}}
\put(6811,-2101){\makebox(0,0)[b]{\smash{\SetFigFont{10}{12.0}{\rmdefault}{\mddefault}{\updefault}$\neg burnOut$  $\neg bulbFixed$}}}
\put(6811,-2476){\makebox(0,0)[b]{\smash{\SetFigFont{10}{12.0}{\rmdefault}{\mddefault}{\updefault}$\neg switchOn$}}}
\put(4216,-2116){\makebox(0,0)[b]{\smash{\SetFigFont{10}{12.0}{\rmdefault}{\mddefault}{\updefault}$\neg burnOut$  $\neg bulbFixed$}}}
\put(4216,-2491){\makebox(0,0)[b]{\smash{\SetFigFont{10}{12.0}{\rmdefault}{\mddefault}{\updefault}$switchOn$}}}
\put(7051,-3556){\makebox(0,0)[b]{\smash{\SetFigFont{10}{12.0}{\rmdefault}{\mddefault}{\updefault}$\neg burnOut$  $\neg bulbFixed$}}}
\put(7051,-3931){\makebox(0,0)[b]{\smash{\SetFigFont{10}{12.0}{\rmdefault}{\mddefault}{\updefault}$\neg switchOn$}}}
\put(4501,-5221){\makebox(0,0)[b]{\smash{\SetFigFont{10}{12.0}{\rmdefault}{\mddefault}{\updefault}$burnOut$  $\neg bulbFixed$}}}
\put(4501,-5596){\makebox(0,0)[b]{\smash{\SetFigFont{10}{12.0}{\rmdefault}{\mddefault}{\updefault}$switchOn$}}}
\put(7111,-5236){\makebox(0,0)[b]{\smash{\SetFigFont{10}{12.0}{\rmdefault}{\mddefault}{\updefault}$burnOut$  $\neg bulbFixed$}}}
\put(7111,-5611){\makebox(0,0)[b]{\smash{\SetFigFont{10}{12.0}{\rmdefault}{\mddefault}{\updefault}$\neg switchOn$}}}
\put(2551,-2686){\makebox(0,0)[lb]{\smash{\SetFigFont{10}{12.0}{\rmdefault}{\mddefault}{\updefault}$\Phi_1$}}}
\put(2746,-3151){\makebox(0,0)[lb]{\smash{\SetFigFont{10}{12.0}{\rmdefault}{\mddefault}{\updefault}$checkSwitch$}}}
\put(5461,-2986){\makebox(0,0)[lb]{\smash{\SetFigFont{10}{12.0}{\rmdefault}{\mddefault}{\updefault}$\Phi_3$}}}
\put(5851,-3136){\makebox(0,0)[lb]{\smash{\SetFigFont{10}{12.0}{\rmdefault}{\mddefault}{\updefault}$checkSwitch$}}}
\put(3256,-4516){\makebox(0,0)[lb]{\smash{\SetFigFont{10}{12.0}{\rmdefault}{\mddefault}{\updefault}$checkSwitch$}}}
\put(3061,-4366){\makebox(0,0)[rb]{\smash{\SetFigFont{10}{12.0}{\rmdefault}{\mddefault}{\updefault}$\Phi_1$}}}
\put(3121,-6136){\makebox(0,0)[lb]{\smash{\SetFigFont{10}{12.0}{\rmdefault}{\mddefault}{\updefault}$\Phi_3$}}}
\put(3451,-6286){\makebox(0,0)[lb]{\smash{\SetFigFont{10}{12.0}{\rmdefault}{\mddefault}{\updefault}$checkSwitch$}}}
\put(4201,-1261){\makebox(0,0)[b]{\smash{\SetFigFont{10}{12.0}{\rmdefault}{\mddefault}{\updefault}$burnOut$  $\neg bulbFixed$}}}
\put(4201,-1636){\makebox(0,0)[b]{\smash{\SetFigFont{10}{12.0}{\rmdefault}{\mddefault}{\updefault}$switchOn$}}}
\put(6811,-1276){\makebox(0,0)[b]{\smash{\SetFigFont{10}{12.0}{\rmdefault}{\mddefault}{\updefault}$burnOut$  $\neg bulbFixed$}}}
\put(4216,-3571){\makebox(0,0)[b]{\smash{\SetFigFont{10}{12.0}{\rmdefault}{\mddefault}{\updefault}$\neg burnOut$  $\neg bulbFixed$}}}
\put(4216,-3946){\makebox(0,0)[b]{\smash{\SetFigFont{10}{12.0}{\rmdefault}{\mddefault}{\updefault}$switchOn$}}}
\put(8101,-2911){\makebox(0,0)[rb]{\smash{\SetFigFont{10}{12.0}{\rmdefault}{\mddefault}{\updefault}$checkSwitch$}}}
\put(7726,-4486){\makebox(0,0)[rb]{\smash{\SetFigFont{10}{12.0}{\rmdefault}{\mddefault}{\updefault}$checkSwitch$}}}
\put(8176,-2311){\makebox(0,0)[lb]{\smash{\SetFigFont{10}{12.0}{\rmdefault}{\mddefault}{\updefault}$\Phi_2$}}}
\put(8251,-4261){\makebox(0,0)[rb]{\smash{\SetFigFont{10}{12.0}{\rmdefault}{\mddefault}{\updefault}$\Phi_2$}}}
\end{picture}

%\end{center}

\noindent
The world view associated with $\Phi_1$ on the left hand side of the
figure will have four belief sets.  One will contain the union of the
two sets

$W^1_{s_0} = \{holds(\overline{burnOut},s_0),
 holds(\overline{bulbFixed},s_0),
 holds(switchOn,s_0) \}$

\noindent
and

%**************************
%$W^1_{res(ts,res(cs,s_0))} = \{\\
%~~\;\;\;\;\;\;\;\;\;\;\;
% holds(\overline{burnOut},res(turnSwitch,res(checkSwitch,s_0))),\\
%~~\;\;\;\;\;\;\;\;\;\;\;
% holds(\overline{bulbFixed},res(turnSwitch,res(checkSwitch,s_0)), \\
%~~\;\;\;\;\;\;\;\;\;\;\;
% holds(\ol{switchOn},res(turnSwitch,res(checkSwitch,s_0)))\\
%~~\;\;\;\;\;\;\;\;\;\}$\\
$\begin{array}{ll}
W^1_{res(cs,s_0)} = \{ &
 holds(\overline{burnOut},res(checkSwitch,s_0)), \\ &
 holds(\overline{bulbFixed},res(checkSwitch,s_0),\\&
 holds(switchOn, res(checkSwitch,s_0))\}
\end{array}$

\medskip
\noindent
This union represents the fact that $\{\neg burnOut, \neg bulbFixed,
switchOn\}$ is an initial state (encoded $W^1_{s_0}$) and that the
same set is also a state in $\Phi_1(checkSwitch,\Sigma)$ (encoded in
$W^1_{res(cs,s_0)}$).  The rest of the literals in $W^1$ are the same
as in $W^1_{res(cs,s_0)}$ except that the situation constant in each
literal is replaced by situation constants of the form
$res(checkSwitch, res(
\ldots, res(checkSwitch,s_0)\ldots))$ representing that the state
remains the same after any number of applications of the action
$checkSwitch$ to the state (the loop arc on the left of the figure).

\medskip
\noindent
The second belief set will contain 

$W^2_{s_0} =
\{holds(\overline{burnOut},s_0)$, $holds(\overline{bulbFixed},s_0)$,
$holds(\overline{switchOn},s_0) \}$

\medskip
\noindent
representing that $\{\neg burnOut, \neg bulbFixed, \neg switchOn\}$ is
also an initial state.  However, this state is not part of
$\Phi_1(checkSwitch,\Sigma)$.  Then, we need to suppress this state
from the world view.  We will do that by adding the set
\[\begin{array}{lrl}
W^2_{res(cs,s_0)} &=  \{ &holds(\overline{burnOut},res(checkSwitch,s_0)),\\
& &
holds(\overline{bulbFixed},res(checkSwitch,s_0),\\
&& holds(\ol{switchOn},
res(checkSwitch,s_0)) \\ & \} &
\bigcup  \\ & \{ &
 holds(burnOut,res(checkSwitch,s_0)), \\
&& holds(bulbFixed,res(checkSwitch,s_0),\\
&& 
 holds(switchOn,res(checkSwitch,s_0)), \\
&&
 holds(true,res(checkSwitch,s_0)\}
\end{array}
\]
to the belief set.  Actually, we will have in the domain dependent
translation a rule that adds $holds(true,res(checkSwithc,s_0))$, and
the second domain independent suppression rule will add the rest.  The
rest of the literals in $W^2$ are the same as in $W^2_{res(cs,s_0)}$
except that the situation constant is replaced by situation constants
of the form $res(checkSwitch, res( \ldots,
res(checkSwitch,s_0)\ldots))$ representing that the state remains
suppressed in the result of applying the action $checkSwitch$ to the
state.  This is the effect of the first domain independent suppression
rule.

The other two belief sets $W^3$ and $W^4$ are similar to $W^2$.

\medskip
$W^3_{s_0} =
\{holds({burnOut},s_0)$, $holds(\overline{bulbFixed},s_0)$,
$holds({switchOn},s_0) \}$

$W^4_{s_0} =
\{holds(burnOut,s_0)$, $holds(\overline{bulbFixed},s_0)$,
$holds(\overline{switchOn},s_0) \}$

\medskip
\noindent
The rest of $W^3$ and $W^4$ is exactly as in $W^2$ since the states
they represent are also suppressed from the result.

Note that both $holds(\ol{f},res(checkSwitch,s_0))$ and
$holds(f,res(checkSwitch,s_0))$ are members of the belief sets $W^2$,
$W^3$ and $W^4$, for any fluent $f$.  Therefore, for any fluent
literal $g$, the proof of $holds(g,res(checkSwitch,s_0)$ in the world
view is not affected by these belief sets.  The consequence is that we
are ignoring three states after the execution of $checkSwitch$ under
the model $\Phi_1$.

There are two more world views that correspond to the transitions in
the middle and on the right hand side of
the figure.  The definition is very similar to the first world view.
There are four belief sets in the middle, two of them suppressing
initial states, and four belief sets in the last world view, three of
them suppressing initial states.  

Thus, the domain dependent translation of $D^0_1$ will be:
$$holds(\overline{bulbFixed}, s_0) \leftarrow$$
Rule $x_1$ is the translation of rule $r_1$.   The rest of the rules
correspond to the different suppression cases since states that are not 
suppressed by the transition will be moved to the next situation by the
domain independent rule of inertia.  Take for example, $\Phi_1$.
\[
\Phi_1(checkSwitch,\Sigma) = \{ \sigma\in\Sigma | \sigma\models \neg
burnOut, switchOn\in\sigma \}
\]
Hence, we would like to suppress two kinds of states. 1) States where
$\neg switchOn$ is true, and 2) States where $burnOut$ is true.  The
rule for the first case is: 
{\small\begin{eqnarray}
holds(true,res(checkSwitch,S)) & \leftarrow & 
Kholds(switchOn,res(checkSwitch,S)) \nonumber \\
& & Kholds(\overline{burnOut},res(checkSwitch,S)), \nonumber \\
& &
holds(\ol{switchOn},S)  
\end{eqnarray}}
The first two literals in the body of the rule verify that we are in
the case of $\Phi_1$, that is, both $switchOn$ and $\neg burnOut$ are
true in every state of the resulting situation (i.e. the two literals
$Kholds(switchOn,res(checkSwitch,S))$ and
$Kholds(\ol{burnOut},res(checkSwitch,S))$ are true).  The last
predicate checks that we are suppressing the state where $\neg
switchOn$ is true in the current situation
(i.e. $holds(\ol{switchOn},S)$).

The rule for the second case is very similar.  We only need to change
the last literal to indicate that we are suppressing the state where
$burnOut$ is true (i.e. $holds(burnOut,S)$):
{\small\begin{eqnarray}
holds(true,res(checkSwitch,S)) & \leftarrow & 
Kholds(switchOn,res(checkSwitch,S)) \nonumber \\
& & Kholds(\overline{burnOut},res(checkSwitch,S)), \nonumber \\
& &
holds(burnOut,S) 
\end{eqnarray}}
Let us look now at $\Phi_2$.  
\[
\Phi_2(checkSwitch,\Sigma) = \{ \sigma\in\Sigma | \sigma\models \neg
burnOut, switchOn\not\in\sigma \}
\]
We also suppress two kinds of states. 1) States where $switchOn$ is
true, and 2) States where $burnOut$ is true.  We need to check that
$\neg switchOn$ and $\neg burnOut$ are true in every state of the
resulting situation (i.e. $Kholds(\ol{switchOn},res(checkSwitch,S))$
and $Kholds(\ol{burnOut},res(checkSwitch,S))$ are true) to verify that
we are in the case of $\Phi_2$.  The rules for the cases are:
{\small\begin{eqnarray}
holds(true,res(checkSwitch,S)) & \leftarrow & 
Kholds(\overline{switchOn},res(checkSwitch,S)) \nonumber \\
& & Kholds(\overline{burnOut},res(checkSwitch,S)), \nonumber \\
& &
holds(switchOn,S) \nonumber\\ 
holds(true,res(checkSwitch,S)) & \leftarrow & 
Kholds(\overline{switchOn},res(checkSwitch,S)) \nonumber \\
& & Kholds(\overline{burnOut},res(checkSwitch,S)), \nonumber \\
& &
holds(burnOut,S)   %\nonumber \\
\end{eqnarray}}

\noindent
For $\Phi_3$ all the states where $\neg
burnOut$ holds (i.e.  $holds(\ol{burnOut},S)$) should be suppressed since
\[
\Phi_3(checkSwitch,\Sigma) = \{ \sigma\in\Sigma | \sigma\not\models \neg
burnOut\}
\]
To verify that we are in the case of $\Phi_3$ we need to check there
is at least one state in the result where $burnOut$ holds (i.e. $\neg
Kholds(\ol{burnOut},res(checkSwithc,S))$).  The rule for this case is:
{\small\begin{eqnarray}
holds(true,res(checkSwitch,S)) & \leftarrow & 
\neg Kholds(\overline{burnOut},res(checkSwitch,S)) \nonumber \\
&& holds(\overline{burnOut},S)% \nonumber \\
\end{eqnarray}}

\noindent
There is a condition that must be added to all the rules.  The condition 
is that if the fluent $switchOn$ is already known in the original
situation (for example if we have $\initially \neg switchOn$) then
none of the states is suppressed from the situations.  In other words,
the rules above applied only if $switchOn$ is unknown. To check that
this is the case we must add to the body of each rule the literals $\neg
Kholds(switchOn,S)$ and $\neg Kholds(\ol{switchOn},S)$.  These
literals are not required in this particular example but it must be
part of the general case.   
%%%%%%%%%%%%%%%%%%%%%%%%%%%%%%%%%%%%%%%%%%%%%%%%%%%%%%%%%%%%%%%
%%%%%%%%%%%%%%%%%%%%%%%%%%%%%%%%%%%%%%%%%%%%%%%%%%%%%%%%%%%%%%%
%%%%%%%%%%%%%%%%%%%%%%%%%%%%%%%%%%%%%%%%%%%%%%%%%%%%%%%%%%%%%%%
%%%%%%%%%%%%%%%%%%%%%%%%%%%%%%%%%%%%%%%%%%%%%%%%%%%%%%%%%%%%%%%
%%%%%%%%%%%%%%%%%%%%%%%%%%%%%%%%%%%%%%%%%%%%%%%%%%%%%%%%%%%%%%%
%%%%%%%%%%%%%%%%%%%%%%%%%%%%%%%%%%%%%%%%%%%%%%%%%%%%%%%%%%%%%%%
%Now we can describe our domain dependent translation

\noindent 
In general, knowledge laws of the form ``$a$ ${\bf
causes\;to\;know}$ $f$ ${\bf if}$ $p_{1},\ldots ,p_{n}$" are
translated into the rules

%%%%%%%%%%%%%%%%%%%%%%%%%%%%%%%%%%%%%%%%%%%%%%%
%\[
{\small
\begin{eqnarray}
%%%%%%%%%%%%%%%%%%%%%%%case (1) %%%%%%%%%%%%%%%%%%%%%%%%%%%%%%
holds(true,res(a,S)) & \leftarrow & \neg Kholds(f,S),
\neg Kholds(\o{f},S), \nonumber \\
& & \neg Kholds(p_1,res(a,S)),  \nonumber \\
& & 
holds(p_1,S),\ldots,holds(p_n,S) \nonumber \\
 &\vdots  \nonumber \\
holds(true,res(a,S)) & \leftarrow & \neg Kholds(f,S),
\neg Kholds(\o{f},S), \nonumber \\
& & \neg Kholds(p_n,res(a,S)),
 \nonumber \\
& & 
holds(p_1,S),\ldots,holds(p_n,S) \label{s3}\\
%%%%%%%%%%%%%%%%%%%%%%%case (2) %%%%%%%%%%%%%%%%%%%%%%%%%%%%%%
holds(true,res(a,S)) & \leftarrow & \neg Kholds(f,S),\neg Kholds(\o{f},S),
Kholds(\o{f},res(a,S)), \nonumber \\
& & Kholds(p_1,res(a,S)),\ldots,Kholds(p_n,res(a,S)), \nonumber \\
& & holds(f,S) \label{r2}\\ 
holds(true,res(a,S)) & \leftarrow & \neg Kholds(f,S),\neg Kholds(\o{f},S),
Kholds(\o{f},res(a,S)), \nonumber \\
& & Kholds(p_1,res(a,S)),\ldots,Kholds(p_n,res(a,S)), \nonumber \\
& & holds(\o{p_1},S) \nonumber \\
&\vdots  \nonumber \\
holds(true,res(a,S)) & \leftarrow & \neg Kholds(f,S),\neg Kholds(\o{f},S),
Kholds(\o{f},res(a,S)), \nonumber \\
& & Kholds(p_1,res(a,S)),\ldots,Kholds(p_n,res(a,S)), \nonumber \\
& & holds(\o{p_n},S) \label{s2}\\
%%%%%%%%%%%%%%%%%%%%%%%case (3) %%%%%%%%%%%%%%%%%%%%%%%%%%%%%%
holds(true,res(a,S)) & \leftarrow & \neg Kholds(f,S),\neg Kholds(\o{f},S),
Kholds(f,res(a,S)), \nonumber \\
& & Kholds(p_1,res(a,S)),\ldots,Kholds(p_n,res(a,S)), \nonumber \\
& &
holds(\o{f},S) \label{r1}\\ 
%%%%%%%%%%%%%%%%%%%%%%%%%%%%%%%%%%%%%%%%%%%%%%%%%%%%%%%%%%%%%%%%
holds(true,res(a,S)) & \leftarrow & \neg Kholds(f,S),\neg Kholds(\o{f},S),
Kholds(f,res(a,S)), \nonumber \\
& & Kholds(p_1,res(a,S)),\ldots,Kholds(p_n,res(a,S)), \nonumber \\
& &
holds(\o{p_1},S)   \nonumber \\&\vdots \nonumber  \\
holds(true,res(a,S)) & \leftarrow & \neg Kholds(f,S),\neg Kholds(\o{f},S),
Kholds(f,res(a,S)), \nonumber \\
& & Kholds(p_1,res(a,S)),\ldots,Kholds(p_n,res(a,S)), \nonumber \\
& &
holds(\o{p_n},S) \label{s1}
\end{eqnarray}
} %of \small
%\]
%%%%%%%%%%%%%%%%%%%%%%%%%%%%%%%%%%%%%%%%%%%%%%%
We have added to every rule the condition $\neg
Kholds(f,S), \neg Kholds(\o{f},S)$.  None of these rules apply if $f$
is currently known.  In this case, by inertia everything stays the
same after the execution of the sensing action $a$.  Assume now that
neither $f$ nor $\o{f}$ holds in the ``situation'' $S$.  Thus,
according to the definition of compatibility (Def.~\ref{def:comp}), we
would have three types of world views.  1) One type for which we can
find a $p_i$ for each of the belief sets such $holds(p_i,res(a,S))$
does not hold; 2) World views in which $Kholds(\o{f},res(a,S))$ and
every $Kholds(p_i,res(a,S))$ hold; 3) World views in which
$Kholds({f},res(a,S))$ and every $Kholds(p_i,res(a,S))$ hold.

Intuitively, to capture these three cases the logic programming rules
will suppress the belief set that breaks the rules.  To suppress a
belief set in the situation $res(a,S)$ the rules will add
$holds(true,res(a,S))$ to the belief set, and this atom together with
the second suppression rule will add $holds(l,res(a,S))$, for every
fluent literal $l$.  Recall that the effect of having every literal
hold for a particular situation in a belief set is that the belief set
can be ignored when checking if the literal holds in the world
view.  Case (1) is captured by the first set of rules (\ref{s3}).  Case
(2) is captured by rule (\ref{r2}) and the set of rules
(\ref{s2}). Case (3) is captured by rule (\ref{r1}) and the set of
rules (\ref{s1}).

A non-deterministic effect proposition of the form
``$a$ ${\bf may\;affect}$ $f$ ${\bf if}$ $p_{1},\ldots ,p_{n}$" is
translated into

{%\medskip\noindent
\[\begin{array}{ll}
holds(f,res(a,S))\leftarrow & \Not holds(\o{f},res(a,S)), \\
& \holdp1,\ldots,\holdpn\\
holds(\o{f},res(a,S))\leftarrow & \Not holds(f,res(a,S)),\\
& \holdp1,\ldots ,\holdpn
\end{array}\]\[
\begin{array}{ll}
ab(f,a,S) \leftarrow & \Not  %\footnote{lo mismo}
holds(\o{f},res(a,S)),
\holdp1,\ldots ,\holdpn, \\
& \Not holds(true,S)\\
ab({\overline f},a,S)\leftarrow &
\Not holds(f,res(a,S)), \holdp1,\ldots ,\holdpn,\\
& \Not holds(true,S)
\end{array}
\]
%$~~~{\bf not} \;holds(f,res(a,S)), \holdp1,\ldots ,\holdpn$.
}
To illustrate this translation, we use Rule $t_6$ from domain
description $D_5$, and show how it will work from the initial
situation $s_0$. We also include the translation for $t_8$,
$holds(noDrops,s_0)$ and $holds(solidIce,s_0)$ along with the inertia
rule to get the following program. Notice that the suppression rules
do not apply here since we are not considering any knowledge laws.
\[\begin{array}{ll}
holds(solidIce,res(drop,s_0))\leftarrow  
& \Not  holds(\ol{solidIce},res(drop,s_0)),\\
& holds(noDrops,s_0).\\
holds(\ol{solidIce},res(drop,s_0))\leftarrow 
& \Not  holds(solidIce,res(drop,s_0)),\\
& holds(noDrops,s_0).
\end{array} \]\[
\begin{array}{ll}
ab(solidIce,drop,s_0)\leftarrow 
& \Not  holds(\ol{solidIce},res(drop,s_0)), \\
& holds(noDrops,s_0), 
%\\ & 
\Not holds(true,s_0).\\
ab(\overline{solidIce},drop,s_0)\leftarrow
& \Not holds(solidIce,res(drop,s_0)),\\
&  holds(noDrops,s_0),
% \\ & 
\Not holds(true,s_0).
\end{array}\]\[
\begin{array}{l}
holds(\ol{noDrops}, res(drop, s_0))\leftarrow holds(noDrops,s_0).\\
ab(\overline{noDrops},drop, s_0)\leftarrow holds(noDrops,s_0).
\end{array}\]\[
\begin{array}{l}
holds(noDrops,s_0) \leftarrow\\ 
holds(solidIce,s_0) \leftarrow\\ 
\end{array}\]\[
\begin{array}{ll}
holds(noDrops,res(drop,s_0))\leftarrow 
& holds(noDrops,s_0), \\
& \Not ab(\overline{noDrops},drop ,s_0).\\
holds(\ol{noDrops},res(drop,s_0))\leftarrow 
& holds(\ol{noDrops},s_0),\\
& \Not ab(noDrops, drop,s_0).\\
holds(solidIce,res(drop,s_0))\leftarrow 
& holds(solidIce,s_0), \\
& \Not ab(\overline{solidIce},drop ,s_0).\\
holds(\ol{solidIce},res(drop,s_0))\leftarrow
& holds(\ol{solidIce},s_0), \\
& \Not ab(solidIce, drop,s_0).
\end{array}
\]
The program only has objective formulas. Thus, its semantics is given by its
world view which consists of {\em  belief sets\/} (belief sets are the same as 
answer sets in extended logic programs). 
The world view $W$ of the above program is $W\;=\;\{B_1,\;B_2\}$
{\[\begin{array}{rl}
B_1 = \{ & holds(solidIce, 
res(drop,s_0)),\;ab(solidIce,drop,s_0), \\
& holds(solidIce,s_0),\; holds(\ol{noDrops},res(drop,s_0)), \\
& ab(\overline{noDrops},drop,s_0),\;holds(noDrops,s_0) \\
\} \\
B_2 = \{ &holds(\ol{solidIce}, 
res(drop,s_0)),\;ab(\overline{solidIce},drop,s_0), \\
& holds(solidIce,s_0),\;holds(\ol{noDrops},res(drop,s_0)), \\
& ab(\overline{noDrops},drop,s_0), holds(noDrops,s_0) \\
\}
\end{array}
\]}
Notice that the query $holds(noDrops,res(drop,s_0))$ evaluates to true 
for the above 
belief sets. If we were to ask the queries $holds(solidIce,res(drop,s_0))$
or $holds(\ol{solidIce},res(drop,s_0))$,  we see neither would be able to 
produce an answer of $yes$ or $no$. Both queries' answer is $unknown$.

The recursion through negation provides the desired effect of two
possible interpretations for the effect of $a$ in $f$ (note that  
the two first rules of the example have the form $c \leftarrow \Not b$ and
$b \leftarrow \Not c$ and this program has two answer sets, one 
is $\{c\}$ and the other is $\{b\}$).

The {\em translation\/} of a domain $D$ is defined as the union of the domain
dependent and domain independent rules.

\subsection{General Domains}\label{general domains}

The assumption that we made for simple domains was that for any
sensing action $a_s$ and fluent $f$ there is at most one knowledge law 
of the form 
\begin{equation}\label{simple}
a_s {\bf \;causes\;to\; know}\; f {\bf \;if}\; \precon
\end {equation}
Suppose now we have the following domain
\[
D\left\{
\begin{array}{ll}
r_1 : lookInRoom {\ctk} boardClean {\bf\;if}\; curtainOpen \\
r_2 : lookInRoom {\ctk} boardClean {\bf\;if}\; lightOn \\
\end{array} \right. \] 
and assume we start with the following situation
%
%\[\begin{array}{ll}
%\Sigma = \{ & \{boardClean, curtainOpen\} , \{boardClean,lightOn\}, \\
%&             \{boardClean\}, \{\} \}
%\]
{\small\[\begin{array}{l}
\Sigma = \{ \{boardClean, curtainOpen\} , \{boardClean,lightOn\},
           \{boardClean\}, \{\} \}
\end{array}\]}
There is one model $\Phi_1$ that will result in the states where the
fluent $boarClean$ is true and the knowledge precondition $curtainOpen
\vee lightOn$ of $boardClean$ with respect to $lookInRoom$ is also
true (this corresponds to the third case of compatibility).
{\small\[
\Phi_1(lookInRoom,\Sigma) = 
    \{ \{boardClean, curtainOpen\} , \{boardClean,lightOn\} \}
\]}
We will need a suppression rule similar to the rules in group
(\ref{s1}) of the translation of simple domains.  The rule will be
something like
{\[
\begin{array}{ll}
holds(true,&res(lookInRoom,S)) \\
&\begin{array}{ll}
\leftarrow &
 \neg Kholds(boardClean,S),\neg Kholds(\ol{boardClean},S), 
\nonumber \\
& Kholds(boardClean,res(lookInRoom,S)), \nonumber \\
& \mbox{``}Kholds(curtainOpen \vee lightOn ,res(lookInRoom,S))\mbox{''},
 \nonumber \\
&  holds(\ol{curtainOpen},S), holds(\ol{lightOn},S),
\end{array}
\end{array}
\]}
%%
%{\small\[
%\begin{array}{rl}
%holds(true,res(lookInRoom,S)) \\
%\leftarrow & \neg Kholds(boardClean,S),\neg Kholds(\ol{boardClean},S), 
%\nonumber \\
%& 
%Kholds(boardClean,res(lookInRoom,S)), \nonumber \\
%& \mbox{``}Kholds(curtainOpen \vee lightOn ,res(lookInRoom,S))\mbox{''},
%\nonumber \\
%&  holds(\ol{curtainOpen},S), holds(\ol{lightOn},S),
%\end{array}
%\]}
%
The question is how to encode ``$Kholds(curtainOpen \vee lightOn
,\ldots)$''? We will do it by adding two rules to the
program
\[
\begin{array}{l}
holds(p^{lookInRoom}_{boardClean},S) \leftarrow holds(curtainOpen,S) \\
holds(p^{lookInRoom}_{boardClean},S) \leftarrow holds(lightOn,S) 
\end{array}
\]
Now the disjunction can be replaced by
``$Kholds(p^{lookInRoom}_{boardClean}, res(lookInRoom,S))$''. The
symbol $p^{lookInRoom}_{boardClean}$ is a new constant symbol not
appearing anywhere else in the program. We complete the program with the 
rule
\[
\begin{array}{l}
holds(\ol{p^{lookInRoom}_{boardClean}},S) \leftarrow \Not
holds(p^{lookInRoom}_{boardClean},S) 
\end{array}
\]
and the translation becomes
\[
\begin{array}{ll}
holds(true, &res(lookInRoom,S)) \\
&\begin{array}{ll}
\leftarrow & \neg Kholds(boardClean,S),\neg Kholds(\ol{boardClean},S), 
\nonumber \\
& 
Kholds(boardClean,res(lookInRoom,S)), \nonumber \\
& Kholds(p^{lookInRoom}_{boardClean} ,res(lookInRoom,S)),
\nonumber \\
&  holds(\ol{p^{lookInRoom}_{boardClean}},S)
\end{array}
\end{array}
\]
In general, if $D$ is a domain description (not necessarily simple) then, for any
sensing action $a$ and any fluent $f$, if $\varphi_1 \vee\ldots \vee
\varphi_m$ with $ \varphi_i = p_1^i \wedge \ldots \wedge p^i_{k_i}, i
= 1,\ldots, m$, is the knowledge precondition of $f$ with respect to
$a$ in the domain $D$, we will have a new constant symbol $p_f^a$
in the language of the logic program.  Then for the knowledge
laws: $$
\begin{array}{l} 
 a\;\;{\bf
causes\;to\;know}\; f \;{\bf if} \;\varphi_{1} \\ \vdots \\
a\;\; {\bf
causes\;to\;know}\; f \;{\bf if}\; \varphi_{m}
\end{array}
$$
\noindent 
the domain dependent translation will have the rules:
%%%%%%%%%%%%%%%%%%%%%%%%%%%%%%%%%%%%%%%%%%%%%%%%%
{\small\begin{eqnarray}
holds(p_f^a,S) & \leftarrow & holds(p^1_1,S),\ldots,holds(p^1_{k_1},S)
                              \nonumber \\
               & \vdots \nonumber \\
holds(p_f^a,S) & \leftarrow & holds(p^m_1,S),\ldots,holds(p^m_{k_m},S) \\
holds(\overline{p_f^a},S) & \leftarrow & \Not holds(p_f^a,S) \\
%%%%%%%%%%%%%%%%%%%%%%%case (1) %%%%%%%%%%%%%%%%%%%%%%%%%%%%%%
holds(true,res(a,S)) & \leftarrow & \neg Kholds(f,S),
\neg Kholds(\o{f},S),\nonumber \\ & &
 \neg Kholds(p_f^a,res(a,S)),  \nonumber \\
& & holds(p^a_f,S) \label{2s3}\\
%%%%%%%%%%%%%%%%%%%%%%%%%%%%%%%%%%%%%%%%%%%%%%%%%%%%%%%%%%%%%%%%%%%%
holds(true,res(a,S)) & \leftarrow & \neg Kholds(f,S),\neg Kholds(\o{f},S),
Kholds(\o{f},res(a,S)), \nonumber \\
& & Kholds(p_f^a,res(a,S)),holds(f,S)  \label{2s1}\\ 
%%%%%%%%%%%%%%%%%%%%%%%%%%%%%%%%%%%%%%%%%%%%%%%%%%%%%%%%%%%%%%%%%%%%%%%%%
%%%%%%%%%%%%%%%%%%%%%%%%%%%%%%%%%%%%%%%%%%%%%%%%%%%%%%%%%%%%%%%%%%%%%%%%%
holds(true,res(a,S)) & \leftarrow & \neg Kholds(f,S),\neg Kholds(\o{f},S),
Kholds(\o{f},res(a,S)), \nonumber \\
& & Kholds(p_f^a,res(a,S)), \nonumber \\
& & holds(\overline{p^a_f},S)
 \label{2s4} \\
%%%%%%%%%%%%%%%%%%%%%%%%%%%%%%%%%%%%%%%%%%%%%%%%%%%%%%%%%%%%%%%%%%%%%%%%%%%%%%
holds(true,res(a,S)) & \leftarrow & \neg Kholds(f,S),\neg Kholds(\o{f},S),
Kholds(f,res(a,S)), \nonumber \\
& & Kholds(p_f^a,res(a,S)),holds(\o{f},S) \label{2s2} \\
%%%%%%%%%%%%%%%%%%%%%%%%%%%%%%%%%%%%%%%%%%%%%%%%%%%%%%%%%%%%%%%%%%%%%%%%%%%%%% 
%%%%%%%%%%%%%%%%%%%%%%%%%%%%%%%%%%%%%%%%%%%%%%%%%%%%%%%%%%%%%%%%%%%%%%%%%
holds(true,res(a,S)) & \leftarrow & \neg Kholds(f,S),\neg Kholds(\o{f},S),
Kholds(f,res(a,S)), \nonumber \\
& & Kholds(p_f^a,res(a,S)), \nonumber \\
& & holds(\overline{p^a_f},S)
\label{2s5}
\end{eqnarray}}
The set of rules (\ref{s3}) corresponds to rule (\ref{2s3}). Rule
(\ref{r2}) corresponds to rule (\ref{2s1}).  The set of rules
(\ref{s2}) correspond to rule (\ref{2s4}).  Rule (\ref{r1})
corresponds to rule (\ref{2s2}) and rule (\ref{s1}) corresponds to
rule (\ref{2s5}).

\subsection{Query Translation}

To answer queries in the epistemic logic program we need to include
rules to implement the evaluation functions $\Gamma$. 
The query ``$f \after \alpha$'' will be true in a consistent domain
$D$ if and only if $holds\_after\_plan(f,\alpha)$ is true in the
epistemic logic program obtained from $D$ plus the rules

%%%%%%%%%%%%%%%%%%%%%%%%%%%%%%%%%%%%%%%%%%%%%%%%%%%%%%%%%%
%%%%%%%%%%%%%%%%%%%%%%ORIGINAL%%%%%%%%%%%%%%%%%%%%%%%%%%%%
%{\small$$
%\begin{array}{rll}
%holds\_after\_plan(F,P) & \leftarrow & find\_situation(P,s_0,S), holds(F,S)\\
%find\_situation([],S,S)&\leftarrow \\ 
%find\_situation([a|\alpha],S,S_1)&\leftarrow
%& find\_situation(\alpha,res(a,S),S_1)\\
%find\_situation([\If \varphi\;\Then \alpha_1|\alpha_2],S,S_1) &\leftarrow &
%Kholds(\ol{\varphi},S),find\_situation(\alpha_2,S,S_1)\\
%find\_situation([\If \varphi\;\Then \alpha_1|\alpha_2],S,S_1)& \leftarrow &
%Kholds(\varphi,S),find\_situation(\alpha_1,S,S'),\\
%               & & find\_situation(\alpha_2,S',S_1)\\
%find\_situation([\If \varphi \; \Then \alpha_1 \; \Else \;
%\alpha_1'|\alpha_2],\\S,S_1)& \leftarrow &
%Kholds(\o{\varphi},S),find\_situation(\alpha_1',S,S'),\\
%                   & & find\_situation(\alpha_2,S',S_1)\\
%find\_situation([\If \varphi \; \Then \alpha_1\; \Else \;
%\alpha_1'|\alpha_2],\\S,S_1) &\leftarrow &
%Kholds(\varphi,S),find\_situation(\alpha_1,S,S'),\\
%               & & find\_situation(\alpha_2,S',S_1)\\
%find\_situation([\While\varphi\; \Do \alpha_1|\alpha_2],S,S_1) &\leftarrow 
%&
%Kholds(\o{\varphi},S),\\
%& &
%find\_situation(\alpha_2,S,S_1)\\
%find\_situation([\While \varphi \; \Do \alpha_1|\alpha_2],S,S_1)
%& \leftarrow &
%Kholds(\varphi,S),\\& & find\_situation(\alpha_1,S,S'),\\& &
%find\_situation([\While\varphi\;\Do \alpha_1|\alpha_2],S',S_1)
%\end{array}
%$$}
%%%%%%%%%%%%%%%%%%%%%%%%%%%%%%%%%%%%%%%%%%%%%%%%%%%%%%%%%
%%%%%%%%%%%%%%%%%%%%%%%%%%%%%%%%%%%%%%%%%%%%%%%%%%%%%%%%%
{\small$$
\begin{array}{rl}
holds\_after\_plan(F,P) \\ \leftarrow & find\_situation(P,s_0,S), holds(F,S)\\
find\_situation([],S,S)\\\leftarrow \\ 
find\_situation([a|\alpha],S,S_1) \\ \leftarrow
& find\_situation(\alpha,res(a,S),S_1)\\
find\_situation([\If \varphi\;\Then \alpha_1|\alpha_2],S,S_1) \\\leftarrow &
Kholds(\ol{\varphi},S),find\_situation(\alpha_2,S,S_1)\\
find\_situation([\If \varphi\;\Then \alpha_1|\alpha_2],S,S_1) \\\leftarrow &
Kholds(\varphi,S),find\_situation(\alpha_1,S,S'),\\
               & find\_situation(\alpha_2,S',S_1)\\
find\_situation([\If \varphi \; \Then \alpha_1 \; \Else \;
\alpha_1'|\alpha_2],\\S,S_1)\\ \leftarrow &
Kholds(\o{\varphi},S),find\_situation(\alpha_1',S,S'),\\
                   &  find\_situation(\alpha_2,S',S_1)\\
find\_situation([\If \varphi \; \Then \alpha_1\; \Else \;
\alpha_1'|\alpha_2],\\S,S_1) \\\leftarrow &
Kholds(\varphi,S),find\_situation(\alpha_1,S,S'),\\
               & find\_situation(\alpha_2,S',S_1)\\
find\_situation([\While\varphi\; \Do \alpha_1|\alpha_2],S,S_1) \\ \leftarrow 
&
Kholds(\o{\varphi},S),\\
& 
find\_situation(\alpha_2,S,S_1)
\end{array}$$
$$
\begin{array}{rl}
find\_situation([\mbox{\bf while}\;\varphi \;\mbox{\bf do}\;\alpha_1|\alpha_2],S,S_1)
\\ \leftarrow &
Kholds(\varphi,S),\\ & find\_situation(\alpha_1,S,S'),\\ &
find\_situation([\mbox{\bf while}\;\varphi\;\mbox{\bf do}\;\alpha_1|\alpha_2],S',S_1)
\end{array}
$$}
%%%%%%%%%%%%%%%%%%%%%%%%%%%%%%%%%%%%%%%%%%%%%%%%%%%%%%%%%
%%%%%%%%%%%%%%%%%%%%%%%%%%%%%%%%%%%%%%%%%%%%%%%%%%%%%%%%%
As you may note from the rules, $holds\_after\_plan(F,P)$ works
in two steps.  First, it finds the situation $s$ that results from
applying $P$ to the initial situation (using
$find\_situation(P,s_0,S)$) and then shows that $F$ holds in that
situation.  Since the translation of the domain may have several world
views the program needs to find a situation for each world view.  The
following example illustrates how the process works. 
\begin{example}~\\
\[D^1_{1}\left\{
\begin{array}{ll}
r_1 : {\bf \;initially}\; \neg burnOut \\
r_2 : {\bf \;initially}\; \neg bulbFixed \\
r_3 : changeBulb {\bf \;causes}\; burnOut {\bf \;if}\; switchOn \\
r_4 : changeBulb {\bf \;causes}\; bulbFixed {\bf \;if}\; \neg switchOn \\
r_5 : turnSwitch {\bf \;causes}\; switchOn {\bf\;if}\; \neg switchOn \\
r_6 : turnSwitch {\bf \;causes}\; \neg switchOn {\bf\;if}\; switchOn \\
r_7 : checkSwitch {\bf\;causes\;to\;know}\;switchOn{\bf\;if}\;\neg burnOut \\
\end{array} \right. \]
\end{example}
Assume we would like to show that
{\small\[
\begin{array}{ll}
D^1_1 \models bulbFixed \after [ & checkSwitch,
\\                               & \If switchOn \Then [turnSwitch], 
\\                               & changeBulb]
\end{array}
\]}
The states in the initial situation of this example are:
{\[
\begin{array}{ll}
\Sigma = \{ &
\{\neg burnOut, \neg bulbFixed, switchOn\}, \\ &
\{\neg burnOut, \neg bulbFixed, \neg switchOn\}\} 
\end{array}
\]}
% in Example~\ref{ex-d01}. 
 It has two models $\Phi_1$ and $\Phi_2$ that
for the sensing action $checkSwitch$ behave very much like in
Example~\ref{ex-d01}.  Then, the logic program translation of this
domain has two world views.  The world view $\cal W$ corresponding to
$\Phi_1$ has two belief sets $W^1$ and $W^2$, such that $W^1_{s_0}
\cup W^1_{res(cs,s_0)}$ from Example \ref{ex-d01} is a subset of $W^1$
and $W^2_{s_0}\cup W^2_{res(cs,s_0)}$ is a subset of $W^2$.  $W^1$
also contains the sets

$\begin{array}{rl}
W^1_{res(ts,res(cs,s_0))}& = \\  &\{ 
 holds(\overline{burnOut},res(turnSwitch,res(checkSwitch,s_0))),\\&
 holds(\overline{bulbFixed},res(turnSwitch,res(checkSwitch,s_0)), \\&
 holds(\ol{switchOn},res(turnSwitch,res(checkSwitch,s_0)))\}
\end{array}$

\noindent
and

$W^1_{res(cb,res(ts,res(cs,s_0)))} =$
  
$\begin{array}{l}
\{ holds(\overline{burnOut},res(changeBulb,res(turnSwitch,res(checkSwitch,s_0)))),
\\
holds(bulbFixed,res(changeBulb,res(turnSwitch,res(checkSwitch,s_0)))),\\ 
holds(\ol{switchOn},res(changeBulb,res(turnSwitch,res(checkSwitch,s_0))))\}
\end{array}$

\noindent
and $W^2$ the set

$$
\begin{array}{ll}
W^2_{res(ts,(res(cs,s_0)))} = &\{ 
holds(\overline{burnOut},res(turnSwitch,res(checkSwitch,s_0))),\\& 
holds(\overline{bulbFixed},res(turnSwitch,res(checkSwitch,s_0))), \\& 
holds(\ol{switchOn},res(turnSwitch,res(checkSwitch,s_0)))\} \\
& \bigcup \\& 
\{ holds(burnOut,res(turnSwitch,res(checkSwitch,s_0))),\\ &
holds(bulbFixed,res(turnSwitch,res(checkSwitch,s_0))), \\ &
holds(switchOn,res(turnSwitch,res(checkSwitch,s_0))), \\ &
holds(true,res(turnSwitch,res(checkSwitch,s_0)))\}
\end{array}
$$ 

\noindent
There is also a similar set $W^2_{res(cb,res(ts,(res(cs,s_0))))}$, with
the same elements of $W^2_{res(ts,(res(cs,s_0)))}$, replacing the
situation argument with
$$res(changeBulb,res(turnSwitch,(res(checkSwitch,s_0))))$$
This
corresponds to the sequence 
$$\begin{array}{l}
\Sigma,\\
\Phi_1(checkSwitch,\Sigma),\\
\Phi_1(turnSwitch,\Phi_1(checkSwitch,\Sigma)),\\
\Phi_1(changeBulb, \Phi_1(turnSwitch,\Phi_1(checkSwitch,\Sigma)))
\end{array}$$
Thus, in this world view the predicate 
\[\begin{array}{ll}
find\_situation([& checkSwitch,
\\&  \If switchOn\; \Then [turnSwitch],
\\ & changeBulb], s_0, S)
\end{array}
\]
will hold in $\cal W$ iff $S =
res(changeBulb,res(turnSwitch,res(checkSwitch,s_0)))$. The second step
will check if
$$holds(bulbFixed,res(changeBulb,res(turnSwitch,res(checkSwitch,s_0)))$$
is in $\cal W$.  The answer is yes since the atom belongs to both $W^1$
and $W^2$.

The world view associated with $\Phi_2$ is defined in a similar
manner, but in this world view $S = res(changeBulb,res(checkSwitch,s_0))$.

\medskip
Let $\Pi_D$ be the epistemic logic program corresponding to the
translation of a domain description $D$, and denote by $\Pi_D^Q$ the
union of $\Pi_D$ and the rules to interpret queries given above.
Then we can show:

\begin{theorem} \label{entailmentRel}
Given a consistent domain description $D$ and a plan $\beta$.
$D\models F \after \beta$ iff $\Pi_D^Q \models
holds\_after\_plan(F,\beta)$.
\end{theorem}
{\bf Proof:} see appendix.

\section{Relation to Other Work} \label{sec:other-work}
In \cite{lev:PlanSens} there is also a programming language based on
the situation calculus which uses sensing actions.  This work is based
on previous work from \cite{sl:KPA}, in which knowledge is represented
using two levels.  There is a representation of the actual situation
(called $s$) in which the agent is in, and there are situations
accessible from $s$ (called $s'$) which the agent thinks it might be
in.  Something is known to the agent as being true (false) if it is
true (false) in all situations $s'$ which are accessible from the
actual situation $s$ and is unknown otherwise.  In other words
\cite{sl:KPA} distinguishes between what is known by the agent and
what is true in world.  We only represent what is know by the agent,
and assume that this knowledge might be incomplete but always
correct. Something is known in our representation if its value is the
same throughout the states in a situation and unknown otherwise.

In \cite{lev:PlanSens,sl:KPA} the authors use preconditions which are
executability conditions for an action's execution. For example, a
precondition to clean a white-board is one must be in front of the
white-board.  Our preconditions differ in that they are conditions on
the effects.  We can always execute an action but its effect varies
according to its precondition in the effect propositions.  Extending
$\Ak$ to include executability conditions can be done as for
extensions of $\cal A$.  The use of conditions on effects however allows
us to represent a phenomenon of sensing in which the value of
previously unknown preconditions are learned along with the fluent we
are trying to gain knowledge about. This is shown in examples
\ref{ex:door} and \ref{ex:room1} where the robot will know whether or
not it is facing the door after executing the action $look$.

In \cite{lev:PlanSens} once knowledge is
gained it is never lost. We, on the other hand, explore the use of
non-deterministic actions as a mechanism to remove knowledge.
%Our use of non deterministic actions is 
Our use of non-deterministic actions is similar
to \cite{th:Raelp} where the effect of a non-deterministic action is
to make a fluent true or false, but exactly which is indeterminate.
As might be expected, there are cases where the possible outcome
is not intuitive. 
Take for example a deterministic action $Shoot$ that causes
Ollie to be dead. Any observation, which depends on Ollie being alive,
such as ``Ollie is walking'' can be made false using the same action
Shoot. Shoot can be used as a restriction which causes Ollie not to
walk. In the resulting situation, Ollie will not be alive and
therefore will not be walking around. This is not the case when Shoot
has the non-deterministic effect of making Ollie dead or leaving Ollie
alive (suppose that the gun is not working well).  With the same
restriction, Ollie may be dead and not walking in one situation and
alive and not walking in the other. Assuming one can walk as long as
one is alive, then the later situation makes no sense. The same holds
true without the restriction but this time Ollie will be dead and
walking in one situation. If shoot also has a non-deterministic effect
on walking, we are no better off. 

%Take for example a situation where the result of 
%shooting is non-deterministic. The target of this action Ollie is 
%alive and walking around. After the action of shooting, Ollie may 
%still be alive or may be dead. In the situation where Ollie is still alive,
%inertia will cause walking to be true in the resulting state.  
%In the situation where Ollie has died, walking will still be true by
%inertia in the resulting state. 
These cases are prevented with integrity constraints as in
\cite{kl:Aie}.  Our language could be
extended to include constraints as in \cite{bgp:RepAct} 
%our focus
%in this paper is on the effect non-determinism has on knowledge. 
but our interest in non-determinism is its effect on
knowledge. We discuss the topic of integrity constraints in Section
\ref{sec:future}.

Most translations for dialects of ${\cal A}$ are to extended logic
programs. Our translation is to epistemic logic programs because of
its ability to represent knowledge and incomplete information.  To the
best of our knowledge this is the first use of epistemic logic
programs in a translation from action languages.  The closest work
related to our results is presented in \cite{bts:sensing}.
In that paper $\cal A$ is also extended to handle sensing
actions but the semantics is some what limited because they work with
a three value semantics and only approximate knowledge.  Furthermore,
in their language sensing actions have no conditional effects.  These
restrictions allow Baral and Son to write translations into extended
logic programs.  Showing whether is possible to find a translation
into extended logic programs or first order logic of our domains is an
open question.

%%%%%%%%%%%%%%%%%%%%%%%%%%%%%%%%%%%%%%%%%%%%%%%%%%%%%%%%%%%%%%%%%%%%
%%%%% Future Work %%%%%%%%%%%%%%%%%%%%%%%%%%%%%%%%%%%%%%%%%%%%%%%%%%
%%%%%%%%%%%%%%%%%%%%%%%%%%%%%%%%%%%%%%%%%%%%%%%%%%%%%%%%%%%%%%%%%%%%

\section{Future Work} \label{sec:future}

We already mentioned the need to clarify the complexity of adding
sensing actions to domain descriptions.  Our translation suggests that
it might be computationally more complex to deal with conditional
sensing actions than sensing with no conditions.
Two other possible directions of research are: First, the ability of
an agent to query itself about what it knows (i.e introspection).
This is useful when the cost of executing a series of plans is
expensive maybe in terms of time. Allowing an agent to query whether
it knows that it knows something may be a cheaper alternative and cost
effective.  The use of a modal operator as shown below may be
sufficient to accomplish this.

%\begin{equation} \label{introspec}
%D \models {\bf knows}(\varphi )\after [\alpha ] 
%\end{equation}

\begin{equation} \label{introspec}
 \If \neg {\bf knows}(\varphi ) \;{\bf then} \;[\alpha] 
\end{equation}

\noindent where $\varphi$ is a test condition (as defined in Section
\ref{sec:queries}), $\alpha$ a plan, and ${\bf knows}(\varphi)$ would
be an introspective operator on the test condition.

Take for example Agent A in Fig. 1 from Section \ref{sec:domain}. Agent A
knows that Ollie is wet (denoted by $wet$), but does not know if it 
is raining outside (denoted by $rain$). Agent A would have to find a window
and then look out that window to see if it is raining outside. Suppose the 
program or control module for finding a window in a building is long
and very costly as far as battery power, Agent A would have to find a 
window and then check for rain. Agent B would benefit from the 
conditional below

\medskip
$~~~\If$ $\neg {\bf knows}(\varphi ) \;{\bf then} \;[findWindow, lookOutside]$ 

\medskip \noindent
Without an introspective operator, both Agent A and Agent B would have 
to find a window and then look outside. Agent B can save on
battery power if it has the ability to query itself on what it knows. 

%Further investigation into this is definite.
Second, we could investigate expanding the initial epistemic state.
At present, domains only may start from a situation with only one
initial epistemic state.  For more states or to represent multiple
initial situations in a domain, the language to describe domains must be
extended with modal operators.
%It is quite valid to assume that this is not always the case. 

As mentioned earlier in this paper, integrity constraints could be
added. Integrity constraints define dependency relationships between fluents. 
Taking the example from the previous section, walking depends on Ollie
being alive. This could be represented following the approach outlined in
\cite{kl:Aie,lif:TwoComp} 

\begin{equation} \label{eq:integ}
{\bf never}\; \varphi \If \psi
\end{equation}

\noindent
where $\varphi$ and $\psi$ are conjunctions of fluent literals. 
It states that $\varphi$ can not be true when $\psi$ is true. Our 
example with Ollie would look like this

\medskip
$~~~{\bf never}\; walking \If \neg alive $ 

\medskip \noindent
Conditions of effect are used throughout this paper. An example of
a executability condition is the fact that one has to be at a
light bulb to change the light bulb. 
Executability 
conditions found in \cite{lev:PlanSens} could be implemented using
the methods found in \cite{kl:Aie,lif:TwoComp}.

\begin{equation} \label{eq:execond}
{\bf impossible} \;A \If \psi
\end{equation}

\noindent where $A$ is an action and $\psi$ is a conjunction of fluent
literals. The execution of action $A$ cannot take place as long as
$\psi$ is true.

\noindent
Using (\ref{eq:execond}) we could express the constraint that in order to
change the bulb, one has to be at the lamp as

\medskip
$~~~{\bf impossible} \;changeBulb \If \neg atLamp$ 

\medskip 

This paper will conclude with three thoughts. One is relaxing the
assumption that an agent has incomplete but always correct knowledge
of its world. One could imagine the agent not only reasoning on
information that it knows is true, but also reasoning on what it
believes is true. At present we have not explored this topic.  The
other idea is given that $\Ak$ is a high level action description
language that deals with incomplete information across multiple
possible worlds, it stands to reason that $\Ak$ could be translated to
formalisms, such as Levesque's \cite{lev:PlanSens,sl:KPA} or
autoepistemic logic \cite{mo:ScNmL,mt:AeL}, which also hold this
property. The third refers to the termination of routines; there are
certain tasks for which routines can be limited by a sensing action
that determines the ``size'' of the problem.  Take for example the
number of pages in a book or the number of doors on the second floor
of an office building.  The number of pages contained in a book will
ensure the termination of a search for a word through that book. The
same applies to the number of doors on the second floor with respect
to a security routine which checks that all the doors on the second
floor are locked.  For this type of task, a counter is sufficient.  To
include counters, we do not require constraints but variables in
$\Ak$.  These loops correspond to for-loops in regular programming
languages.  Consider the situation described in \ref{ex:cans}.  On the
floor of a room there are cans. An agent is given an empty bag and
instructed to fill the bag with cans. We assume that there are more
than enough cans on the floor to fill the bag.  We can model the space
left in the bag by having initially true one (and only one) of the
following fluents.
$$
\begin{array}{l}
spaceLeft(0)\\
spaceLeft(s(0))\\
\vdots\\
spaceLeft(s^n(0))\\ \vdots
\end{array}
$$
\noindent
The effect of $drop$ can be now described by the effect proposition:
%\begin{example}
% \label{ex:cans}

\[
%D^1\left\{
\begin{array}{ll}
%r_1 : {\bf \;initially}\; \neg bagFull \\
r_1 : drop {\bf \;causes}\; spaceLeft(x){\bf\;if}\; spaceLeft(s(x))\\
%r_2 : drop {\bf \;causes}\;\neg spaceLeft(s(x)) 
%{\bf\;if}\; spaceLeft(s(x) \\
%r_4 : lookInBag \;\ctk \;  bagFull \\
%r_5 : pickUp {\bf \;causes}\; canInHand {\bf\;if}\;  \neg canInHand \\
\end{array}
% \right. 
\] 
%\end{example}
However, we need to restrict the world to only allow one $spaceLeft$
fluent to be true at any moment.  This can be described with a
constraint of the form
\[
\mbox{\bf never}\;  spaceLeft(x) \wedge  spaceLeft(y) \If x\neq y
\]
Note that the constraint encodes a ramification of $drop$ since not
only the execution of the action $drop$ makes $spaceLeft(x)$ true, but
also indirectly causes $spaceLeft(s(x))$ to become false.

An orthogonal problem to the issue of constraints, is that we still
need in our domain a value proposition that tells us how much space we
initially have in the bag.  Adding the initial value proposition is
not a completely satisfactory solution since the plan
\[
\mbox{\bf while}\; nospaceLeft(0) \Do [dropCanInBag]
\]
\noindent
fills the bag irrespectively of the initial situation and (in normal
circumstances) the plan will always terminate. Furthermore, in a
realistic setting, plans need to consider limitation of resources.
Plans may need to limit the amount of time devoted to any task or
limit the amount of energy that can be used. These bounds can be
applied to all tasks, but still a counter is required.  Further
research in termination, specially in a common-sense approach to proof
of termination is necessary to deal with loops in plans.

%%%%%%%%%%%%%%%%%%%%%%%%%%%%%%%%%%%%%%%%%%%%%%%%%%%%%%%%%%%%%%%%%%%%
%%%%% Conclusion %%%%%%%%%%%%%%%%%%%%%%%%%%%%%%%%%%%%%%%%%%%%%%%%%%%
%%%%%%%%%%%%%%%%%%%%%%%%%%%%%%%%%%%%%%%%%%%%%%%%%%%%%%%%%%%%%%%%%%%%

%\COMMENT
%##################The beginning############################
%
%####################The End###############################
%\ENDCOMMENT

%%%%%%%%%%%%%%%%%%%%%%%%%%%%%%%%%%%%%%%%%%%%%%%%%%%%%%%%%%%%%%%%%%%%
%%%%% Bibliography %%%%%%%%%%%%%%%%%%%%%%%%%%%%%%%%%%%%%%%%%%%%%%%%%
%%%%%%%%%%%%%%%%%%%%%%%%%%%%%%%%%%%%%%%%%%%%%%%%%%%%%%%%%%%%%%%%%%%%
%\singlespace

%%%%%%%%%%%%%%%%%%%%%%%%%&&&&&&&&&&&%%%%%%%%%%%%%%%%%%%%%%%
%%%%%%%%%%%%%%%%%%%%%%%%%&&&&&&&&&&&%%%%%%%%%%%%%%%%%%%%%%%
%%%%%%%%%%%%%%%%%%%%%%%%%&&&&&&&&&&&%%%%%%%%%%%%%%%%%%%%%%%

\paragraph{Acknowledgments}  We would like to thank the anonymous
referees for their invaluable suggestions.  Section~\ref{general
domains} is a direct result of one of their comments.

\bibliographystyle{tlp}

%\bibliography{stuart}
%\bibliography{/home/jorge/research/actions/know-act/stuart}

%\end{document}

\appendix

\section{While-evaluation Function}
\label{denotational sem}

\begin{definition}
  Let $\E$ be the set of all situations and $\P$ the set of all total
  functions $f$  mapping situations into situations, ${\P} = \{f \;|\;
  f:\E \rightarrow \E$ \}.  We say that for any pair of functions
  $f_1,f_2 \in \P, f_1 \leq f_2$ if and only if for any $\Sigma \in
  \E$ if $f_1(\Sigma) \neq \emptyset$, then $f_1(\Sigma) =
  f_2(\Sigma)$.
\end{definition}   
The next proposition follows from the above definition.
\begin{proposition}
The above relation $\leq$ defines a partial order in $\P$.
\end{proposition}
Moreover, this partial order is a complete semi-lattice with the bottom
element equal to the function that maps every situation to
$\emptyset$.  We will denote the bottom element by $f_{\emptyset}$.

\begin{definition}
  Let $\alpha$ be a plan and $\Gamma$ a function that maps plans and
  situations into situations. Let $\varphi$ be a conjunction of fluent 
literals.
  Then, we define the function $\Fap:\P \rightarrow \P$ such that for
  any function $f \in \P,$ 

  $\Fap(f)(\Sigma) = \left\{
  \begin{array}{ll}
    \ \Sigma & \mbox{if $\varphi$ is false in $\Sigma$} \\
  f(\Gamma(\alpha,\Sigma)) & \mbox{if
  $\varphi$ is true in $\Sigma$ }\\
  \emptyset & \mbox{otherwise}
  \end{array}\right.$
\end{definition}

Our goal is to show that $\Fap$ is continuous. For this, we will need
to show that for any directed set $D\subseteq \P$, the least upper
bound of $D$, denoted by $\bigsqcup D$ exists, and that $\Fap(\bigsqcup D)
= \bigsqcup\{\Fap(d)\;|\; d\in D\}$. A directed set is a set such that for
any finite subset of it, the least upper bound of that set exists, and
belongs to the directed set.  The existence of $\bigsqcup D$
follows from the following proposition.
\begin{proposition} 
\label{directed} Let $D$ be a directed subset of $\P$, and let
  $d\in D$. If $d(\Sigma) = \Sigma^{'}\neq \emptyset$, for a situation
  $\Sigma$, then for any $d^{'}\in D$ either $d^{'}(\Sigma)=\emptyset$
  or $d^{'}(\Sigma)=\Sigma^{'}$.
\end{proposition}
It follows from this proposition that,
$$\bigsqcup D(\Sigma) = 
\left
\{\begin{array}{ll}
\emptyset & \mbox{if $\forall d\in D, d(\Sigma) = \emptyset$}\\
\Sigma^{'} & \mbox{if $\exists d\in D$ such that $d(\Sigma) = \Sigma^{'}$ and
$\Sigma\neq \emptyset$}
\end{array}\right.$$
A similar function is defined by $\bigsqcup\{\Fap(d)\;|\; d\in
D\}$. This function will be used in the proof of the following
theorem.
\begin{theorem}
For any plan $\alpha$ and any conjunction of fluent literals $\varphi$, the
function $\Fap$ is continuous with respect to the order
$\leq$.
\end{theorem}

\noindent{\bf Proof:}
Let $\bigsqcup \Fap[D]$ denote the function $\bigsqcup\{\Fap(d)\;|\; d\in
D\}$. To prove the theorem, it suffices to show that, for any
directed set $D \subseteq \P$, $\Fap(\bigsqcup D) = \bigsqcup \Fap[D]$.\\
\noindent Let $\Sigma$ be a situation. \\
\noindent {\bf (a)} If $\varphi$ is false in
$\Sigma$ then for any $f \in \P$, $\Fap(f)(\Sigma) = \Sigma$. Hence,

\[ \Fap(\bigsqcup D)(\Sigma) = \Sigma = \bigsqcup\Fap[D](\Sigma).\]
\medskip \noindent {\bf (b)} If $\varphi$ is true in $\Sigma$, then
$\Fap(\bigsqcup D)(\Sigma) = \bigsqcup D(\Gamma(\alpha,\Sigma))$.  Let
$\bigsqcup\Fap[D](\Sigma) = \Sigma^{'}$. By
Proposition~\ref{directed}, $\Sigma^{'} = \emptyset$ iff
$\Fap(d)(\Sigma) = \emptyset$, for any $d\in D$ since $\Fap(d)(\Sigma) =
d(\Gamma(\alpha,\Sigma))$ and $D$ is directed. Therefore,
$\Fap(\bigsqcup D)(\Sigma)$ must be $\emptyset$. If
$\Sigma^{'}\neq\emptyset$, then for every $d\in D$ such that
$\Fap(d)(\Sigma)\neq\emptyset$, it must be the case that
$\Fap(d)(\Sigma) = \Sigma^{'}$ since $\Fap(d)(\Sigma) =
d(\Gamma(\alpha,\Sigma))$ and $D$ is directed. Then
$\Fap(\bigsqcup D)(\Sigma) = \Sigma^{'}$.

\medskip \noindent
{\bf (c)} When $\varphi$ is neither true nor false in $\Sigma$, the
proof is similar to part (a) since $\Fap(f)(\Sigma) = \emptyset$ for
any $f\in\P$.
\qed

\noindent
We define the powers of $\Fap$ as follows:
\begin{enumerate}
\item $\Fap\uparrow 0 = f_{\emptyset}$.
\item $\Fap\uparrow n+1 = \Fap(\Fap\uparrow n)$.
\item $\Fap\uparrow\omega = \bigsqcup\{\Fap\uparrow n \;|\; n\leq
\omega\}$.
\end{enumerate}
From the continuity of $\Fap$ the corollary below follows.
\begin{corollary} \label{fixpoint}
The least fix-point of $\Fap$ is $\Fap\uparrow\omega$.
\end{corollary}
%

%%%%%%%%%%%%%%%%%%%%%%%%%%%%%%%%%%%%%%%%%%%%%%%%%%%%%%%%%%%%%%%%%%%%
%%%%%%%%%%%%%%%%%%%%%%%%%%%%%%%%%%%%%%%%%%%%%%%%%%%%%%%%%%%%%%%%%%%%
%%%%%%%%%%%%%%%%%%%%%%%%%%%%%%%%%%%%%%%%%%%%%%%%%%%%%%%%%%%%%%%%%%%%

\section{Proofs}
 
In this section we present the proof of Theorem \ref{entailmentRel} by
givingn a detailed proof of the correctness of the translation for
simple domains.  The proof for the general case is a direct extension.
In our proofs we will use the splitting lemma of extending logic
programs \cite{lt:Split}.  For completeness we will include some 
definitions and the
statement of the lemma below.

%The following definitions are from \cite{lt:Split}.  
Consider a
nonempty set of symbols called atoms.  A literal is an atom possibly
preceded by the classical negation symbol $\neg$. A rule is determined
by three finite set of literals - the set of head literals , the set
of positive subgoals and the set of negated subgoals. The rule with
the head literals $L_1,
\ldots, L_q$, the positive subgoals $L_{i+1}, \ldots, L_m$ and the
negated subgoals $L_{m+1}, \ldots, L_n$ is written as 

$\;\;\;\;\;\;\;\;\;\;\;L_1\;or\; \ldots\;or\; L_q \leftarrow L_{q+1}, \ldots 
, L_m, \Not L_{m+1}, \ldots, L_n$

The three parts of a rule $r$ are denoted by $head(r)$, $pos(r)$ and 
$neg(r)$; $lit(r)$ stands for $head(r) \cup pos(r) \cup neg(r)$.  
\begin{definition}{\it (Splitting set)} \cite{lt:Split}
A splitting set for a logic program $\Pi$ is any set $U$ of literals 
such that, for every rule $r \in \Pi$, if $head(r) \cap U \not = \emptyset$ 
then $lit(r) \subseteq U$. If $U$ is a splitting set for $\Pi$, 
we also say that $U$ splits $\Pi$. The set of rules $r \in \Pi$ such that 
$lit(r) \subseteq U$ is called the {\it bottom} of $\Pi$ 
relative to the splitting set $U$ and is denoted by $b_U(\Pi)$. 
The subprogram $\Pi - b_U(\Pi)$ is called the {\it top} of 
$\Pi$ relative to $U$.   
\end{definition}

\begin{definition}{\it(Partial evaluation)} \cite{lt:Split}
The partial evaluation of a program $\Pi$ with splitting set $U$
 w.r.t. a set of literals $X$ is the program $e_U(\Pi,X)$ defined as follows.
 For each rule $r \in \Pi$ such that: \\
$(pos(r) \cap U) \subseteq X \wedge (neg(r) \cap U) \cap X = \emptyset$
\\
put in $e_u(\Pi,X)$ the rule $r'$ which satisfies the following property:\\
$head(r') = head(r), pos(r') = pos(r) - U, neg(r') = neg(r) - U$.
\end{definition}

\begin{definition}{\it (Solution)} \cite{lt:Split}
Let $U$ be a splitting set for a program $\Pi$. A solution to 
$\Pi$ w.r.t. $U$ is a pair $(X,Y)$ of sets of literals such that:

 \begin{itemize}
\item $X$ is an answer set for for $b_U(\Pi)$;
\item $Y$ is an answer set for $e_U(\Pi - b_U(\Pi),X)$;
\item $X \cup Y$ is consistent.
\end{itemize}

\end{definition}

\begin{lemma}{\it (Splitting lemma)} \cite{lt:Split} 
Let $U$ be a splitting set for a program $\Pi$. A set $A$ of literals 
is a consistent answer set of $\Pi$ if and only if $A = X \cup Y$ 
for some solution $(X,Y)$ to $\Pi$ w.r.t. U.

\end{lemma}

 From now on we will refer to the {\em simple domain description\/}
 \ref{simple}, as {\em domain description\/} to simplify the statements.

The proof of Theorem \ref{entailmentRel}
({\bf Theorem:}
 %\label{entailmentRel}
Given a consistent domain description $D$ and a plan $\beta$.
$D\models F \after \beta$ iff $\Pi_D^Q \models
holds\_after\_plan(F,\beta)$.)
 is organized as follows:
\begin{enumerate}
\item 
First, we will prove that the epistemic logic program translation
models correctly the execution of a single non-sensing action.
Intuitively this can be done by looking at all the predicates
of the form $holds(f,res(a,s_0))$, for any non-sensing action $a$.
Furthermore we should be able to replace the initial constant $s_0$
with any fixed situation constant $s$ of the form
$res(a_1,\ldots,res(a_k,s_0)\ldots)$.  In the proof we will show that
given any situation constant $s$, state $\sigma$, and 0-model of the
domain $\Phi_0$, we can find a sub-set of the program $ground(\Pi_D)$
in which assuming $s$ to be the initial situation constant one of its
belief sets corresponds to $\Phi_0$. We will also prove the other
direction.  That is, for any belief set of the mentioned sub-set of
$ground(\Pi_D)$, there exists a corresponding function $\Phi_0$,
0-model of $D$.  This covers the general case of a single non-sensing
action applied to a situation since, by the definition of
\ref{d-model}, this reduces to the application of 0-interpretations to
each of the states in the situation.

%given a domain description $D$, without value propositions or
%knowledge laws, we associate with any pair of 0-interpretation and
%state $(\Phi_0,\sigma)$ and any situation constant $s$, a set of
%literals $ A_{(\Phi_0,\sigma,s)}$ such that it is a belief set of
%$\Pi^1_{(D,s)}$, if $(\Phi_0,\sigma)$ is a 0-specific model of $D$,
%and we prove that any belief set of $\Pi^1_{(D,s)}$ has such a form
%for some 0-specific-model $(\Phi_0,\sigma)$ of $D$.

\item
The second part of the proof extends the first part to cover the
execution of sensing actions.  In this case the
sub-set of $ground(\Pi_D)$ includes rules with the modal operator
$K$.  We  show that each world view of the sub-program corresponds to
an interpretation $\Phi$, model of $D$.  We also show that for any model
$\Phi$ of $D$ there is an associated world view of the sub-program.

\item
The next step extends step 2 from the application of a single action
to the application of any sequence of actions by induction.

\item 
The final step extends the proof from sequence a of actions to complex
plans.  The proof shows by structural induction on the complexity of
the plans that given a fixed world view any plan (that terminates) can
be reduced to the execution of a sequence of actions.

\end{enumerate}

Given a situation constant $s$, denote by $\Pi^1_{(D,s)}$ the
subprogram of $Ground(\Pi_D)$ that is restricted to those rules in
$ground(\Pi_D)$, such that either the only situation constant
appearing in the heads is of the form $res(a,s)$ for an action symbol
$a$, or is of the form $ab(f,a,s)$ for a fluent literal $f$ and action
symbol $a$.
 
For any possible action $a$, we will denote by $\Pi^1_{(D,a,s)}$
the subprogram of $\Pi^1_{(D,s)}$ that is restricted to those rules in
$\Pi_D$ that only involve the action $a$ in its predicates, besides
other action symbols occurring in $s$.
% and are not translation of $\initially$ propositions.
We call a domain description a {\em universal\/} domain if there are no
value propositions in the domain.  Given a universal domain
description $D$, and state $\sigma$, we denote by $D_\sigma$ the
domain consisting of $D \cup \{\initially f: f\in \sigma\} \cup
\{\initially\neg f: f \not\in \sigma\}$. 
% We will denote by $D^{n.s}$
%the set of value and effect propositions in $D$, and by $D^{sen}$ the
%set of knowledge laws in $D$.
%Observe that when a domain description $D$ has no knowledge laws,
%$\Pi_D$ is a normal logic program.

%
\begin{definition}\label{A1}  For a domain description $D$,
let $\sigma$ be a state, $s$ a situation constant, and $\Phi_0$ a
0-interpretation, We define the set of literals
$A_{(\Phi_0,\sigma,s)}$ as follows:\\

\noindent
For any action $a$ and any fluent $f$,
 
\begin{enumerate}
  
\item $holds(f,s) \in A_{(\Phi_0,\sigma,s)} \iff f \in \sigma $,

\item $holds(\o{f},s) \in A_{(\Phi_0,\sigma,s)} \iff f \not \in
  \sigma$,

\item $holds(f, res(a,s)) \in A_{(\Phi_0,\sigma,s)} \iff f \in
  \Phi_0(a,\sigma) $,

\item $ holds(\o{f},res(a,s)) \in A_{(\Phi_0,\sigma,s)}\iff f
  \not \in \Phi_0(a,\sigma)$.

\item $ab(f,a,\sigma) \in A_{(\Phi_0,\sigma,s)}$ if and only if there
  exists an object effect proposition of the form
  $$a\;\causes\;f\;\If\;p_1\ldots p_m$$ such that $p_1\ldots p_m$
  holds in $\sigma$ or a non deterministic effect proposition of the
  form $$a \may f \If p_1,\ldots,p_m$$in $D$ such that
  $p_1,\ldots,p_m$ holds in $\sigma$ and $f$ holds $\Phi_0(a,\sigma)$.

\item $ab(\bar f,a,\sigma) \in A_{(\Phi_0,\sigma,s)}$ if and only if
  there exists an object effect proposition of the form
  $$a\;\causes\;\neg f\;\If\;p_1\ldots p_m$$ such that $p_1\ldots
  p_m$ or a non deterministic effect proposition of the form $$a \may
  f \If p_1,\ldots,p_m$$ in $D$ such that $p_1,\ldots,p_m$ and $f$
  does not hold in $\Phi_0(a,\sigma)$.

\end{enumerate}
Nothing else belongs to $A_{(\Phi_0,\sigma,s)}$.
\end{definition}

\begin{definition}
Let $D$ be a domain description,
%without {\it knowledge laws} 
$\Phi_0$ a {\it 0-interpretation}, and $\sigma$ a state.  We will
say that the pair $(\Phi_0,\sigma)$ is a {\it 0-specific model} of
the domain description $D$ if $\sigma$ is an initial state of $D$
and  $\Phi_0$ one of its 0-models.
\end{definition}

In the next two theorems we will prove that the logic program models
correctly the execution of a single non-sensing action.

\begin{theorem}
Let $D$ be a consistent universal domain description with no knowledge 
laws, and $\sigma$ a state. If $(\Phi_0,\sigma)$ is a 0-specific model 
of $D_{\sigma}$, then $A_{(\Phi_0,\sigma,s) }$ satisfies every rule in
  $\Pi^1_{D_{(\sigma,s)}}$, for any situation constant $s$.
\end{theorem}

\subsection*{Proof}

 Assume that $(\Phi_0,\sigma)$ is a {\it 0-specific
    model} of $D_{\sigma}$.
% we will prove that
%  $A_{(\Phi_0,\sigma,s) }$ is a belief set of $\Pds{1}$.  
  Then any fact  
%\noindent
 $holds(f,s)$ or $holds(\o{f},s)$ in $\Pds{1}$ is such that  
either $holds(f,s)$ is in $\Aps$ or $holds(\o{f},s)$ is
  in $\Aps$, which is obvious.  Furthermore, for any literal of the form 
$holds(f,s)$
  (resp. $holds(\o{f},s)$) in $\Pds{1}$ obtained from the translation
  of a proposition of the form $\initially f$ (resp. $\initially \neg
  f$), we will have  by construction that $holds(f,s)\in\Aps$
  (resp. $holds(\o{f},s)\in\Aps$).  Now, let us take a pair of rules
  of the form

$~~~~holds(f,res(a,s)) \leftarrow holds(p_1,s)
\ldots,holds(p_m,s)$  

$~~~~ab(f,a,s)\leftarrow holds(p_1,s),\ldots,holds(p_m,s),
                         \Not holds(true,s)$

\noindent
obtained from the translation of a proposition of the form 
$$a \causes f \If p_1,\ldots, p_m$$, and assume that $holds(p_1,s),\ldots,holds(p_m,s)
\in A_{(\Phi_0,\sigma,s)}$. Then by construction, $p_1,\ldots,p_m$
holds in $\sigma$ and $holds(true,s)
 \not\in A_{(\Phi_0,\sigma,s)}$.
Therefore $ab(f,a,s) \in A_{(\Phi_0,\sigma,s)}$ and $f$ is in
$\Phi_0(a,\sigma)$.  Consequently, $holds(f,res(a,s))$ holds in
$A_{(\Phi_0,\sigma,s)}$.

The rules:

$~~~~holds(true,res(A,S)) \leftarrow holds(true,S)$

$~~~~holds(F,S) \leftarrow holds(true,S)$

\noindent
are trivially satisfied since there are no atoms of the form
$holds(true,s)$ in $A_{(\Phi_0,\sigma,s)}$.

Now we will make several considerations on $A_{(\Phi_0,\sigma,s)}$ to
evaluate the other rules (ground instances of the inertia rule and
rules obtained from the translation of non-deterministic effect
propositions):

\begin{enumerate}
\item $ab(f,a,s)$ holds in
 $A_{(\Phi_0,\sigma,s)}$. \\
Any rule of the form
$holds(\o{f},res(a,s)) \leftarrow holds(f,s),\Not ab(f,a,s)$
(instance of the inertia rule) is removed from $\Pds{1}$ to verify that
$A_{(\Phi_0,\sigma,s)}$ is a belief set of $\Pds{1}$.  Moreover, by
the definition of $A_{(\Phi_0,\sigma,s)}$, there must be an effect
proposition with one of the following forms 
$$a \causes  f \If p_1, \ldots, p_m $$  with $p_1,\ldots, p_m$
true in $\sigma$ or 
$$a\may f \If p_1,\ldots,p_m$$ in $D_{\sigma}$ with $p_1,\ldots, p_m$
true in $\sigma$ and $holds(f,res(a,s))$ member of
$A_{(\Phi_0,\sigma,s)}$ by case (3) above. So any pair of rules of the form

$$\begin{array}{ll}
ab(f,a,s) \leftarrow & \Not
 holds(\o{f},res(a,s)), 
\\ & holds(p_1,s)),\ldots holds(p_n,s)),
 \Not holds(true,s)\\
holds(f,res(a,s)) \leftarrow  & \Not
holds(\o{f},res(a,s)), \\ & holds(p_1,s),\ldots holds(p_n,s)
\end{array}$$
coming from the translation of a non-deterministic effect proposition
of the form 
$$a \may f \If p_1,\ldots,p_m$$
will be trivially satisfied in $A_{(\Phi_0,\sigma,s)}$.  The
other two rules obtained from the non-deterministic effect
propositions are of the form
$$\begin{array}{ll}
ab(\o{f},a,s) \leftarrow & \Not
holds(f,res(a,s)),
\\ & holds(p_1,s)),\ldots, holds(p_n,s),
\Not holds(true,s) \\
holds(\o{f},res(a,s)) \leftarrow & \Not
holds(f,res(a,s)), \\
& holds(p_1,s),\ldots holds(p_n,s)
\end{array}$$
and they will also be removed from $\Pds{1}$ to verify that
$A_{(\Phi_0,\sigma,s)}$ is a belief set of $\Pds{1}$, since 
$holds(\o{f},res(a,s)) \in \Aps$, and this concludes the proof for this case.

\item  $ab(\bar f,a,s)$ holds in $A_{(\Phi_0,\sigma,s)}$.

Similar to previous case.

\item Neither $ab(f,a,s)$ nor $ab(\bar f,a,s)$ are in
  $A_{(\Phi_0,\sigma,s)}$.\\
In this case we will have no effect propositions of the form:
\begin{itemize}
\item $a \causes f \If p_1,\ldots,p_n$
\item $a \causes \bar f \If p_1,\ldots,p_n$
\item $a \may f \If p_1,\ldots,p_n$
\end{itemize}
 in $D_{\sigma}$ with $p_1,\ldots,p_n$ true in $\sigma$. 
Therefore any rule $r$ in $\Pds{1}$
 with predicates involving $\Not$,
$a$ and $f$, will be such that the body of $r$ does not 
hold in $A_{(\Phi_0,\sigma,s)}$, unless, possibly  for those rules of
the form

$~~~~holds(f,res(a,s)) \leftarrow holds(f,s), \Not ab(\o{f},a,s)$

$~~~~holds(\o{f},res(a,s)) \leftarrow holds(\o{f},s),
 \Not ab( f,a,s)$

instances of the inertia rule.  So we will verify these by cases,
\begin{itemize}
\item  $holds(f,s)$ holds in $A_{(\Phi_0,\sigma,s)}$, then 
$holds(\o{f},s)$ does not belong to $\Aps$ and there is nothing to
verify for the second rule.  The first rule is transformed into

$~~~~holds(f,res(a,s)) \leftarrow
holds(f,s)$

and it is satisfied by $A_{(\Phi_0,\sigma,s)}$ because $f$
is in $\sigma$ and since there are no effect
propositions of the above types, and $f$ is true in
$\Phi_0(a,s)$, by definition of $A_{(\Phi_0,\sigma,s)}$, 
we have $holds(f,res(a,s)) \in \Aps$.
\item $holds(\o{f},s)$ holds in $A_{(\Phi_0,\sigma,s)}$

The proof is similar to the previous case.
  
\end{itemize}   
\end{enumerate}
\qed

\begin{theorem}
\label{A is a belief set}
Let $D$ be a consistent universal domain description with no knowledge 
laws, and $\sigma$ be a state. If $(\Phi_0,\sigma)$ is a 0-specific model 
of $D_{\sigma}$, then $A_{(\Phi_0,\sigma,s) }$ 
is a belief set of
 $\Pi^1_{D_{(\sigma,s)}}$, for any situation constant $s$.

\end{theorem}
By the above theorem 
we just need to prove
%what remain to prove is 
that $A_{(\Phi_0,\sigma,s)}$ is minimal in 
the family of models of $\Pi^1_{D_{(\sigma,s)}}$.  Let $B$
be a proper subset of $A_{(\Phi_0,\sigma,s)}$ and $Q$ some predicate
in $A_{(\Phi_0,\sigma,s)} \setminus B$.  Then $Q$ could be a literal of one of
the following five types:
\begin{itemize}
\item[i.-] $Q = holds(f,s)$, in this case there will be a fact in
$\Pds{1}$ not covered by $B$, so it would not be a belief set of $\Pds{1}$.

\item[ii.-] $Q = holds(f,res(a,s))$. $f \in \Phi_0(a,\sigma)$ since $Q$ is in
  $A_{(\Phi_0,\sigma,s)}$,%
\footnote{Note that by consistence of $D_{\sigma}$ (we are assuming
that $(\Phi_0,\sigma)$ is a 0-specific model) there is no rule of
the form ``$a
  \causes \neg f \If p_1,\ldots,p_m$'' in $D_{\sigma}$ with $p_1,\ldots
  p_m$ holding in $\sigma$.} therefore,

\begin{itemize}
\item If there is a rule ``$a \causes f \If p_1,\ldots,p_m$'' in
  $D_{\sigma}$ with $p_1,\ldots,p_m$ holding in $\sigma$, there is a
  rule $holds(f,res(a,s)) \leftarrow
  holds(p_1,s),\ldots,holds(p_m,s)$\\ in $\Pds{1}$ with
  $holds(p_1,s)\ldots,holds(p_m,s)$ members of $A_{(\Phi_0,\sigma,s)}$
  and by (i), \\$holds(p_1,s)\ldots,holds(p_m,s)$ hold in $B$,
  therefore this rule will not be satisfied in $B$.
\item If there is a rule $a \may f \If p_1,\ldots,p_m$ with
  $p_1,\ldots,p_m$ in $\sigma$, since\\ $holds(\o{f},res(a,s))$ can
 not be in $B$ (otherwise $\Aps$ would be inconsistent) and
 $holds(f,res(a,s))$ is not in $B$, we will have that $B$ does not
 satisfies the rule \\ $holds(f,res(a,s)) \leftarrow \Not 
 holds(\o{f},res(a,s)), holds(p_1,s),\ldots,holds(p_m,s)$.
\item If there are no effect propositions in $D_{\sigma}$ involving
  $a$ and $f$, then, we have that $f$ is in $\sigma$, because in this
  case $f \in \Phi_0(a,\sigma)$ if and only if $f \in \sigma$, and the
  rule that will not be satisfied by $B$ is the (ground instance of
  the) inertia rule $holds(f,res(a,s))\leftarrow holds(f,s), \Not
  ab(\bar f,a, s)$.
%  \\ because $ ab(\bar f,a,s)$ is not in
%  $A_{(\Phi_0,\sigma,s)}$, since there is not rule $a \causes f \If
%  p_1,\ldots,p_m$ or $a \may f \If p_1,\ldots,p_m$ in $D_{\sigma}$
%  such that $p_1,\ldots,p_m$ hold in $\sigma$.
\end{itemize}
\item[iii.-] $Q = holds(\o{f},res(a,s))$ 

The proof of this case is
similar to the previous case.

\item[iv.-] $Q = ab(f,a,s)$\\
In this case we have that there is either an effect proposition of the
form $$a \causes f \If p_1,\ldots,p_m$$ with $p_1,\ldots,p_m$ true in
$\sigma$ or $$a \may f \If p_1,\ldots,p_m$$ in $D_{\sigma}$ with
$p_1,\ldots,p_m$ true in $\sigma$, and $holds({f},res(a,s)) \in
A_{(\Phi_0,\sigma,s)}$.

%Note that since $ab(f,a,s)$ is in $A_{(\Phi_0,\sigma,s)}$, by consistence,
%  $ab(\bar f,a,s)$ can not be in $B$; otherwise both
%$holds(f,res(a,s))$ and $holds(\o{f},res(a,s))$ will be in $\Aps$.
Hence, one of the following two rules are not satisfied in $B$
$$\begin{array}{ll}
ab(f,a,s) \leftarrow & holds(p_1,s) \ldots holds(p_m,s),
                          \Not holds(true,s) \\
ab(f,a,s) \leftarrow & \Not holds(\o{f},res(a,s)),\\
& holds(p_1,s),
\ldots, holds(p_m,s), \Not holds(true,s).
\end{array}$$

\item[v.-] $Q= ab(\bar f,a,s)$.

Similar to previous case.
\qed
\end{itemize}
%\end{enumerate}
%
We prove completeness in two steps.  First, we show that if a belief
set of $\Pi^1_{(D_\sigma,s)}$ is defined as in Defintion~\ref{A1} then 
$(\Phi_0,\sigma)$ is a 0-specific model of $D_\sigma$.  Then we show
that every belief set of $\Pi^1_{(D_\sigma,s)}$ must be of this form.

\begin{theorem}
\label{if Phi0 is a 0,1 sp model}
  Let $D$ be a consistent universal domain description with no
  knowledge laws, $\Phi_0$ a 0-interpretation, and $\sigma$ a
  state. If $A_{(\Phi_0,\sigma,s)}$ is a {\it belief set} of
  $\Pi^1_{(D_{\sigma},s)}$, then $(\Phi_0,\sigma)$ is a {\it
  0-specific model} of $D_{\sigma}$.
% = D \cup \{\initially f: f\in  \sigma\} \cup \{\initially\neg f: f \not\in
%  \sigma\}$ 
%if and only if $A_{(\Phi_0,\sigma,s)}$ is a {\it belief set} of
%$\Pi^1_{(D_{\sigma},s)}$, for any situation constant $s$.
\end{theorem}

\subsection*{Proof}

%\begin{enumerate}
 Let $A_{(\Phi_0,\sigma,s)}$ be a belief set
of $\Pds{1}$. Clearly, by construction,  $\sigma$ is an initial state
of $D_\sigma$.  Now let $f$ be a fluent such that there is an effect
proposition of the form ``$a \causes f \If p_1,\ldots, p_m$'' in
$D_{\sigma}$, and assume that $p_1,\ldots,p_m$ hold in $\sigma$. Then,
by construction, there is a rule of the form

$~~~~holds(f,res(a,s)) \leftarrow holds(p_1,s),\ldots,holds(p_m,s)$

\noindent
in $\Pds{1}$ such that
$holds(p_1,s), \ldots, holds(p_m,s) \in A_{(\Phi_0,\sigma,s)}$.

Therefore $holds(f,res(a,s)) \in A_{(\Phi_0,\sigma,s)}$ 
and hence $f
\in \Phi_0(a,\sigma)$.  The proof is analogous for effect propositions of the
form $a \causes \neg f \If p_1,\ldots, p_m$.  If $f$ is a fluent such
that there are no effect propositions of the above two types we have
two possible situations. 

\medskip\noindent
{\bf (i)} If there are no non-deterministic
effect propositions of the form $$a \may f \If p_1,\ldots,p_m$$ with
$p_1,\ldots,p_m$ holding in $\sigma$, we will have that there are no
rules in $\Pds{1}$ such that $ab(f,a,s)$ appears in the head of the
rule and whose body holds in $A_{(\Phi_0,\sigma,s)}$. Therefore by the
rules

$~~~~holds(f,res(a,s))\leftarrow holds(f,s),\Not ab(f,a,s)$ and 

$~~~~
 holds(\o{f},res(a,s))\leftarrow holds(\o{f},s),\Not ab(f,a,s)$, 

\noindent
(ground instances of the inertia rule), we will have that either
$holds(f,res(a,s))$ is in $A_{(\Phi_0,\sigma,s)}$ or $
holds(\o{f},res(a,s))$ is in $A_{(\Phi_0,\sigma,s)}$, since either
$f\in\sigma$ or $f\not\in\sigma$, forcing either $holds(f,s)$ or $
holds(\o{f},s)$ to be in $\Aps$. Thus, $holds(f,res(a,s))$ (resp. 
$holds(\o{f},res(a,s))$) is in $\Aps$ if and only if $holds(f,s)$
(resp. $holds(\o{f},s)$) is in $A_{(\Phi_0,\sigma,s)}$.  Therefore
$f\in \Phi_0(a,\sigma)$ if and only if $f$ is in $\sigma$.

\medskip\noindent
{\bf (ii)} On the other hand, if there is a proposition of the form
$$a \may f \If p_1,\ldots,p_m$$ in $D_\sigma$ with $p_1,\ldots,p_m$
holding in $\sigma$, by construction, we will have in $\Pds{1}$ the
following rules $$
\begin{array}{lll}
holds(f,res(a,s)) &\leftarrow &
\Not holds(\o{f},res(a,s)),\\&&
holds(p_1,s),\ldots,
holds(p_m,s)  \\ 
holds(\o{f},res(a,s)) &\leftarrow &  \Not 
holds(f,res(a,s)),\\ &&
holds(p_1,s),\ldots,
holds(p_m,s)  \\
ab(f,res(a,s)) & \leftarrow & \Not 
holds(\o{f},res(a,s)) ,  \Not holds(true,s),\\ &&
holds(p_1,s),\ldots,
holds(p_m,s)  \\ ab(\bar
f,a_n,res(a,s)) & \leftarrow   & \Not 
holds(f,res(a,s)), \Not holds(true,s), \\ & &
holds(p_1,s),\ldots,
holds(p_m,s)  
\end{array}
$$

Thus, since $\Aps$ is a belief set of $\Pds{1}$ and the $holds(p_i,s)$ are
assumed to belong to $\Aps$ for every $i$, then either
$holds(f,res(a,s))$ or $holds(\o{f},res(a,s))$ must be in
$A_{(\Phi_0,\sigma,s)}$, but not both.  Therefore it does not matter
if $f$ is or is not part of $\Phi_0(a,\sigma)$. 
Hence $(\Phi_0,\sigma)$ is a  0-specific model of $D_{\sigma}$.
\qed

Observe that for any domain description $D$, any state $\sigma$ 
and any initial situation constant $s$, a 
 set of predicates $A$ will be a belief set of $\Pi^1_{(D,s)}$ 
if and only if $A$ is the union of belief sets of 
$\Pi_{(D,s,a)}$,  for each possible action $a$.
 $A = \bigcup\{A_a: a\;\; \mbox{is a possible action}\}$ 
 with each $A_a$ a belief set of $\Pi_{(D,s,a)}$. This is because if $a_1$ and 
$a_2$ are two different  actions then  
none of the predicates in rules in $\Pi_{(D,s,a_1)}$ appear in any 
predicate of any rule in $\Pi_{(D,s,a_2)}$, so the computation of the 
belief sets for one of the programs does not affect the computation
for the other one.

\begin{theorem}
%\label{A is a belief set}
\label{if A is a belief set}
Given a consistent domain description $D$, and a situation constant
$s$.  If $A$ is a belief set for $\Pi^1_{(D,s)}$ , then there exists a state
$\sigma$, and a 0-specific model $\Phi_0$ of $D_{\sigma}$ such that
$A = A_{(\Phi_0,\sigma,s)}$.
\end{theorem}

\subsection*{Proof}

By definition of $\Pi^1_{D,s}$, $A$ must be complete.
That is, for any fluent $f$ we have that either $holds(f,s)$ is in $A$
or $holds(\o{f},s)$ is in $A$.  Thus, if we let $\sigma = \{f:
holds(f,s)\in A\}$ and $\Phi_0$ be  such that for any possible action $a$,
$f \in
\Phi_0(a,\sigma)$ if and only if $holds(f,res(a,s)) \in A$, we will
have by completeness that $f \not \in \sigma$ if and only if
$holds(\o{f},s)$ is in $A$ and $f \not \in \Phi_0(a,s)$ if and only if
$holds(\o{f},res(a,s)) \in A$. Moreover, if some predicate $ab(f,a,s)$
is in $A$ then one of the following facts holds:
\begin{itemize}
\item There is a rule in $\Pi^1_{(D,s)}$ 
whose body is $$holds(p_1,s),\ldots,holds(p_m,s), \Not holds(true,s)$$
and whose head is $ab(f,a,s)$ such that $holds(p_i,s) \in A$ for any
$i=1,\dots,m$, and $holds(true,s)\not\in A$. Thus, there must be an
effect proposition of the form ``$a \causes f \If p_1,\dots,p_m$'' in $D$
with $p_1,\ldots,p_m$ true in $\sigma$.

\item There is a rule in $\Pi^1_{(D,s)}$ whose body is

$\Not
holds(\o{f},res(a,s))$, $holds(p_1,s),\dots,holds(p_m,s)$, $\Not
holds(true,s)$ 

 whose head is $ab(f,a,s)$ such that
$holds({f},res(a,s))$ and each $holds(p_i,s)$ are in $A$, for each
$i=1,\dots,m$, and $holds(true,s)$ is not in $A$.  Therefore, in this
case there exists a non-deterministic effect proposition of the form
``$a
\may f \If p_1,\ldots,p_m$'' in $D$ with $f,p_1,\ldots p_m$ true in
$\sigma$.

\end{itemize}

If for some fluent $f$, $ab(f,a,s)$ is in $A$, we will have by similar
reasons that there exists a proposition of the form 
$$a \causes \neg
 f \If p_1,\ldots,p_m \;\mbox{ or}\; a \may f \If
 p_1,\ldots,p_m$$ in $D$ such that $p_1,\ldots,p_m$ hold in $\sigma$
 and $f \not \in \Phi_0(a,\sigma)$.

So we have proved that $A = A_{(\Phi_0,\sigma,s)}$, and by
Theorem~\ref{if A is a belief set}, $(\Phi_0,\sigma)$ is a 0-specific
model.
\qed

\medskip

We now extend the proof to handle sensing actions.  Let
$Mod_0(D,\Sigma)$ denote the set $\{(\Phi_0,\sigma):\;\;
\mbox{0-specific model of}\;\; D,\;\; \sigma \in \Sigma \} $, where
$\Sigma$ is the set of initial states of $D$.  Note that the set can be
empty if there is no state in $\Sigma$ that is an initial state of
$D$.

\begin{definition}
Given a consistent situation $\Sigma$, and an interpretation $\Phi$.
  $(\Phi,\Sigma)$ will be a $1$-{\it specific model} of a consistent domain
  description $D$ if:

\begin{enumerate}
\item $\Sigma$ is an initial situation of  $D$.

\item
For any non-sensing action $a$,
$\Phi(a,\Sigma) = 
\bigcup_{\Phi_0 \in Mod_0(D,\Sigma)}\bigcup_{\sigma\in\Sigma}\{\Phi_0
(a,\sigma)\}$.

\item For each sensing action $a$, if

$$
\begin{array}{l} 
 a\; \toknow\;f_1\;\If\;\varphi_1\\
\vdots \\ 
a\; \toknow \; f_{s} \; \If \; \varphi_{s_a}
\end{array}
$$
are all the knowledge laws in $D$ where $a$ occurs. 
Then, $\Phi(a,\Sigma)$
 must be consistent and if $s_a=0$, $\Phi(a,\Sigma)= \Sigma$; 
otherwise $\Phi(a,\Sigma) = 
\bigcap_{l=1,\ldots,s}\Sigma_l$ such that each $\Sigma_l$ is 
a situation $(f_l,\varphi_l)$-compatible with $\Sigma$.  (Recall that
since $D$ is simple, all the $f_i$ are different).
 
\end{enumerate}

\end{definition}

\begin{definition}
\label{world view definition}

Let $(\Phi,\Sigma)$ be a 1-specific model of a domain description $D$. 
Denote by $\Asso{D}{\Phi}{\Sigma}$ the set of 0-specific models of $D$
such that for any non-sensing action $a$, $\Phi(a,\Sigma) =
\bigcup_{\sigma \in \Sigma}\{\Phi_0(a,\sigma)\}$.

\noindent
Let $a$ be a sensing action.  Define:

$A^a_{(\sigma,s)} = \{ holds(f,res(a,s)): \sigma\models f$ with $f$ a
fluent literal \} if $\sigma \in \Phi(a,\Sigma)$.  Otherwise,

$A^a_{(\sigma,s)} = \{ holds(true,res(a,s)) \} \cup \{
holds(f,res(a,s)):$ $f$ fluent literal \}.

Let $A'_{(\Phi_0,\sigma,s)} = A_{(\Phi_0,\sigma,s)} \cup\bigcup_{a\in
Sensing} A^a_{(\sigma,s)}$.

Let ${\bf A}_{(\Phi,\Sigma,s)} = \{ A'_{(\Phi_0,\sigma,s)}:
(\Phi_0,\sigma) \in \Asso{D}{\Phi}{\Sigma} \}$.

As a straightforward consequence of
 this definition we have that for any fluent $f$ 
and any action $a$, $\Phi(a,\Sigma)\models f$ iff
 $ {\bf A}_{(\Phi,\Sigma,s)}\models holds(f,res(a,s))$.

\end{definition}

For any set $A$ of literals we will denote by 
$\sigma_A$ the state $\sigma_A = \{ f : holds(f,s) \in
A\}$

We will denote by $D^{n.s}$ the set of value and effect propositions
in $D$, and by $D^{sen}$ the set of knowledge laws in $D$.

As a corollary of the theorems [\ref{A is a belief set}, \ref{if A is
a belief set}] we will have the soundness and completeness of the
logic program translation for the execution of a single action
(sensing or not).  The next corollary shows soundness and
Corollary~\ref{For 1-plans,2} shows completeness.

\begin{corollary}
\label{For 1-plans}
Let $D$ be a consistent  domain description. If 
$(\Phi,\Sigma)$ is a 1-specific model of $D$ then ${\bf
A}_{(\Phi,\Sigma,s)}$ is a world view of
$\Pi^1_{(D,s)}$.
\end{corollary}

\noindent
%$(\Leftarrow)$ Reciprocally
\subsection*{Proof}

Let us suppose that $(\Phi,\Sigma)$ is a 1-specific model of $D$.  We
will prove that $\bfAs$ is a world view of $\Pi^1_{(D,s)}$. In other
words, we will show that $\bfAs$ is the collection of belief sets of
$[\Pi^1_{(D,s)}]_{\bfAs}$ (see Section \ref{ELP} for the definition of
$[\Pi]_{\bf A}$ and $[\Pi]^A_{\bf A}$). Given an $A
\in \bfAs$, let $A =  A'_{(\Phi_0,\sigma,s)}$, and denote by $\Pi$
the program $[\Pi^1_{(D,s)}\setminus\Pi^1_{(D^{n.s},s)}]^A_{\bfAs}$
which is equal to
$[\Pi^1_{(D^{sen},s)}]^A_{\bfAs}$ union all the rules of the form
$holds(true,res(a,s))\leftarrow holds(true,s)$ and
$holds(f,res(a,s))\leftarrow holds(true,res(a,s))$ where $f$ is a
fluent literal and $a$ is a sensing action.

The set $U = lit(\Pi^1_{(D^{n.s},s)})$ split $[\Pi^1_{(D,s)}]_{\bfAs}$, 
and by theorem \ref{A is a belief set} $\Aps$ is a belief set 
of $\Pi^1_{(D^{n.s},s)}$, moreover $b_U(\Pi^1_{(D,s)}) = \Pi^1_{(D^{n.s},s)}$
 and any answer set of 
$e_U([\Pi^1_{(D,s)}]_{\Aps} \setminus \Pi^1_{(D^{n.s},s)},
 A_{\Phi_0,\sigma,s)})$ is a belief set of  $A_{\Phi_0,\sigma,s)})$. 
 
%By the splitting lemma and Theorems [\ref{A is a belief set},
%\ref{if A is a belief set})],
Hence by {\it Splitting Lemma} we only
need to prove that $A$ is a belief set of $A_{(\Phi_0,\sigma,s)}
\cup \Pi$.
We first prove that all the rules in the program hold in $A$ and then
we show that $A$ is minimal.

Obviously any fact in $A_{(\Phi_0,\sigma,s)} \cup \Pi$ is in $A$, thus
we will prove that any rule $R$ in $\Pi$ holds in $A$. Let, for a
given sensing action $a$,
 $$
\begin{array}{l}
a\;\toknow\;f_1\;\If\;p^1_1,\ldots,p^1_{n_1}\\ \vdots \\
a\;\toknow\;f_{s_a}\;\If\;p^{s_a}_1,\ldots,p^{s_a}_{n_{s_a}}
\end{array}
$$
be all the knowledge laws in $D$, involving $a$, and 
$\Phi(a,\Sigma) = \bigcap_{l=1}^{s_a}\Sigma_l$ where each $\Sigma_l$ is 
$(f_l,p^l_1,\ldots,p^l_{n_l})$-compatible with $\Sigma$. The rules $R$ in
$\Pi$  that
mention $a$ either in its body or in its head will be evaluated as follows:

\begin{enumerate}
\item If $\Sigma\models f_l$ or $\Sigma\models \neg f_l$, there are no
   rules in the program with the predicate $holds(true,res(a,s))$ in
   the head that are not ground instances of domain independent rules
   ( because any rule in $\Pi^1_{(D^{sen},s)}$ will be removed, to get
   $[\Pi^1_{(D^{sen},s)}]^A_{\bfAs}$ after checking

 $\neg
Kholds(f_1,res(a,s)), \neg Kholds(\o{f_1},res(a,s))$). 

\item If $\Sigma\not\models f_l$ and $\Sigma\not\models \neg f_l$, then
  either 

  (a) $\Phi(a,\Sigma) \models p^l_1,\ldots,p^l_{n_l}$, and $\Phi(a,\Sigma)
  \models f_l$.  In this case,  $R$ must be one of the following:

$holds(true,res(a,s)) \leftarrow holds(\o{p}^l_1,s)$\\
$\vdots$

$holds(true,res(a,s)) \leftarrow holds(\o{p}^l_{n_l},s)$

$holds(true,res(a,s)) \leftarrow holds(\o{f}_l,s)$

\noindent and each of these rules are verified in $A$, 
 because; if $ holds(\o{f}_l,s)$ is in $A$  or 
for some $i$  $holds(\o{p}^l_i,s) \in A$  then $\sigma_A \models \o{p}^l_i$ 
or  $\sigma_A \models \o{f_l}$ 
and in both cases $\sigma_A \not\in \Phi(a,\Sigma)$, 
and hence, $holds(true,res(a,s))\in A$,by definition of $A^a_{(\sigma,s)}$.

\noindent
(b) $\Phi(a,\Sigma) \models p^l_1,\ldots,p^l_{n_l}$, and $\Phi(a,\Sigma)
  \models \o{f_l}$
 This case is similar to (a) changing the last rule for

$holds(true,res(a,s)) \leftarrow holds(f_l,s)$.

\item $\Sigma\not\models f_l$ and $\Sigma\not\models \neg f_l$, and there
  exists $i=1,\ldots,n_l$, such that $\Phi(a,\Sigma) \not\models
  p^l_i$. In this case, there is only one rule that remains in the program;

$holds(true,res(a,s)) \leftarrow holds(p^l_1,s), \ldots,
holds(p^l_{n_l},s)$

\noindent
Now, if for any $i=1,\ldots,n_l$, $holds(p^l_i,s) \in A$, then
$\sigma_A \models p^l_i$, for every $i=1,\ldots,n_l$, and
$\sigma_A \not\in \Phi(a,\Sigma)$. Thus, $holds(true,res(a,s)) \in A$.

\item From the domain independent rules, $R$ could also be of the form
$$holds(true,res(a,s)) \leftarrow holds(true,s)$$ But, by construction
of $A$, $holds(true,s) \not\in A$, and thus, $R$ is satisfied by $A$.

\item The last rules to consider, also coming from the domain
  independent rules, are all the rules of the form $holds(f,res(a,s))
\leftarrow holds(true,res(a,s))$, where $f$ is a fluent literal.  If
  $holds(true,res(a,s))$ belongs to $A$ then, by construction, $\sigma_A
  \not\in \Phi(a,\Sigma)$.  Therefore, by construction too,
  $holds(f,res(a,s))$ also belongs to $A$, for every $f$, fluent literal.

\end{enumerate}

%%%%%%%%%%%%%%%%%%%%%%%%%%%%%%%%%%%%%%%%%%%%%%%%%%%%%%%%%%%%%
%%%%%%%
%%%%%%%%%%%%%%%%%%%%%%%%%%%%%%%%%%%%%%%%%%%%%%%%%%%%%%%%
\noindent
To prove the minimality of $A$, let $C$ be a proper subset of $A$ and
$h$ a predicate in $A \setminus C$.\footnote{Recall that $A =
A'_{(\Phi_0,\Sigma,s)}$.}  We will find a rule $R$ in
$[\Pi^1_{(D,s)}]^A_{\bfAs}$ such that $R$ does not hold in $C$.

If $h$ is of the form $ab(f,a,s)$, $holds(f,s)$ or
$holds(f,res(a,s))$, with $a$ a non-sensing action, then $R$ can be
found in $[\Pi^1_{(D^{n.s},s)}]^A_{\bfAs}$ by Theorem~[\ref{A is a
belief set}].  Thus, all these $h$ must be in $C$.  Therefore, it
suffices to consider the case when $h = holds(f,res(a,s))$, with $a$ a
sensing action. If $h=holds(f,res(a,s))$, with $f$ a fluent literal,
and the rule $holds(f,res(a,s)) \leftarrow holds(true,res(a,s))$ is
satisfied in $C$, then $holds(true,res(a,s)) \not\in C$, in which
case, if $holds(true,res(a,s))$ is in $A'_{(\Phi_0,\sigma,s)}= A$ then
$\sigma
\not \in \Phi(a,\Sigma)$, hence  we will have that
there exists an $l=1,\ldots,s_a$, such that
% $\sigma$ is not $f_l,p^l_1,\ldots,p_{n_l}$- compatible with it self,
 %therefore there exist $\sigma_1,\sigma_2,\sigma^1,\ldots,\sigma^{n_l}$ 
%in $\Sigma$ such that $f \not \in \sigma_1$ $f \in \sigma_2$ and $p_i^l 
%\in \sigma^i$ for $i = 1,\ldots,n_l$.
 $\sigma \not \in \Sigma_l$,then since  
 $\Phi(a,\Sigma)$ is
 $f_l,p^l_1,\ldots,p_{n_l}$-compatible with $\Sigma$ and 
the remark at the end of [\ref{world view definition}] we will have
 that one rule $R$ of the following:
%Hence 
% if there is a $\sigma \in \Sigma$ which is not in $\Sigma_l$ it because
% therefore $\Sigma$ is not model of some ${p^l_i}$
% and it is not a model of $f_l$ or $\o{f_l}$, so we have that 
%in $A^a_{(\sigma_A,s)}$ is any predicate
%of the form  $holds(f,res(a,s))$ with a $f$ a fluent literal or $true$
% and there are some $...$ 
 
%or a 
%since $\Sigma_l$ is  then neither $\forall i=1,
%\ldots,n_l p^l_i\in \sigma_A \vee 
%f_l \in A$ or  $\forall i=1,\ldots,n_l p^l_i\in \sigma_A \vee 
%\o{f_l} \in A$ or  one of the following 
$$
\begin{array}{ll}

holds(true,res(a,s)) & \leftarrow holds(\o{f}_l,s) \\
holds(true,res(a,s)) & \leftarrow holds(f_l,s) \\
holds(true,res(a,s)) & \leftarrow holds(\o{p}^l_1,s) \\
& \vdots \\
holds(true,res(a,s)) & \leftarrow holds(\o{p}^l_{n_l},s) \\
holds(true,res(a,s)) & \leftarrow holds(p^l_1,s), \ldots,
                                  holds(p^l_{n_l},s)
\end{array}
$$ 
has to be such that both i) $R \in [\Pi^1_{(D^{sens},s)}]^A_{\bfAs}$ and
ii) the fluent literals appearing on the body of $R$ will be in
$\sigma_A$.  Hence the body of $R$ will be true in $A$ and therefore in
$C$, thus we can conclude that $R$ is not satisfied by $C$.
%@@@@@@@@@@@@@@@@@@@@@@@@@@@@@@@@@@@@@@@@@@@@@@@@@@@@@@@@@@@@@@ 
If $holds(true,res(a,s))$ is not in $A$, then $\sigma \in \Phi(a,\Sigma)$
 and $holds(f,res(a,s))$ is in $A$, but this happens
  if and only if $f \in \sigma$, which is true iff $holds(f,s) \in A$, and the 
  inertia rule will not be true in $C$.
%@@@@@@@@@@@@@@@@@@@@@@@@@@@@@@@@@@@@@@@@@@@@@@@@@@@@@@@@@@@@@@@@@@@
To complete the proof we need to show that any belief set of
$[\Pi^1_{(D,s)}]_{\bfAs}$ is of the form $A'_{(\Phi_0,\sigma,s)}$, for
some $(\Phi_0,\sigma) \in \Asso{D}{\Phi}{\Sigma}$.  Take now $A$, a belief
set of $[\Pi^1_{(D,s)}]_{\bfAs}$. By the splitting lemma, the set $A_0 = A
\setminus \{holds(f,res(a,s)): a$ sensing action \}, is a belief set of
$\Pi^1_{(D^{n.s},s)}$.  Then by (\ref{if A is a belief set}) there exists
$(\Phi_0,\sigma)$ in $\Asso{D}{\Phi}{\Sigma}$ such that $A_0 =
A_{(\Phi_0,\sigma,s)} = A_{(\Phi_0,\sigma_A,s)}$ taking  $\Phi(a,\Sigma) =
\bigcup_{\sigma \in \Sigma}\{\Phi_0(a,\sigma)\}$, it only remains
to be shown that for any sensing action $a$, both of the following are
satisfied: (i) If $\sigma_A \in \Phi(a,\Sigma)$ then $f \in \sigma_A
\Leftrightarrow holds(f,res(a,s)) \in A$, and (ii)  $\sigma_A \not\in
\Phi(a,\Sigma) \Leftrightarrow holds(true,res(a,s)) \in A$.

\noindent
For case (i), let $\sigma_A \in \Phi(a,\Sigma)$. Then, for any
$l=1,\ldots, s_a$, $\sigma_A \in \Sigma_l$.  The rules with heads of the
form $holds(f,res(a,s))$ and $f$ a fluent literal are:
$holds(f,res(a,s)) \leftarrow holds(true,res(a,s))$ and
the one of the ground instances of the inertia rule.  The body of the
first rule is false in $A$ because any rule with $holds(true,res(a,s))$ in
its head must have its body false in $A$. Then, $holds(f,res(a,s)) \in A
\Leftrightarrow holds(f,s) \in A \Leftrightarrow f \in \sigma_A$.

\noindent
For (ii),  $holds(true,res(a,s)) \in A$ if and only if there exists a rule
$R$ which body is true in $A$ and its head $holds(true,res(a,s))$.  Hence,
$R$ must be one of the following rules:

$$
\begin{array}{ll}
holds(true,res(a,s)) & \leftarrow holds(\o{f}_l,s) \\
holds(true,res(a,s)) & \leftarrow holds(f_l,s) \\
holds(true,res(a,s)) & \leftarrow holds(\o{p}^l_1,s) \\
& \vdots \\
holds(true,res(a,s)) & \leftarrow holds(\o{p}^l_{n_l},s) \\
holds(true,res(a,s)) & \leftarrow holds(p^l_1,s), \ldots,
                                  holds(p^l_{n_l},s)
\end{array}
$$
for some $l=1,\ldots,s_a$, and in any case, the rule $R$ belongs to
$[\Pi^1_{(D,s)}]_{\bfAs}$ and its body is true in $A$, if and only if
$\sigma_A \not\in \Sigma_l$. Hence, $\sigma_A \not\in \Phi(a,\Sigma)$.
\qed

\begin{corollary} (Completeness)
\label{For 1-plans,2}
Let $D$ be a consistent  domain description. If  $\bf A$ is a world view of
$\Pi^1_{(D,s)}$ then there exists a 1-specific model
$(\Phi,\Sigma)$ of $D$ such that ${\bf A} = {\bf
A}_{(\Phi,\Sigma,s)}$.
\end{corollary}

\subsection*{Proof}
%$(\Rightarrow)$
 Let $\bf A$ be a world view of $\Pi^1_{(D,s)}$.  Let $\Sigma = \{
 \sigma_A : A\in {\bf A}\}$.  If $A \in \bf A$, $\Phi^A_0$ will be a
 0-interpretation such that
$ A_{(\Phi^A_0,\sigma_A,s)} = A \setminus \{ holds(f,res(a,s)): a$ is
a sensing action and $f$ is $true$ or a fluent literal\}, which
can be found making use of Theorem~[\ref{if A is a belief set}].  We define
an interpretation $\Phi$ such that $\Asso{D}{\Phi}{\Sigma} = \{
(\Phi^A_0,\sigma_A) : A \in {\bf A}\}$, and $\Phi(a,\Sigma) =
\{\sigma_A: holds(true,res(a,s)) \not\in A\}$ for any sensing action
$a$.  Note that if $(\Phi,\Sigma)$ is a 1-specific model of $D$
 then $A = A'_{(\Phi_0^A,\sigma_A,s)}$ for any $A \in \bf A$.
  Thus, we will show that $(\Phi,\Sigma)$ is a
1-specific model of $D$ and we will have that
 ${\bf A }= {\bf A}_{(\Phi,\Sigma,s)}$.
%%%%%%%%%%%%%%%%%%%%%%%%%%%%%%%%%%%%%%%%%%%%%%%%%%%%%%%%%%%%%%%%%%%%%%%%%%%%%%%
It is clear that $\Sigma$ is the initial situation of $D$.  Then, if
$a$ is a non-sensing action, by definition, $\Phi(a,\Sigma) =
\bigcup\{\{\Phi^A_0(a,\sigma_A)\}: A \in {\bf A}\} = 
\bigcup\{\{\Phi_0(a,\sigma)\}: (\Phi_0,\sigma) \in
\Asso{D}{\Phi}{\Sigma}\}.$

If $a$ is a sensing action and,  
%
%\begin{array}{ll}
$a\;\toknow\; f_l\;\If\;p^l_1,\ldots,p^l_{n_l}, l = 1,\ldots,s_a$
%\\ \vdots \\
%a\;\toknow\; f_{s_a}\;\If\;p^{s_a}_1,\ldots,p^{s_a}_{n_{s_a}}
%\end{array}
%
are exactly the knowledge laws where $a$ appears, we need to show that for
each $l=1,\dots,s_a$, there exists a $\Sigma_l$,
$(f_l,p^l_1,\dots,p^l_{n_l})$-compatible with $\Sigma$ such that
$\Phi(a,\Sigma) = \bigcap_{l=1,\dots,s_a} \Sigma_l$.

\begin{enumerate}

\item
If ${\bf A}\models holds(f_l,s)$ or ${\bf A}\models holds(\o{f}_l,s)$ then
let $\Sigma_l = \Sigma$.

\item  
If ${\bf A}\not\models holds(f_l,s)$ and ${\bf A}\not\models
holds(\o{f}_l,s)$ and ${\bf A}\models holds(p^l_i,res(a,s))$, for
$i=1,\dots,n_l$, then:

\begin{enumerate}

\item if ${\bf A} \models holds(f_l,res(a,s))$, then let
$\Sigma_l =  \{ \sigma\in\Sigma: \sigma \models p^l_1,\dots,p^l_{n_l},f_l\}$.

\item
if ${\bf A} \models holds(\o{f}_l,res(a,s))$, then let
$\Sigma_l =  \{ \sigma\in\Sigma: \sigma \models p^l_1,\dots,p^l_{n_l},
\o{f}_l\}$.

\end{enumerate}

\item
  
If ${\bf A}\not\models holds(f_l,s)$ and ${\bf A}\not\models
holds(\o{f}_l,s)$ and ${\bf A}\not\models holds(p^l_i,res(a,s))$, for
some $i=1,\dots,n_l$, we let $\Sigma_l = \{ \sigma\in\Sigma :
\exists k=1,\dots,n_l, \sigma \not\models p^l_k \}$.

\end{enumerate}

We need to show next that $\Phi(a,\Sigma) = \bigcap_{l=1}^{s_a} \Sigma_l$.
For that, we will prove that for any $\sigma_A\in\Sigma$,
$\sigma_A\not\in \bigcap_{l=1,\dots,s_a} \Sigma_l$ if and only if
$holds(true,res(a,s)) \in A$.

First we will show that if $\sigma_A\not\in\bigcap_{l=1,\dots,s_l}
\Sigma_l$, then $holds(true,res(a,s))\in A$.  Let $k$ be an $l$ such
that $\sigma_A\not\in\Sigma_l$.  Thus, since $\sigma_A\in \Sigma$ then
$\Sigma\not\models f_k$ and $\Sigma\not\models \neg f_k$ and case
(1) above does not occur, and one of the following cases must hold:

\begin{enumerate}
\item[2.a.]
${\bf A}\models holds(p^k_{i},res(a,s))$, for every $i=1,\dots,n_k$, and
${\bf A}\models holds(f_k,res(a,s))$.  Therefore, either
$\sigma_A\not\models p^k_j$ for some $j=1,\dots,n_k$ or
$\sigma_A\not\models f_k$.  Hence, $holds(\o{p}^k_j,s) \in A$ or
$holds(\o{f}_k,s) \in A$ and the following rules will be part
of $[\Pi^1_{(D,s)}]^A_{\bfAs}$:

$holds(true,res(a,s)) \leftarrow holds(\o{p}^k_j,s)$

$holds(true,res(a,s)) \leftarrow holds(\o{f}_k,s)$

Thus, $holds(true,res(a,s)) \in A$.

\item[2.b.]
${\bf A}\models holds(p^k_{i},res(a,s))$, for $i=1,\dots,n_k$, and
${\bf A}\models holds(\o{f}_k,res(a,s))$, is similar to (2.a).
%
%Therefore, either
%$\sigma_A\not\models p^k_j$ for some $j=1,\dots,n_k$ or
%$\sigma_A\not\models \neg f_k$.  Hence, $holds(\o{p}^k_j,s) \in A$ or
%$holds(f_k,s) \in A$ and the the following rules will be part
%of $[\Pi^1_{(D,s)}]^A_{\bfAs}$:
%
%$holds(true,res(a,s)) \leftarrow holds(\o{p}^k_j,s)$
%
%$holds(true,res(a,s)) \leftarrow holds(f_k,s)$
%
%Thus, $holds(true,res(a,s)) \in A$.

\item[3.]  ${\bf A}\not\models holds(p^k_{i},res(a,s))$, for some
  $i=1,\dots,n_k$.  Therefore, for every $j=1,\ldots,n_k$,
  $\sigma_A \models p^k_j$.  Hence $A \models holds(p^k_j,s)$ and the
  following rule will be part of $[\Pi^1_{(D,s)}]^A_{\bfAs}$:

$holds(true,res(a,s)) \leftarrow holds(p^k_1,s),
\ldots,holds(p^k_{n_k},s)$

Thus, $holds(true,res(a,s)) \in A$.
\end{enumerate}

\noindent
For the other direction, assume $holds(true,res(a,s)) \in A$.  Then,
it must be the case that there exists a rule in
$[\Pi^1_{(D,s)}]^A_{\bfAs}$ with $holds(true,res(a,s))$ in the head
and its body true in $A$.  Note that this rule cannot be
$holds(true,res(a,s)) \leftarrow holds(true,s)$ by the construction of
$A$.  Thus, ${\bf A} \not\models holds(f,s)$ and ${\bf A} \not\models
holds(\o f,s)$; otherwise there will be no rule in
$[\Pi^1_{(D,s)}]^A_{\bfAs}$ with $holds(true,res(a,S))$
in its head (these are ground instances of rules derived from
knowledge laws).  We will inspect the remaining rules with
$holds(true,res(a,s))$ in the head and we will show that there exists
$\Sigma_k$ such that $\sigma_A
\not\in \Sigma_k$.

\begin{enumerate}

\item 
If the rules are of the form:

$holds(true,res(a,s)) \leftarrow holds(\o{p}^k_j,s)$

$holds(true,res(a,s)) \leftarrow holds(\o{f}_k,s)$

then ${\bf A}\models holds(p^k_{i},res(a,s))$, for $i=1,\dots,n_k$, and
${\bf A}\models holds(f_k,res(a,s))$.  Therefore, since
$holds(\o{p}^k_j,s)$ or $holds(\o{f}_k,s)$ has to belong to $A$, $\sigma_A
\not\in \Sigma_k$.

\item
If the rules are of the form:

$holds(true,res(a,s)) \leftarrow holds(\o{p}^k_j,s)$

$holds(true,res(a,s)) \leftarrow holds(f_k,s)$

then, similar to 1, ${\bf A}\models holds(p^k_{i},res(a,s))$, for
$i=1,\dots,n_k$, and ${\bf A}\models holds(\o{f}_k,res(a,s))$.
Therefore, since $holds(\o{p}^k_j,s)$ or $holds(f_k,s)$ has to belong
to $A$, $\sigma_A \not\in \Sigma_k$.

\item

If the rule is of the form:

$holds(true,res(a,s)) \leftarrow holds(p^k_1,s),
\ldots,holds(p^k_{n_k},s)$

then ${\bf A}\not\models holds(p^k_{i},res(a,s))$, for some
  $i=1,\dots,n_k$.  Therefore, since  for every $j=1,\ldots,n_k$ 
   $ holds(p^k_j,s)$ has to be in $A$, $\sigma_A \not \in \Sigma_k$.
\qed
\end{enumerate}

The next step is to show soundness and completeness for sequences of
actions.  Sequences of actions are the most simple plans.  We then
extend the proof to plans of all classes.  The general proof will be
by induction on the complexity of the plans.  Thus, we start by
formally defining complexity and other definitions required for the
inductions.
%%%%%%%%%%%%%%%%%%%%%%%%%%%%%%%%%%%%%%%%%%%%%%%%%%%%%%%%%%%%%%%%%%
\begin{definition}
We will define the {\em complexity\/} of a plan  $\beta$ ($comp(\beta)$) by: 
if the empty plan is $[]$, $comp([]) = 0$. 
For an action $a$ $comp(a)=1$. For complex plans, 
$comp(\If\;\varphi \; \Then \;\alpha)$ and 
$comp(\While\;\varphi\;\Do\;\alpha)$ is $comp(\alpha)+1$ and 
$comp(\If\;\varphi\;\Then\;\alpha_1\;\Else\;\alpha_2)$ and 
$comp([\alpha_1,\alpha_2]$ is  
$comp(\alpha_1) + comp(\alpha_2)$ 

We will say that a plan $\alpha$ is an $n$-{\em plan\/} if it has
complexity $n$, it will be an $\leq n$-{\em plan\/} if it has complexity
less or equal than $n$. $P_n$ will denote the set of $n$-plans, and
$P_{\leq n}$ the set of $\leq n$-plans.

We define the complexity of a situation constant $s$ inductively as 0
if $s= s_0$; or 1 plus the complexity of $s'$ if $s = res(a,s')$, for
any action $a$.  A situation $s$ will be called an $n$-{\em situation\/}
if its complexity is $n$. The complexity of a predicate of the form
$holds(f,s)$ with $f$ a fluent literal, or $holds(true,s)$, will be
the complexity of $s$, the complexity of predicates of the form
$ab(f,a,s)$, with $f$ a fluent literal and $a$ an action will be equal
to the complexity of $s$ plus one, the complexity of predicates of the
form $\Find{\beta}{s_1}{s}$ will be the complexity of $\beta$ plus the
complexity of $s_1$, and the complexity of a predicate of the form
$\Holds{F}{\beta}$ will be the complexity of the plan $\beta$.  We will
say that a predicate $h$ is an $\leq n$-{\em predicate\/}, if $h$ has
complexity $m$ and $m \leq n$. Given a plan $\alpha$, $[\alpha^1]$
will denote the plan $\alpha$ and $[\alpha^{n+1}]$ will denote the
plan $[\alpha|[\alpha^n]]$.  Denote by $\Pi^n_D$ the subprogram of
$\Pi_D$ restricted to those rules in $\Pi_D$ with $\leq n$-predicates.
Note that in any k-predicate in $\Pi^n_D$, the constant situation is 
a sequence of k actions.
\end{definition}

Given a domain description $D$ denote by $D^r$ the sub-domain 
of $D$ obtained when we remove from $D$ any value proposition.

\begin{definition}
  For any $n>0$, we will say that a pair $(\Phi,\Sigma)$ where $\Phi$
  is an interpretation and $\Sigma$ a situation, is an $n+1$-{\it
  specific model} of $D$ if and only if it is an $n$-specific model of
  $D$ and for any sequence of actions $seq_n = a_1,\ldots,a_n$,
  $(\Phi,\Gamma_\Phi([seq_n],\Sigma))$ is a 1-specific model of $D^r$
  (i.e. $D$ minus the value propositions). $(\Phi,\Sigma)$ will be a
  specific model of $D$ if it is an $n$-specific model of $D$ for any
  $n \geq 1$.

\end{definition}

\begin{definition}

  Given a sequence of actions $seq = a_1,\ldots,a_n$ and a situation
  constant $s$, $res((seq),s)$ denotes the situation constant
  $res(a_n,\ldots,res(a_1,s))$, $seq_{\emptyset}$ denotes the empty
  sequence and $res((seq_{\emptyset}),s)$ will be equal to $s$.  Let
  $Act_n$ be the set of all the sequences of $n$ actions. For any set
  of literals $A$ we take $\sigma_{(A,(seq))}$ as the state such that
  $f\in \sigma_{(A,(seq))} \Leftrightarrow holds(f,res((seq),s_0))\in
  A$.
 
\end{definition}

\begin{definition}
\label{bfA}

Given a pair $(\Phi,\Sigma)$ of interpretation and situation, $n \geq
0$ and the situation constant $s_0$, we will denote by ${\bf
A}^{n+1}_{(\Phi,\Sigma,s_0)}$ the following family of sets:

\begin{itemize}
\item  if $n=1$,
${\bf A}^1_{(\Phi,\Sigma,s_0)}
= {\bf A}_{(\Phi,\Sigma,s_0)}$ 

%\{A \cup \{\Find{a}{s_0}{res(a,s_0)}:a\in Act_1\} \cup
%\{\Holds{f}{a}: holds(f,res(a,s_0)) \in A\}: A \in 

\item 
If $n \geq 1$, let for any set $A$ of $(\leq n+1)$-predicates, $A_n$
and $A_1$ denote the sets of $(\leq n)$-predicates in $A$ and
$(n+1)$-predicates in $A$ (resp.).  Then, ${\bf
A}^{n+1}_{(\Phi,\Sigma,s_0)}$ is defined as a family of sets $A$ of
$(\leq n+1)$-predicates, such that the following is satisfied:

\begin{enumerate}
\item $A_n \in \bfA{n}$ 

%\item $A = A_n \cup \bigcup_{seq_n \in Act_n} \bigcup_{(\Phi_0,\sigma) \in 
%Asso_D(\Phi,\Gamma_\Phi([seq_n],\Sigma))}A,_{(\Phi_0,\sigma,res((seq_n),s_0))}$
%\footnote{Aqui cambie muchisimo la definicion pero esto es lo 
%que necesitamos y con esto la prueba es inmediata}.
%\COMMENT

\item  
If $holds(true,(seq_n,s_0))$ is in $A_n$ then $A_1 = \bigcup_{seq_n
\in Act_n} A_{(seq_n)} $ with $A_{(seq_n)}$ the set of all the
predicates of the form $holds(true,res((seq_n,a),s_0))$ or $
holds(f,res((seq_n,a),s_0))$

\item
If $holds(true,(seq_n,s_0))$ is not in $A_n$ then $A_1  = A_n
\cup \bigcup_{seq_n \in Act_n} A_{(seq_n)} $ where $A_{(seq_n)}$ is
$A'_{(\Phi_0,\sigma_{(A_n,seq_n)},res((seq_n),s_0))}$ for some
$(\Phi_0,\sigma_{(A_n,(seq_n))})$ in \\
$Asso_{D^r}(\Phi,\Gamma_{\Phi}((seq_n),\Sigma)$.

\end{enumerate}
\end{itemize} 
\end{definition}

\begin{lemma}
\label{f in Phi iff holds(f) in A}

Given $n > 0$, a consistent domain description $D$, a pair
$(\Phi,\Sigma)$, $n$-specific model of $D$, the initial situation
constant $s_0$.  We will have that for any fluent literal $f$ and any
sequence of actions $seq = a_1,\ldots,a_n$, $holds(f,res((seq),s_0)$
holds in ${\bf A}^n_{(\Phi,\Sigma,s_0)}$
% and $S$ is
%an $m$-situation such that $holds(true,S) \not \in A$ then there
%exists a state $\sigma_S$ such that for any $(n-m)$-plan $\alpha$,
%both $\sigma_S \in 
if and only if $f \in \Gamma_{\Phi}([seq],\Sigma)$. 
%\Gamma_\Phi(\alpha_S,\Sigma))$ 
% and for any fluent literal $f$ 
% $holds(f,res(\alpha,S)) \in A \Leftrightarrow f \in 
%\sigma^{\alpha}_{A}$. Therefore we have that for any $n$-plan $\beta$,   
%$\Gamma_{\Phi}(\beta,\Sigma) \models f$  
%is an $n$-specific model of $D$
% if and only if $ {\bf A}^n_{(\Phi,\Sigma,s_0)}
% \models holds(f,res(\beta,s_0))$.

\end{lemma}

\subsection*{Proof}
The proof is by induction on $n$
 and it is straightforward from the definition 
of  $ {\bf A}^{n+1}_{(\Phi,\Sigma,S)}$ 
%\begin{itemize}
%For $n=1$ the results is direct from (\ref{For 1-plans}).  For $n>1$
%the proof is straightforward by definition of .
\qed

%\end{itemize}

The next corollary proves by induction on the length of the sequence 
of actions that the logic program translation is sound and complete 
for the execution of a sequence of actions.

\begin{corollary}
\label{bf A is a world view}
Given a consistent domain description $D$ and $n >0$.  ${\bf A}$ is a
 world view of $\Pi^n_D$ if and only if there exists a pair
 $(\Phi,\Sigma)$, $n$-specific model of $D$ such that
$$
\begin{array}{l}
{\bf A} = {\bf A}^n_{(\Phi,\Sigma,s_0)}
%\bigcup_{a\in Act_1}

% {\bf A}^{(n,a)}_{(\Phi,\Sigma,s)}
%    \{A_{(\Phi_0,\sigma,s)}: 
%(\Phi_0,\sigma)\in \Asst{n}{D}{\Phi}{\Sigma}\}.
\end{array}
$$
\end{corollary}

\subsection*{Proof}

\noindent
$\bullet$ The base case ($n=0$) follows from \ref{For 1-plans} and
\ref{For 1-plans,2}.

\noindent
$\bullet$ Suppose the result is valid for any $m \leq n$.

\noindent

$\bullet$  $(\Rightarrow)$ Let {\bf A} be a world view of
$\Pi^{n+1}_D$. As in Definition \ref{bfA},
  define for any 
%$m\leq n$ and for any
 $A \in {\bf A}$, $A_n$ to be
  the subset of $A$ restricted to those predicates in $A$ involving
  just $\leq n$-predicates, so ${\bf A}_n = \{A_n: A \in {\bf A}\}$ will be
  a world view of $\Pi^{n}_D$, and by inductive hypothesis
  there is a pair $(\Phi_1,\Sigma)$, $n$-specific model of $D$ such
  that ${\bf A}_n = {\bf A}^n_{(\Phi_1,\Sigma,s_0)}$.

We first define an interpretation $\Phi$ such that
$(\Phi,\Sigma)$ is $n+1$-specific model of $D$.
%\msg{Gisela, $Act_n$ son las non-sensing actions? No deberia ser todas
%las acciones?}
Given an action $a$, let $A''$ be the set 
%$$
%\begin{array}{l}
$\{holds(f,res((seq_n),s_0)) \in A_{n}:   holds(true,res((seq_n),s_0)) 
\not \in  A_{n} \wedge seq_n \in  Act_n\}$
%\end{array}
%$$ 
 and $A_1$ be $ A''$ union
$$
\begin{array}{l}  
[\{holds(f,res((seq_n,a),s_0)): seq_n 
   \mbox{ is any sequence of } n \mbox{ actions}
 \}\\ \cup %\\  
 \{ab(f,a,res((seq_n),s_0)): seq_n
  \mbox{ is any sequence of } n \mbox{ actions} \} \\ \cup %\\
  \{holds(true,res((seq_n,a),s_0)): seq_n 
   \mbox{ is any sequence of } n \mbox{ actions} \}] \cap A
\end{array}
$$
%%%%%%%%%%%%%%%%%%%%%%%%%%%%%%%%%%%%%%
Note that $A = A_n \cup A_1$.  
By the splitting lemma, $ A_1$ is a belief set of $A'' \cup
\bigcup_{seq_n \in Act_{n}} [\Pi^1_{(D^r,res((seq_n),s_0)}]^A_{\bf
  A}$.  

Hence by \ref{For 1-plans,2} there exists an interpretation
$\Phi_2$ and a 0-interpretation $\Phi_0$ such that for any sequence of
$n$ actions $seq_n$ with $holds(true,res((seq_n),s_0)) \not\in A_n$,
%we can find a state $\sigma_{(seq_n)}$ that satisfies the following
%properties:
the following properties are satisfied:
\begin{enumerate} 
\item  $(\Phi_2,\Gamma_{\Phi_1}([seq_n],\Sigma))$ is 
 a 1-specific model  of $D^r$,

\item $(\Phi_0,\sigma_{(A_n,(seq_n))})$ is
in $\Asso{D^r}{\Phi_2}{\Gamma_{\Phi_1}([seq_n],\Sigma)}$.
%for every $\alpha \in P_n$,  
% verifying that,  
% for any fluent literal $f$, and every $\alpha \in P_n$ 

%\item For any fluent literal $f$, $holds(f,res((seq_n),s_0)) \in A''$ 
%if and only if $f\in\sigma_{(seq_n)}$.
  
\item $A_1 =$ \\ $\bigcup 
\{ A'_{(\Phi_0,\sigma_{(A_n,(seq_n))},res((seq_n),s_0))}:seq_n \in
Act_{n} \wedge holds(true,res((seq_n),s_0)) \not\in A_n \} \cup A''$.

\end{enumerate}

Defining $\Phi$ such that for any sequence of actions $seq_n \in Act_n$
 if the atomic formula $holds(true,res((seq_n),s_0))$ is not in $A_n$,
then $\Gamma_{\Phi}([seq_n],\Sigma)$ is equal to
$\Gamma_{\Phi_1}([seq_n],\Sigma)$, we will have that $(\Phi,\Sigma)$
is an $n$-specific model of $D$.  Moreover if we take $\Phi$ such that
for any action $a$ and any sequence of actions $seq_n \in Act_n$
, $\Phi(a,\Gamma_{\Phi}([seq_n],\Sigma)) =
\Phi_2(a,\Gamma_{\Phi}([seq_n],\Sigma))$, we will have that for any
sequence of actions $seq_n \in Act_n$,
$(\Phi,\Gamma_{\Phi}([seq_n],\Sigma))$ is a $1$-specific model of
$D^r$.  Therefore, $(\Phi,\Sigma)$ is an $n+1$-specific model of $D$.

%It is clear by construction that 
%if that  is equal to ${\bf
% A}^{n+1}_{(\Phi,\Sigma,s_0)}$.  By the splitting lemma we will have
%that $A \in {\bf A}$ iff
% $A = A_{n} \cup A_1$, where $A_1$ is a belief
%set of $A_n \cup \bigcup_{seq_n \in
%  Act_n}\Pi^1_{(D^r,res((seq_n),s_0)))}$.
 It is clear by construction that either 
% be a member of${\bf A}$ and $a$ an action.  
%by the definition of $\Phi$,that  for any
%sequence of actions $seq_n \in Act_n$ either the predicate 
the predicate $holds(true,res((seq_n),s_0))$ is a member of $A_n$, and by the
inertia rules, $A_1$ is the set of predicates of the form 
 $holds(true,res((seq_n,a),s_0))$ and
$holds(f,res((seq_n,a),s_0))$ with $a$ an action, 
 $(seq_n)$ a sequence of $n$ actions  and $f$ a
 fluent literal, or $holds(true,res((seq_n),s_0)) \not \in A_n$ and
for any sequence of $n$ actions $seq_n$  the  
  following propositions hold:
\begin{enumerate}

\item $(\Phi_0,\sigma_{(A_n,(seq_n))}) \in
  \Asso{D^r}{\Phi}{\Gamma_\Phi([seq_n],\Sigma)}$,

%\item $holds(f,res((seq_n),s_0)) \in A_n \Leftrightarrow f \in
%\sigma_{seq_n}$ and

\item $ A_1 = A'_{(\Phi_0,\sigma_{(A_n,(seq_n))},res((seq_n),s_0))}$ 

\end{enumerate}
%\cup
%  \{holds(f,res((seq_n,a),s_0)): f \in \sigma_{seq_n}\}$ is the set of
%  all predicates in $A$, with $res((seq_n,a),s_0)$ as their situation
%  argument.

Therefore by definition of $\bfA{n+1}$ we have proved that $A \in {\bf
  A}$ if and only if it is in $\bfA{n+1}$.

\medskip

\noindent
$(\Leftarrow)$
Reciprocally, let $(\Phi,\Sigma)$ be an $(n+1)$-specific model of $D$.  Then,
by \ref{For 1-plans}, induction and 
by the splitting lemma, to prove that ${\bf
  A}^{n+1}_{(\Phi,\Sigma,s_0)}$ is a world view of $\Pi^{n+1}_D$, it
suffices to show that for any $A$, $A\in {\bf A}^{n+1}_{(\Phi,\Sigma,s_0)}$
if and only if $A = A_n \cup B_A$ where $A_n$ is the set of predicates in
$A$ of $\leq n$-complexity and $B_A$ is a belief set of the program

$$ \Pi = A_n \cup [\Pi^{n+1}_D - \Pi^{n}_D]_
    {{\bf A}^{n+1}_{(\Phi,\Sigma,s_0)}}
=
A_n \cup [\bigcup_{seq_n \in Act_n} \Pi^1_{(D^r,res((seq_n),s_0))}]_ 
    {{\bf A}^{n+1}_{(\Phi,\Sigma,s_0)}}
$$

In order to prove this, 
 we will show that the beliefs sets of the program $\Pi$
% $A_n \cup 
%[\bigcup_{seq_n \in Act_{n}} \Pi^1_{(D^r,res((seq_n),s_0))}
%]_{\bfA{n+1}}$
 are exactly the sets  of the form $A_n \cup A_1$ defined in 
 \ref{bfA}.

First note that the belief sets of $[\bigcup_{seq_n \in Act_{n}}
\Pi^1_{(D^r,res((seq_n),s_0))}]_{\bfA{n+1}}$ are the unions of belief
sets of $[\Pi^1_{(D^r,res((seq_n),s_0))}]_{\bfA{n+1}}$ where $seq_n$
is varying over the set of sequences of $n$ actions.
% a sequence of  
%$n$ actions,
This is so because these programs are independent of each other.

Hence, we will calculate the belief sets of each program $
\Pi^1_{(D^r,res((seq_n),s_0))}]_{\bfA{n+1}}$ where $seq_n \in Act_n$
and we will prove that the set of these union is ${\bf
A}^{n+1}_{(\Phi,\Sigma,s_0)}$.

To show this, let $seq_n$ be a sequence of 
$n$ actions. Then there are two possible cases:

\begin{enumerate}
\item $ holds(true,res((seq),s_0)) \in A_n$,
 and $ A_n \cup [\Pi^1_{(D^r,res((seq_n),s_0))}]_{\bfA{n+1}}$ 
has only the belief set $B_A = A_n \cup A_1$ with $A_1$ the set of 
predicates of the form $holds(true,res((seq,a),s_0))$ or 
$holds(f,res((seq,a),s_0))$ where $a$ is any action and $f$ is 
any fluent literal.

\item $holds(true,res((seq),s_0)) \not \in A_n$ and by theorem [\ref{For 
1-plans,2}], any belief set of

 $A_n \cup
 [\Pi^1_{(D^r,res((seq_n),s_0))}]_{\bfA{n+1}}$

 is equal to $A_n \cup 
 A'_{(\Phi_0,\sigma_{(A_n,seq_n)},res((seq_n),s_0))}$ for some 
 0-interpretation such that $(\Phi_0,\sigma_{(A_n,seq_n)}) \in 
Asso_{D^r}(\Phi,\Gamma_{\Phi}(([seq_n,a],\Sigma))$. 

\end{enumerate}

Therefore the belief sets of $ A_n \cup
[\Pi^1_{(D^r,res((seq_n),s_0))}]_{\bfA{n+1}}$ are precisely the elements on
$\bfA{n+1}$, and this implies that $\bfA{n+1}$ is a world view of
$\Pi^{n+1}_D$.
\qed

%\hspace{\fill}$\Box$

\begin{definition}
  Given a domain description $D$, an initial situation constant $s_0$
  and a specific-model $(\Phi,\Sigma)$ of $D$, we will denote by
  $\bfA{\omega}$ the following family of sets, $A \in {\bf
    A}^{\omega}_{(\Phi,\Sigma,s_0)}$ if and only if for any $n \geq 1$
  there exist $A_n \in \bfA{n}$ and $A_{n+1} \in \bfA{n+1}$ with $A_n
  \subseteq A_{n+1}$, and $A = \bigcup_{n\geq 1}A_n$
  We will denote by ${\bf A}^{Q}_{(\Phi,\Sigma,s_0)}$ the family of
  sets $A$, such that $A$ is the union of sets $A_{\Pi}$ and $A_Q$
  where $A_{\Pi}$ is an element of ${\bf
    A}^{\omega}_{(\Phi,\Sigma,s_0)}$ and $A_Q$ is a belief set of
  $A_{\Pi} \cup [Q]^{A_{\Pi}}_{{\bf A}^{\omega}_{(\Phi,\Sigma,s_0)}}$
\end{definition}

 \begin{lemma}
\label{bf A Q}
Given a consistent domain description $D$ and the initial situation
constant $s_0$, ${\bf A}$ is a world view of $\Pi_D$, if and
only if there exists a specific model $(\Phi,\Sigma)$ of $D$ such that
${\bf A} = {{\bf A}^{Q}_{(\Phi,\Sigma,s_0)}}$.
\end{lemma}

\subsection*{Proof}
Given a world view ${\bf A}$  of  $\Pi^{\omega}_D$, 
and  $A \in {\bf A}$, 
we will denote by $A_{\Pi}$ the subset of $A$ restricted to 
those predicates in $A$ of the form $holds(f,s)$ or $ab(f,a,s)$ where 
$s$ is a situation constant and $a$ is an action, then the set 
${\bf A}_{\Pi} = \{A_{\Pi}: A \in {\bf A}\}$ 
is a world view of $\Pi_D$, and  
  by (\ref{bf A is a world view}) 
there exists a specific model of $\Pi_D$ such 
that ${\bf A}_{\Pi}$ is equal to $\bfA{\omega}$,   
since $[Q]^A_{\bf A}$ is equal to 
$[Q]^A_{\bfA{\omega}}$,  applying the splitting lemma to 
$[\Pi^{Q}_D]^A_{\bf A} = [\Pi_D]^A_{\bf A} 
\cup [Q]^A_{\bf A}$ we have the result.   
\qed

\begin{theorem}
\label{seq(plan,s)}
Let $D$ be a domain description, $s_0$ be an initial constant
situation and $(\Phi,\Sigma)$ be a specific model of $D$. Then, given
a sequence of actions $seq = a_1,\ldots,a_k$, and the situation
constant $s = res((seq),s_0)$, we will have that, for any situation
constant $s_1$ and any plan $\beta$, there exists a sequence of
actions $seq_{(\beta,s)}$ such that $\Gamma_{\Phi}
([seq,seq_{(\beta,s)}],\Sigma)$ is equal to $\Gamma_{\Phi}(\beta,
\Gamma_{\Phi}([seq],\Sigma)$ and for any $A \in \bfA{Q}$,
$find\_situation(\beta,s,s_1)$ belongs to $A$ if and only if $s_1
=res((seq,seq_{(\beta,s)}),s_0)$.
% Moreover for any plans $\alpha,\alpha'$ and 
%any situation constant $s = res((seq),s_0)$, we will have that the sequence  
%$seq_{([\alpha,\alpha'],s)}$ is equal to   
%$[seq_{(\alpha,s)},seq_{(\alpha',res((seq_{(\alpha,s)}),s_0))}]$.

\end{theorem}

\subsection*{Proof}
The proof will be a double induction, using the loop nesting in the
plan and the complexity of the plans.  Thus, the while-complexity of a
plan $\beta$, denoted by $wcomp(\beta)$, is 0 if $\beta$ is either the
empty plan $[]$ or an action $a$,
$max(wcomp(\alpha_1),wcopm(\alpha_2)$ If $\beta =\If
\varphi \Then \alpha_1 \Else \alpha_2$ or $\beta =
[\alpha_1|\alpha_2]$, $1+wcomp(\alpha)$ if $\beta = \While \varphi \Do 
\alpha$.

Taking $wcomp(\beta) = 0$ we will do induction on $comp(\beta)$.
\begin{itemize}

\item If $\beta = []$, the result is immediate, because
  $$\Gamma_{\Phi}(\beta,\Gamma_{\Phi}([seq],\Sigma)) = \Gamma_{\Phi}( [
  ],\Gamma_{\Phi}([seq],\Sigma)),$$ and for any $A \in \bfA{Q}$
  $find\_situation([\beta],s,s_1)$ is in $A$ if and only if $s_1 = s$.
  Therefore taking $seq_{(\beta,s)} = seq_{\emptyset}$ the claim
  follows.

\item Suppose the theorem is valid for  $\leq n$-plans.

\item Let $\beta $ be an $(n+1)$-plan. Then we have the following
  possibilities

  \noindent I) $\beta = [a|\alpha]$ where $\alpha$ is an $n$-plan and
  $a$ is an action. Then, by inductive hypothesis
  $\Gamma_{\Phi}(\alpha,\Gamma_{\Phi}([seq,a],\Sigma))$ is equal to
  $\Gamma_{\Phi}([seq,a,seq_{(\alpha,res(a,s))}],\Sigma)$, therefore
  we will have that
$$
\begin{array}{l}
\Gamma_{\Phi}(\beta,\Gamma_{\Phi}([seq],\Sigma)) = \\
\Gamma_{\Phi}([a,\alpha],\Gamma_{\Phi}([seq],\Sigma)) = \\
 \Gamma_{\Phi}(\alpha,\Gamma_{\Phi}([seq,a],\Sigma)) = \\
 \Gamma_{\Phi}([seq_{(\alpha,res(a,s))}],\Gamma_{\Phi}([seq,a],\Sigma)) = \\ 
\Gamma_{\Phi}([a,seq_{(\alpha,res(a,s))}],\Gamma_{\Phi}([seq],\Sigma)) = \\ 
\Gamma_{\Phi}([seq_{(\beta,s)}],\Gamma_{\Phi}([seq],\Sigma)) = \\
 \Gamma_{\Phi}([seq,seq_{(\beta,s)}],\Sigma).
\end{array}
$$

Moreover for any $A$ in $\bfA{Q}$, $find\_situation(\beta,s,s_1)$ is
in $A$ if and only if\\
$find\_situation(\alpha,res(a,s),s_1)$ is in
$A$.

Thus, if we take $seq_{(\beta,s)} = (seq_{(\alpha,res(a,s))},a)$,
since by inductive hypothesis $ find\_situation([\alpha],res(a,s),s_1)$
is in $A$ if and only if $s_1$ is equal to
$res(((seq,a),seq_{(\alpha,res(a,s))}),s_0)$, we have that
$find\_situation([\beta],s,s_1)$ is in $A$ if and only if $s_1 =
res((seq,(a,seq_{(\alpha,s)})),s_0)$ which is equal to the situation
$res((seq,seq_{(\beta,s)}),s_0)$, and
$$
\begin{array}{l}
\Gamma_{\Phi}([\beta],\Gamma_{\Phi}([seq],\Sigma)) = \\
 \Gamma_{\Phi}([seq,seq_{(\beta,s)}],\Gamma_{\Phi}([seq],\Sigma)) = \\
\Gamma_{\Phi}([seq_{(\beta,s)}],\Gamma_{\Phi}([seq],\Sigma)) 
\end{array}
$$
 
\noindent II) $\beta = [\If\; \varphi \;\Then \alpha_1,\alpha_2]$ where
$\alpha_1, \alpha_2$ are $n_1$ and $n_2$-plans respectively, with $n_1
+ n_2 = n$ then, we have that for any $A$ in $\bfA{Q}$,
$find\_situation(\beta,s,s_1)$ is in $A$ if and only if either:
\\
i) $holds(\o \varphi,s)$ holds in $\bfA{Q}$ and
$find\_situation([\alpha_2],s,s_1)$ is in $A$,

or 

ii) $holds(\varphi,s)$ holds in $\bfA{Q}$ and
$find\_situation([\alpha_1,\alpha_2],s,s_1)$ is in $A$.

By inductive hypothesis
$\Gamma_{\Phi}(\alpha_2,\Gamma_{\Phi}([seq],\Sigma))$ is equal to
$\Gamma_{\Phi}([seq,seq_{(\alpha_2,s)}],\Sigma)$ and
$\Gamma_{\Phi}([\alpha_1,\alpha_2],$ $\Gamma_{\Phi}([seq],\Sigma))$ is
equal to $\Gamma_{\Phi}([seq,seq_{([\alpha_1,\alpha_2],s)}],\Sigma)$
for the sequences of actions $seq_{(\alpha_2,s)}$ and
$seq_{([\alpha_1,\alpha_2],s)}$

%Therefore $find-situation (\beta,s,s_1)$ is in $A$ if and only if, either %
%
%i) $holds(\o \varphi,res((seq),s_0))$ holds  in $\bfA{Q}$ and   
%$find-situation([\alpha_2],s,s_1)$ 
%is in $A$ and $\Gamma_{\Phi}(\beta,\Gamma_{\Phi}((seq),\Sigma) = 
%\Gamma_{\Phi}(\alpha_2,\Gamma_{\Phi}([seq],\Sigma))$ 
%
%or 
%
%ii) $hols(\varphi,res((seq),s_0)$ holds in $\bfA{Q}$ and 
%$find-situation([\alpha_1,\alpha_2],s,s_1)$ 
%is in $A$. 

Hence by 
[\ref{f in Phi iff holds(f) in A}] and 
inductive hypothesis, $find\_situation(\beta,s,s_1)$ 
is in $A$ if and only if either 

i) $\o \varphi$ holds in $\Gamma_{\Phi}([seq],\Sigma)$, and     
%$holds(\o \varphi,res((seq),s_0))$ holds  in $\bfA{Q}$, 
  $s_1 = res((seq,seq_{(\alpha_2,s)}),s_0)$. 

or 

ii)   $\varphi$ holds in $\Gamma_{\Phi}([seq],\Sigma)$, and     
%$holds(\o \varphi,res((seq),s_0))$ holds  in $\bfA{Q}$, 
  $s_1 = res((seq,seq_{([\alpha_1,\alpha_2],s)}),s_0)$.  

Thus, if we take $seq_{(\beta,s)} = seq_{(\alpha_2,s)}$ in case i), and in case 
ii)  $seq_{(\beta,s)}$ equal to  $seq_{([\alpha_1,\alpha_2],s)}$
 we will have that 
$\Gamma_{\Phi}(\beta,\Gamma_{\Phi}([seq],\Sigma)) = 
\Gamma_{\Phi}([seq,seq_{(\beta,s)}],\Sigma))$ and 
$find\_situation(\beta,s,s_1)$, belongs to $A$ if and only if 
$s_1$ is equal to the situation $res((seq,seq_{(\beta,s)}),s_0)$. 

\noindent
III) $\beta = [\If\; \varphi\;\Then\alpha_1\;\Else\alpha_1',\alpha_2]$, 
where $\alpha_1,\alpha_1'$ and $\alpha_2$ are $n_1,n_1'$ and $n_2$-plans (resp.)
 with $max(n_1,n_1')+n_2 = n$.  
This case is  similar to the previous one.

\noindent IV) $\beta = [\While\;\varphi\;\Do\;\alpha_1,\alpha_2]$,
where $\alpha_1$ and $\alpha_2$ are $n_1$ and $n_2$-plans (resp.),
with $n_1 + n_2 = n$. Here we may suppose by inductive hypothesis that
for any $k \geq 0$ the plans $[\alpha_1^k]$ and
$[\alpha_1^k,\alpha_2]$ verify the theorem, and we will denote by
$seq_{(k,\alpha_1)}$ the sequence $seq_{(\alpha_1,s)}$ and by
$seq_{(k,\alpha_2)}$ the sequence
$seq_{([\alpha_2],res((seq_{([\alpha_1^k],s)}),s_0))}$.

Using the fix-point operator ${\bf T}_\Pi$ defined by ${\bf T}_\Pi(I)
= \{p: \exists$ a rule $p \leftarrow q_1,\ldots,q_n$ in $\Pi$ with
each $q_i$ a fact in $I \}$, for any positive logic program $\Pi$,  we
know that if ${\bf T}_\Pi\!\!\uparrow^1$ is defined to be equal to ${\bf
  T}_\Pi(\emptyset)$ and ${\bf T}_\Pi\!\!\uparrow^{k+1} = {\bf
  T}_\Pi({\bf T}_\Pi\!\!\uparrow^k)$, then ${\bf
  T}_\Pi\!\!\uparrow^{\omega}$, which is the set $\bigcup_{k \geq
  1}{\bf T}_\Pi\!\!\uparrow^k$, is a fix-point for ${\bf
  T}_\Pi$.  Moreover $A$ is a belief set for
$[\Pi^{\omega}_{(D,s_0)}]^A_{\bfA{Q}}$ if and only if $$A = {\bf
  T}_{[\Pi^{\omega}_{(D,s_0)}]^A_{\bfA{Q}}}\!\!\uparrow^\omega.$$
Therefore, $h = find\_situation(\beta,s,s_1)$ is in $A$ if and only if
there exists $k \geq 0$ such that $h $ is in ${\bf
  T}_{[\Pi^{\omega}_{(D,s_0)}]^A_{\bfA{Q}}}\!\!\uparrow^{k+1}$.

Let $k_0$ be the minimum $k$ such that $h$ belongs to ${\bf
  T}_{[\Pi^{\omega}_{(D,s_0)}]^A_{\bfA{Q}}}\!\!\uparrow^{k+1}$.  We
have that $h \in {\bf
  T}_{[\Pi^{\omega}_{(D,s_0)}]^A_{\bfA{Q}}}\!\!\uparrow^{k_0+1}$, if
and only if there exists $m$ such that the following properties are
satisfied:
\\
1) For any $j < m$  $holds(\varphi,res((seq,seq_{(j,\alpha_1)}),s_0)$ holds in 
$\bfA{Q}$ \\
2) $holds(\o \varphi,s')$ holds in $\bfA{Q}$ \\
3) $find\_situation([\alpha_1^m],s,s') \in A$ \\ 
4) $find\_situation([\alpha_2],s',s_1) \in A$ \\
If we fix $m$ with properties (1), (2), (3) and (4), we have
by inductive hypothesis and
[\ref{f in Phi iff holds(f) in A}] that, $h \in A$ if and only if
 $\varphi$ holds in $\Gamma_{\Phi}([seq,seq_{(m,\alpha_1)}],s_0)$,
  $s' = res((seq_{(m,\alpha_1)}),s_0)$ and
 $s_1 = res((seq_{(m,\alpha_2)}),s_0)$. Hence taking
 $seq_{(\beta,s)} = seq_{(m,\alpha_2)}$, we will have
 that $\Gamma_{\Phi}(\beta,\Gamma_{\Phi}([seq],\Sigma)$
 is equal to $\Gamma_{\Phi}([seq,seq_{(\beta,s)}],\Sigma)$ and
  $h \in A$ if and only if $s_1 = res((seq_{(\beta,s)}),s_0)$.

\end{itemize}
The inductive step on $wcomp(\beta)$ follows the same reasoning as in
the base case.
\qed

\begin{corollary}
  Given a consistent domain description $D$, the initial situation
  constant $s_0$ and a specific model of $D$, $(\Phi,\Sigma)$, we will
  have that for any fluent $f$ and any plan $\beta$,
  $hold\_after\_plan(f,\beta)$ holds in $\bfA{Q}$ if and only if $f
  \in \Gamma_{\Phi}(\beta,\Sigma)$.

\end{corollary}

\subsection*{Proof}
Let $seq_{(\beta,s_0)}$ be the sequence of actions described in
[\ref{seq(plan,s)}], such that $\Gamma_{\Phi}(\beta,\Sigma) =
\Gamma_{\Phi}([seq_{(\beta,s_0)}],\Sigma)$ and
$find\_situation(\beta,s_0,s_1)$ holds in $\bfA{Q}$ if and only if
$s_1 = res((seq_{(\beta,s_0)}),s_0)$. Then since
$hold\_after\_plan(f,\beta)$ holds in $\bfA{Q}$ if and only if
$find\_situation(\beta,s_0,s_1)$ and $holds(f,s_1)$ hold in $\bfA{Q}$,
we have by [\ref{f in Phi iff holds(f) in A}]
% [\ref{bf A is a world view}] that
$hold\_after\_plan(f,\beta)$ holds in $\bfA{Q}$ if and only if $f \in
\Gamma_{\Phi}([seq_{(\beta,s_0)}],\Sigma)=
\Gamma_{\Phi}(\beta,\Sigma)$.
\qed

Hence by [\ref{bf A Q}] we have the following:\\

\begin{corollary} \label{MAIN}
  Given a \underline{simple} and consistent domain description $D$ and
  a plan $\beta$.  $D \models F \After \beta$ if and only if $\Pi^Q_D
  \models hold\_after\_plan(F,\beta)$.

\end{corollary}

\paragraph{Theorem 7.3}
  Given a simple consistent domain description $D$ and
  a plan $\beta$.  $D \models F \After \beta$ if and only if $\Pi^Q_D
  \models hold\_after\_plan(F,\beta)$.

\noindent
\medskip
{\bf Proof:} Direct from Corollary \ref{MAIN}.

\end{document}